\documentclass{article}
\PassOptionsToPackage{numbers,sort&compress}{natbib}
\usepackage[dblblindworkshop, final]{neurips_2026}
\workshoptitle{Agentic AI for Biological Discovery}
\usepackage[utf8]{inputenc}
\usepackage[T1]{fontenc}
\usepackage{amsmath,amssymb}
\usepackage{graphicx,booktabs,array,longtable}
\usepackage{pdflscape,placeins,float}
\usepackage{microtype}
\usepackage{xcolor}
\usepackage{xurl}
\usepackage[hidelinks]{hyperref}
\hypersetup{pdftitle={From Tool State to Scientific Action: Auditing Decision Provenance in Biomedical Research Agents},pdfauthor={Xingyu Yi; Xinya Zhang; Yunshen Zhou; Haoyu Bao},pdfsubject={NeurIPS 2026 Workshop: Agentic AI for Biological Discovery (AgenticLS)},pdfkeywords={}}

\DeclareUnicodeCharacter{2212}{\ensuremath{-}}
\title{From Tool State to Scientific Action: Auditing Decision Provenance in Biomedical Research Agents}
\author{\href{https://openreview.net/profile?id=~Xingyu_Yi1}{Xingyu Yi}\textsuperscript{1}\thanks{Corresponding author.}\quad
\href{https://openreview.net/profile?id=~Xinya_Zhang1}{Xinya Zhang}\textsuperscript{1,2}\quad
\href{https://openreview.net/profile?id=~Yunshen_Zhou1}{Yunshen Zhou}\textsuperscript{3}\quad
\href{https://openreview.net/profile?id=~Haoyu_Bao2}{Haoyu Bao}\textsuperscript{1}\\[5pt]
\normalfont\small
\textsuperscript{1}\href{https://intosyn.com}{INTOSYN AI Labs}\quad
\textsuperscript{2}\href{https://columbia.edu}{Columbia University}\quad
\textsuperscript{3}\href{https://microsoft.com}{Microsoft}\\
\normalfont\small\texttt{\{xingyu.yi,haoyu.bao\}@intosyn.com}\\
\normalfont\small\texttt{sz3475@tc.columbia.edu}\quad\texttt{yunshenzhou@microsoft.com}
}
\begin{document}
\maketitle
\begin{abstract}
A scientific agent can give the same verdict while reading and citing different evidence. BioAgent-DF tests this separation using paired reference and candidate values saved at the same scientific computation site. It restores or transplants those values at their original coordinates, then runs the agent on healthy hardware. Across eight SciFact batches, Qwen3.5-4B shows repeatable citation changes in three capture--claim combinations, and Gemma 3 4B IT shows repeatable read and citation changes in four. These seven combinations cover two claims. Full restoration recovers the reference evidence sets while verdicts remain unchanged. Verdict changes observed under the baseline prompt did not recur under the explicit-example prompt. Complementary tool and optimizer experiments distinguish recovery of numerical state, recovery of ranked lists, and restoration of an agent's visible history. These results show that an audit should track both the evidence an agent uses and its final answer. In hERG cardiac-safety assay allocation, silent numerical faults changed testing decisions while the fixed-budget shortlist kept the same number of compounds labeled as blockers. After the agent had used faulty values, restoring its earlier execution state and supplying certified scores recovered all 39 changed decisions, whereas replacing the prediction file recovered none; no incorrect action was committed. A streaming implementation of the arithmetic checks reduced the total time for forward execution and verification by 42.4\% relative to the original FP64 implementation, with identical flags and recovery outputs.
\end{abstract}
\section{Introduction}\label{sec:1}

Biomedical research agents retrieve literature, inspect molecular and cellular features, and choose what to examine next. The final verdict captures only one part of this process. Two executions can agree on a claim while selecting different documents or citing different sentences. We ask whether silently corrupted tool values change those evidence choices, and how much of the tool's output must be restored to undo the change.

Evidence-path fidelity means agreement with a specified reference execution in recorded retrieval, read, and citation endpoints. Scientific decision fidelity (the DF in BioAgent-DF) is the broader audit, covering actions, item selection, and verdicts as well.

We investigate silent data corruption (SDC) observed on faulty GPUs during distributed training and its downstream effects on scientific agents. Finite, erroneous outputs can pass ordinary NaN/Inf monitoring and reach a tool result. The faulty GPUs fall into two classes: \textbf{Case 1 (size-dependent)} GPUs fail more often on tested large-matrix paths than on short probes; \textbf{Case 2 (expanded-trigger)} GPUs also fail on those short probes.

\begin{figure}[!tbp]
\centering\includegraphics[width=0.78\linewidth,height=0.78\textheight,keepaspectratio]{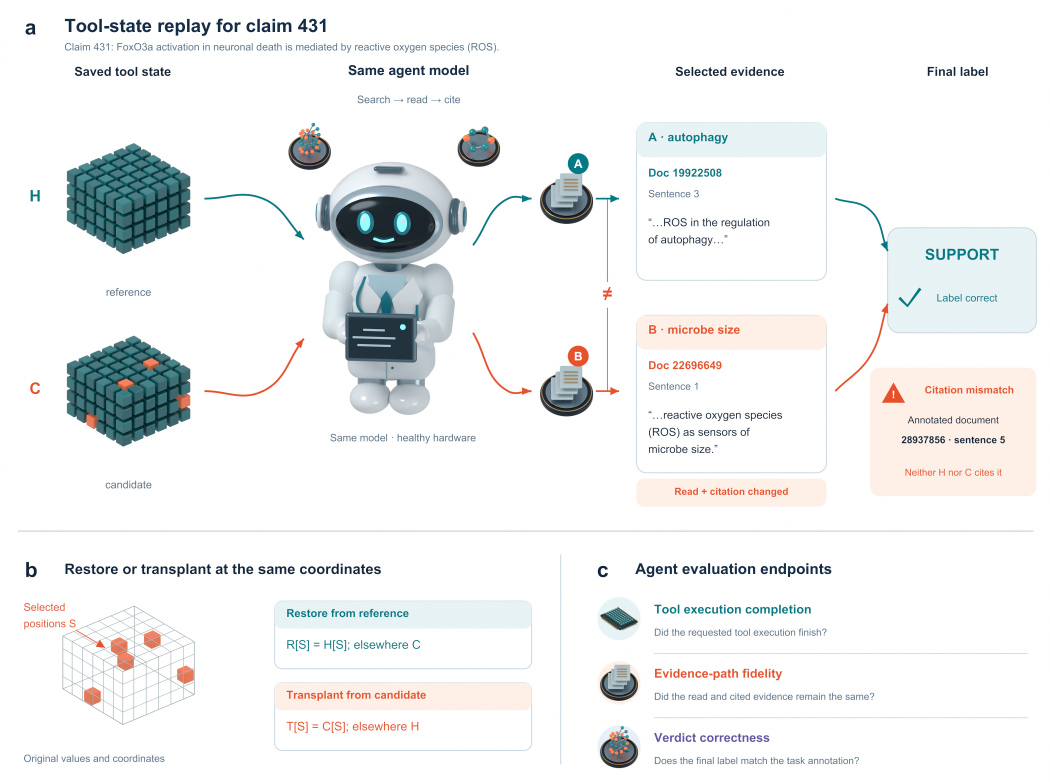}
\caption{The same verdict from different cited evidence. (a) Reference and candidate tool states are supplied to the same agent model on healthy hardware. Claim 431 concerns ROS-mediated FoxO3a activation in neuronal death. Gemma's reference branch cites a sentence about ROS and autophagy (19922508/3), and its candidate branch cites ROS sensing microbe size (22696649/1). Neither sentence supports the stated FoxO3a mechanism, yet both branches return the benchmark-correct SUPPORT label. Neither citation matches the annotated evidence, 28937856/5. Sentence indices are zero-based and document identifiers are SciFact IDs. (b) Restoration and transplantation select values at their original coordinates. (c) Tool execution completion, evidence-path fidelity, and verdict correctness are evaluated. H1 and H2 repeat the same reference state.}\label{fig:1}
\end{figure}

BioAgent-DF uses controlled tool-state interventions to trace and restore biomedical agents' recorded evidence paths (Figure~\ref{fig:1}). We save reference and candidate values at one computation site, apply replacements at their original coordinates, and execute the remaining tool and agent steps on healthy hardware. The experiments test the numerical validity of the paired values, their effect on recorded reads and reported citations, and the replacements that recover reference endpoints under matched continuation conditions (Appendices A--H).

We also test whether numerical errors redirect measurements on prespecified screening inputs and alter hERG assay allocation. Restoring reference values at their original coordinates identifies values responsible for changed actions; selective recomputation repairs decisions, and certification determines when they can authorize action under a fixed computation contract. The hERG experiment compares replacing the prediction file with restoring the execution state after the agent has used faulty values (Appendices~\ref{app:I}--\ref{app:commit}).

\section{Auditing evidence-path fidelity}\label{sec:2}

\subsection{From arithmetic constraints to agent observations}\label{sec:2.1}

With inputs, model, representation, and document order matched, we save reference and candidate states \(h\) and \(c\) at the same computation site, and call each saved candidate state a capture. A tool returns \(o_{t} = T\left( x_{t},u_{t} \right)\), where \(x_{t}\) is its state and \(u_{t}\) its request. The agent selects \(a_{t} = \pi\left( H_{t} \right)\) from the visible history \(H_{t} = \left( o_{1},a_{1},\ldots,o_{t} \right)\). In SciFact, search returns document IDs, titles, and scores. The agent sees these scores but not the correct values.

Numerical comparisons use a rounding-error bound on multiplication, accumulation, intermediate rounding, and final conversion \citep{ref33,ref34} (Appendix A.3).

Exact-integer probes provide a simpler check. Their represented products and all permitted partial sums are exactly representable in FP32, so a correct result must be an integer. The observed value 405613.25, against the exact value 405760, cannot arise under that construction. In the same compiled kernel and destination registers, three runs give 38, 24, and 54 mismatching outputs from the tensor-core matrix multiply-accumulate instruction (MMA), compared with zero for the paired scalar FP32 fused multiply-add (FFMA). FFMA replacement on the target streaming multiprocessor (SM) and non-target sham routing further identify the affected execution path in both GPU classes (Appendix A).

Healthy implementations also differ. In PBMC, disabling intermediate low-precision BF16 reduction reduces Top-10 membership disagreements between references from 174 queries to zero, although two order disagreements remain. We therefore distinguish arithmetic anomalies supported by an applicable bound, differences with unresolved references, ordinary precision effects, and constructed interventions.

\subsection{Tool-state interventions}\label{sec:2.2}

At the saved site, states \(h,c,z\in\mathcal X\) share shape, coordinate indices \(\mathcal I\), and representation. We save the residual \(r=FP32(c)-FP32(h)\). For \(S\subseteq\mathcal I\), \(R_S=select(S,h,c)\) and \(T_S=select(S,c,h)\).

Selection takes the first state inside \(S\), the second elsewhere; \(R_S\) and \(T_S\) perform restoration and transplantation, respectively. Residual implementations are checked after conversion \(Q\) to the recorded storage representation, using empty, full, and zero-residual controls (Appendix B.1).

Let \(F_e\) continue in environment \(e\in\{A,B\}\), with reference \(h_e\in\mathcal X\) at matched inputs, model, and site. For fixed \(F\) and endpoint map \(D\), restoration tests \(D(F(R_S))=D(F(h))\), and transplantation tests \(D(F(T_S))=D(F(c))\). \(F_A(z),F_B(z)\) share any compatible state; \(F_A(Q(h_A+r)),F_B(Q(h_B+r))\) share a residual on environment-specific references. Agent repeats measure within-state variability.

\subsection{Endpoints}\label{sec:2.3}

For an execution environment \(e\) and initial state \(z\), let \(\tau_e(z)=F_e(z)\) denote the resulting agent trajectory. For any trajectory \(\tau\), define

\[P(\tau) = \left( E_{ret},E_{read},E_{cite} \right),\quad\quad D(\tau) = \left( P(\tau),A,V \right).\]

Here \(E_{ret},E_{read},E_{cite}\) record retrieved documents, read documents, and cited sentences; \(A,V\) record structured actions and verdict. We compare full trajectories or named components by membership, order, and repeated access. On fixed evaluation data, \(M_e:\mathcal X\to\mathbb R\) measures continuation quality: \(M_e(h)=M_e(c)\) can coexist with \(P(\tau_e(h))\ne P(\tau_e(c))\).

Primary paired analyses include only groups in which H1, H2, C, and full restoration all complete and access the intervention site (exposed groups). Comparisons without a restoration endpoint require H1, H2, and C. Candidate--reference and reference-repeat differences use the same subset. Failed, interrupted, unexposed, and invalid executions are reported alongside verdict changes.

\subsection{Global sparsity and recovery within the visited region}\label{sec:2.4}

A small fraction of a full score array can contain every changed value that an agent encounters. For a region \(\Omega\subseteq\mathcal I\) specified before confirmation and \(\Delta=\{i\in\mathcal I:h_i\ne c_i\}\), we report

\[b_{global} = C_{\mathrm{frac}}(S),\quad\quad f_{exposed} = \frac{|S \cap \Omega \cap \Delta|}{|\Omega \cap \Delta|},\quad\quad n_{remain} = |\Omega \cap \Delta \smallsetminus S|.\]

Here \(C_{\mathrm{frac}}\) is the selected fraction of eligible blocks (cost units: Appendix C.4). The first SciFact query row defines \(\Omega\); later visits are logged. A zero denominator means the region has no changed score. Partial-row recovery requires endpoint recovery with \(0<f_{exposed}<1\) and \(n_{remain}>0\). For claim 831, 8 of the 38 row blocks replace every changed score in the visited row: \(f_{exposed}=1\), \(n_{remain}=0\). This is full-row recovery despite global sparsity (Appendix \ref{sec:J.1}).

\subsection{Paired audit and recovery cost}\label{sec:2.5}

\textbf{Algorithm 1. Paired audit.} Fix task, paired states, exposed region, coordinate priorities, and continuation; check inputs, document identities, model, representation, and reference. Construct H1/H2/C, full/selected restoration, transplantation, and zero treatment (a zero residual); verify reconstruction. Interleave references with other conditions, saving observations, actions, and the first divergent response. Compare jointly completed, exposed groups; report replacement cost, remaining differences, and incomplete attempts.

Given a set of allowed edits and a cost function, \(\rho_D\) and \(\rho_{\mathrm{state}}\) are the minimum costs of recovering the reference endpoint and the exact state (\(R_S=h\)). Deterministic continuation with other relevant state fixed gives \(\rho_D\leq\rho_{\mathrm{state}}\); a successful budget upper-bounds the minimum. Score-error bounds give sufficient conditions for Top-K stability \citep{ref20,ref21,ref27} (Appendix C.4). Reads and citations still have to be checked directly, because the agent sees the scores and its outputs vary across identical requests.
When reference predictions are unavailable, valid error bounds and trusted recomputation can certify a decision before it authorizes action.

\section{SciFact agent experiments}\label{sec:3}

\subsection{Agent protocol}\label{sec:3.1}

The SciFact agent executes \texttt{search}, \texttt{read}, and \texttt{finish}, returning SUPPORT, REFUTE, or NOT\_ENOUGH\_INFO (NEI) with sentence citations \citep{ref25,ref26}. Tools check that cited sentences were delivered by a read action; only the evaluator sees the evidence annotations. The confirmation protocol binds the original claim to its saved query row, permits one search and up to six document reads, and limits each episode to 16 calls and each response to 384 output tokens. H and C may differ in retrieved membership, order, and visible scores; this experiment evaluates their combined effect.

When the agent model itself ran on Case 1 hardware, all 220 candidate requests triggered vLLM's request-level NaN diagnostic, and no tool calls were completed. An earlier Case 2 series records 240 flagged requests and no candidate completions. Each cohort ran Qwen3.5-4B BF16 through vLLM 0.28.0 on its own class of faulty RTX GPU (Appendix F.2). The evidence-path experiment replays finite saved retrieval scores to agents on healthy hardware.

Qwen3.5-4B~\citep{ref58} and Gemma 3 4B IT~\citep{ref59} use their official BF16 weights and matched tasks, arrays, masks, and budgets. Reference measurements taken around the capture rounds at query batches 8 and 300 match independent healthy outputs bitwise. Within each Case and query batch, captures 0--1 define a priority of one coordinate per query; captures 2--7 form the confirmation subset. Case 1 initially shows numerical differences in 15/16 captures and Top-10 membership changes in 0/16.

For the mechanism study, we select 27 Case 2 capture--claim combinations, covering 11 claims, by upstream Top-10 change. Three technical repeats of each request at temperature zero, across seven conditions, give 567 attempts per model. A later explicit-example prompt adds JSON examples and explanations of rejected calls to the baseline prompt, with the same tasks, budgets, sampler, and sampling settings. With content-masking and unselected controls, each prompt has 1,386 attempts per model.

Baseline Qwen completes 1,386 attempts; baseline Gemma completes 800, with 559 budget exhaustions and 27 invalid-JSON outputs. With explicit examples, Qwen completes 1,372 and Gemma 1,078; the other 322 attempts across the two models exhaust their budgets (Appendix H).

\subsection{Path and verdict endpoints}\label{sec:3.2}

\begin{table}[!tbp]
\caption{\textbf{Endpoints by prompt.} Each endpoint cell reports candidate--H1 / H2--H1 discordance among jointly completed, exposed groups. A group is one repeat of a capture--claim combination.}\label{tab:1}
\centering\small\setlength{\tabcolsep}{3pt}
\begin{tabular}{@{}>{\raggedright\arraybackslash}p{\dimexpr 0.35000\linewidth-4.8pt\relax}>{\raggedright\arraybackslash}p{\dimexpr 0.20000\linewidth-4.8pt\relax}>{\raggedright\arraybackslash}p{\dimexpr 0.15000\linewidth-4.8pt\relax}>{\raggedright\arraybackslash}p{\dimexpr 0.15000\linewidth-4.8pt\relax}>{\raggedright\arraybackslash}p{\dimexpr 0.15000\linewidth-4.8pt\relax}@{}}
\toprule
\textbf{Prompt / model / query batch} & \textbf{Joint groups / scheduled} & \textbf{Read} & \textbf{Cite} & \textbf{Verdict} \\
\midrule
Baseline / Qwen / 8 & 30/30 & 0 / 0 & 0 / 0 & 1 / 0 \\
Baseline / Qwen / 300 & 51/51 & 12 / 0 & 2 / 0 & 2 / 0 \\
Baseline / Gemma / 8 & 3/30 & 0 / 0 & 0 / 0 & 0 / 0 \\
Baseline / Gemma / 300 & 33/51 & 2 / 0 & 8 / 0 & 6 / 0 \\
Explicit examples / Qwen / 8 & 30/30 & 0 / 0 & 0 / 0 & 0 / 0 \\
Explicit examples / Qwen / 300 & 48/51 & 0 / 0 & 9 / 0 & 0 / 0 \\
Explicit examples / Gemma / 8 & 15/30 & 0 / 0 & 0 / 0 & 0 / 0 \\
Explicit examples / Gemma / 300 & 42/51 & 15 / 0 & 15 / 0 & 0 / 0 \\
\bottomrule
\end{tabular}
\end{table}

Each combination is run three times at temperature zero, so Table~\ref{tab:1} counts three groups per capture--claim combination. With explicit examples, Qwen's nine citation changes are three captures of claim 431: both branches read the same document, H cites one sentence, C cites none, and both return the incorrect NEI. Gemma's 15 read and citation changes are five combinations for claims 343, 431, and 729; both branches return the correct SUPPORT and neither cites the annotated evidence. The full-restoration control matches the reference in every group.

The prompt determines which endpoint moves. In the 36 Gemma groups complete under both prompts, the eight baseline citation differences (all claim 268) disappear with explicit examples and nine appear on other claims, while verdict differences fall from six to zero. For Gemma on claim 431, the candidate state changes the verdict but not the reads or citations under the baseline prompt, and the reads and citations but not the verdict under the explicit-example prompt.

\subsection{Numerical branches can also change verdicts}\label{sec:3.3}

Claim 431 concerns reactive oxygen species, neuronal death, and FoxO3a activation. In six baseline Gemma groups, the read and citation sets remain the same while C changes an incorrect NEI to a correct SUPPORT verdict. Full restoration returns the incorrect reference verdict. Two baseline Qwen records similarly improve correctness for claim 431, while one claim-56 record becomes incorrect. The verdict changes observed under the baseline prompt do not recur under the explicit-example prompt.

\subsection{What restores the evidence path?}\label{sec:3.4}

Full restoration rebuilds the reference array exactly, so it is in effect a third reference run; under the explicit-example prompt, it matches H in all nine Qwen and 15 Gemma groups. Selected restoration and transplantation never reach their targets in these groups (Qwen 0/9; Gemma 0/12, three groups incomplete), and in baseline Qwen selected restoration recovers none of the 12 changed read sets. The nearest case is a one-score restoration for claim 431 that recovers citations and verdict in two repeats but not reads, with 95 row differences remaining; the reverse transplantation does not reproduce C's verdict (Figure~\ref{fig:4}b).

\begin{figure}[!tbp]
\centering\includegraphics[width=0.78\linewidth,height=0.78\textheight,keepaspectratio]{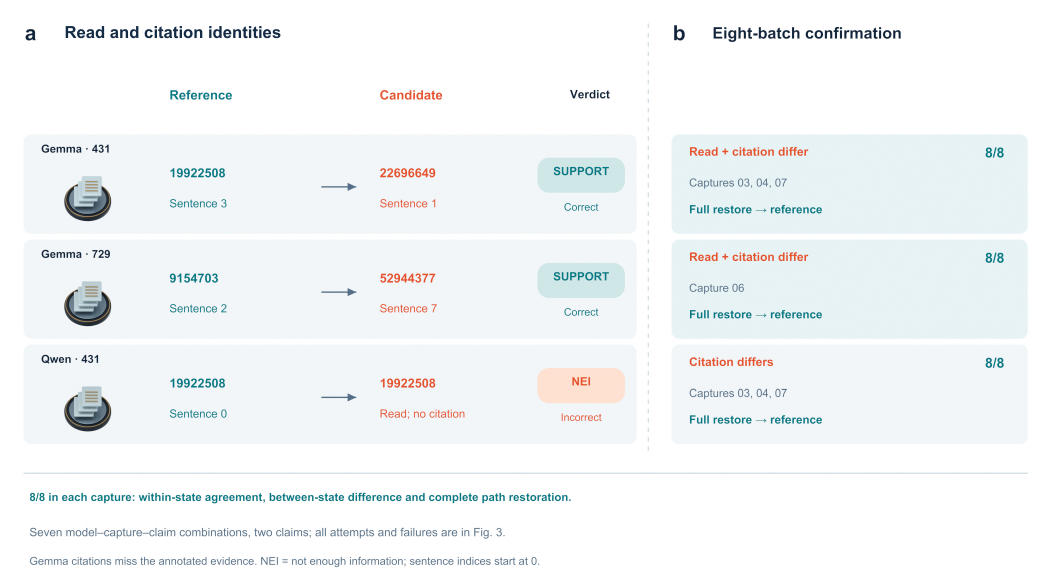}
\caption{Read and citation selections across repeated batches. (a) Actual document and sentence selections. Qwen reads the same document without citing under C; both NEI verdicts are wrong. Gemma gives correct verdicts while selecting different evidence, neither branch matching the annotated evidence. (b) Each marked combination has stable within-state endpoints, different H/C endpoints, and full restoration in all eight batches. Figure~\ref{fig:3} reports all attempts and incomplete groups.}\label{fig:2}
\end{figure}

\begin{figure}[!tbp]
\centering\includegraphics[width=0.78\linewidth,height=0.78\textheight,keepaspectratio]{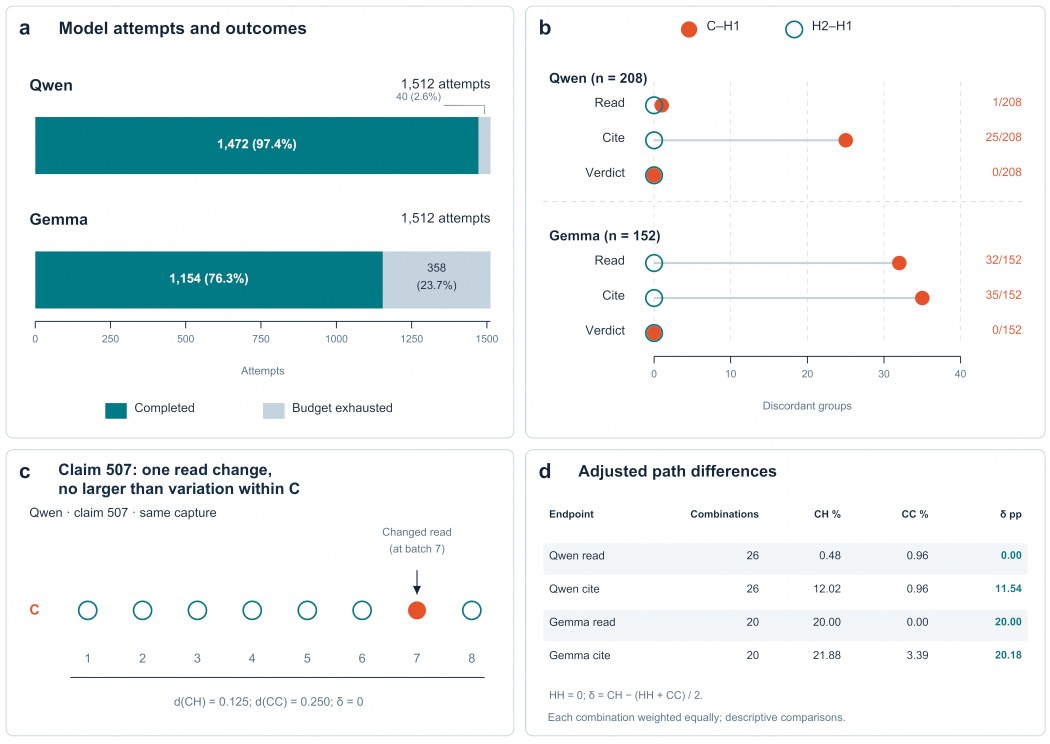}
\caption{All attempts and within-state variability. (a) Each model schedules 27 combinations \ensuremath{\times} eight batches \ensuremath{\times} seven conditions, or 1,512 episodes. (b) Qwen has 208 jointly completed groups and Gemma 152. Between candidate and H1, read, citation, and verdict discordances are 1, 25, and 0 for Qwen and 32, 35, and 0 for Gemma; H2 and full restoration never differ from H1. (c) Claim 507 differs only in the seventh batch; its adjusted read difference is zero. (d) Summaries weight each of 26 Qwen and 20 Gemma combinations equally. CH, HH, and CC are between-state and within-state discordance fractions; \(\delta=\mathrm{CH}-(\mathrm{HH}+\mathrm{CC})/2\), in percentage points (Appendix \ref{sec:J.14}).}\label{fig:3}
\end{figure}

Identical requests do not always give identical outcomes: some C and selected-restoration runs receive the same row and still differ (Appendix~\ref{sec:J.8}). We therefore ran all 27 combinations in eight further batches with the request seed fixed. Three Qwen combinations show the same citation change in every batch and four Gemma combinations the same read and citation change, while H1, H2, and full restoration agree and verdicts do not change (Figures~\ref{fig:2}--\ref{fig:3}). The seven combinations involve claims 431 and 729 only. A claim-507 read change appears in one batch and is no larger than the variation among candidate runs. No partial-row mask both restored the reference path and reproduced the candidate path across the eight batches.

A separate Qwen series includes three claim-831 records from two captures. Reads and citations change while the correct NEI verdict is preserved. Selected restoration recovers 2/3 reference paths and transplantation reproduces 3/3 candidate paths. The selected blocks replace all differences in the visited row, and the claim has no annotated relevant document. These records show full-row path recovery (Appendices F and \ref{sec:J.1}).

\subsection{Content masking and unselected controls}\label{sec:3.5}

Content masking removes titles and abstracts while retaining IDs and scores. Under the baseline prompt, all 243 masked attempts per model finish with uncited NEI. Among H1 pairs completed with both full and masked content, verdicts change in 57/81 Qwen and 28/40 Gemma comparisons. With explicit examples, Gemma completes only 29/243 masked attempts.

Unselected claims 350 and 1180 span both Cases and both query batches, with 576 attempts per model and prompt. Their completed comparison subsets show no read, citation, or verdict changes. Baseline Gemma's completed subset covers only claim 350. A Qwen cohort run under sustained batching, selected without requiring an upstream membership change, has 382 jointly completed, exposed groups out of 421 (Table~\ref{tab:2}). Its candidate--H1 discordance (3, 5, and 2 groups for reads, citations, and verdicts) is no higher than its H2--H1 discordance (4, 8, and 2), and every interval in Appendix F that weights captures equally includes zero. Without selection on an upstream change we see no excess over repeat-to-repeat variation.

\begin{table}[!tbp]
\caption{\textbf{Candidate and healthy-repeat discordance in the sustained-batching cohort.} The remaining 39 groups are incomplete or unexposed. Full restoration also has two verdict discordances.}\label{tab:2}
\centering\small\setlength{\tabcolsep}{3pt}
\begin{tabular}{@{}>{\raggedright\arraybackslash}p{\dimexpr 0.40000\linewidth-4.0pt\relax}>{\raggedright\arraybackslash}p{\dimexpr 0.32000\linewidth-4.0pt\relax}>{\raggedright\arraybackslash}p{\dimexpr 0.28000\linewidth-4.0pt\relax}@{}}
\toprule
\textbf{Endpoint} & \textbf{Candidate versus H1} & \textbf{H2 versus H1} \\
\midrule
Read membership & 3/382 & 4/382 \\
Cited sentences & 5/382 & 8/382 \\
Verdict & 2/382 & 2/382 \\
\bottomrule
\end{tabular}
\end{table}

\section{Numerical mechanisms behind choices and recovery}\label{sec:4}

\subsection{Score competition and evidence selection}\label{sec:4.1}

Tool outputs determine which items an agent can inspect. Four of five new SciFact captures change which documents are retrieved while preserving every query's relevance-label sequence, Recall@5, and nDCG@10 (Appendix~\ref{sec:C.3}; see also Appendix~\ref{sec:E.1}). PBMC shows similar neighbor changes with unchanged votes.

Competition can make an unchanged score consequential. For SciFact claim 684, one relevant document leaves the top five after three competitors change (capture in Appendix C.2). Restoring or transplanting any two competitor scores recovers or reproduces the Recall@5 loss. In the batch-8 capture identified in Appendix C.2, restoring a selected set of three scores recovers all changed memberships while 587/590 score differences and one order difference remain; no proper subset of that set recovers all reference memberships. A candidate deviation can also correct a reference error, as in splice classification (Appendix C.2). Over one fixed evaluation domain,

\[\Delta MSE = 2\operatorname{mean}(e\delta) + \operatorname{mean}\left( \delta^{2} \right).\]

For reference/candidate predictions \(\widehat y_i^H,\widehat y_i^C\) and targets \(y_i\), set \(e_i=\widehat y_i^H-y_i\), \(\delta_i=\widehat y_i^C-\widehat y_i^H\). Both means use the same evaluation rows; \(\Delta MSE=MSE_C-MSE_H\). Favorable alignment can reduce MSE even when selection changes. Differences between the ranking and evaluation domains, strict crossings, and ties are examined in Appendices B--C.

For a compound-activity model that scores each compound independently, 16 captures are split equally into development and confirmation subsets. Two priorities, decision sensitivity and co-occurrence, rank the row blocks using the development captures (Appendix C.1).

\textbf{Algorithm 2. Priority and increasing restoration budget.} Fix the block order before confirmation, then evaluate budgets 0, 1, 2, 4, 8, 16, 32, 64, and 127, recording recovery and remaining differences.

\begin{figure}[!tbp]
\centering\includegraphics[width=0.78\linewidth,height=0.78\textheight,keepaspectratio]{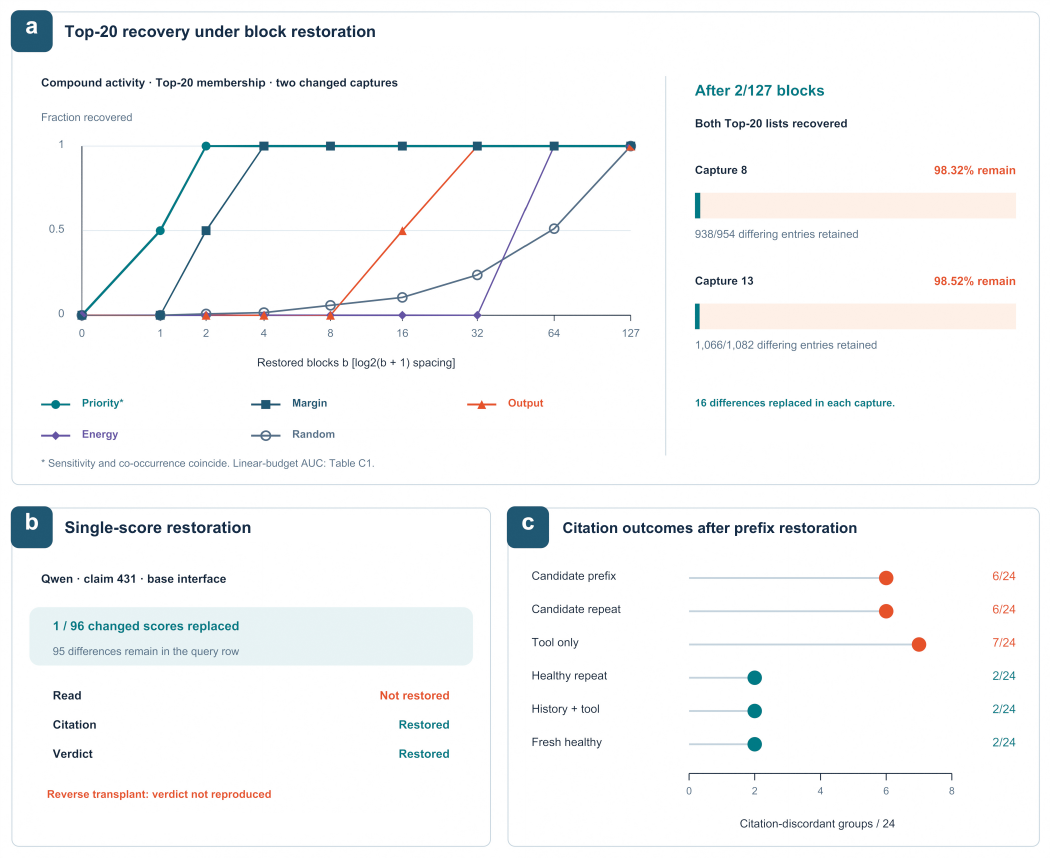}
\caption{Recovery under score and prefix interventions. (a) Recovery for the compound-activity model across nine tested budgets and two changed captures. Random priority averages 128 fixed orders. Horizontal spacing uses log\textsubscript{2}(b + 1), but the AUC in Table~\ref{tab:C1} is computed over b/127. Percentages of remaining differences refer to the two-block budget. (b) Two baseline Qwen repeats for claim 431: one score is restored and 95 differences remain. Citations and verdict recover, reads do not, and transplantation does not reproduce the candidate verdict. (c) Citation discordances after prefix interventions in 24 groups. Healthy repeat has 2/24 discordances.}\label{fig:4}
\end{figure}

The two confirmation captures with changed membership first recover at one and two blocks, respectively; both recover reference membership at the common two-block budget. On a linear budget axis, the recovery AUC is 0.992, compared with 0.498 for random priority (Figure~\ref{fig:4}a; Table~\ref{tab:C1}; Appendix C.1).

Recombining residuals across a panel of eight executions keeps each call's values but changes which differences occur together. Of 4,096 recombined executions, 29 produce lists absent from their source group. The BF16 score grid and tie rule explain all 1,024 enumerated combinations.

\subsection{Persistent state and the next agent choice}\label{sec:4.2}

Tool-state restoration changes subsequent outputs; earlier observations remain in the agent's visible history. Across 24 groups resumed from saved prefixes under seven conditions (scores differ in 13 of them), the candidate prefix and its repeat each have citation discordance in 6/24 groups, and tool-only replacement in 7/24. Healthy repeat, joint replacement of visible history and tool state, and a fresh healthy replay each have 2/24. Verdicts agree throughout (Appendix F.3).
In hERG assay allocation, replacing the prediction file after the agent had read faulty values recovered none of 39 changed decisions. Restoring the execution state saved before that read (visible history, tool variables and derived files) recovered all 39.

The saved AdamW state is \(\sigma_t=(\theta_t,m_t,v_t,\xi_t)\): parameters \(\theta_t\), first/second moments \(m_t,v_t\), and \(\xi_t\) containing per-parameter counters, hyperparameters, random state, and sample stream at update \(t\) (Appendix D.1) \citep{ref30}. After applying H/C gradients on healthy hardware, eight combinations \(\sigma_{abc} = \left( \theta^{a},m^{b},v^{c},\xi \right)\), \(a,b,c \in \{ H,C\}\), each recompute 63 subsequent updates. Initially only three parameters and seven entries of each moment differ, yet the path CCC \ensuremath{\rightarrow} HHC \ensuremath{\rightarrow} HHH produces Top-20 order discordance 0 \ensuremath{\rightarrow} 1 \ensuremath{\rightarrow} 0.

A real All-Reduce dilutes one rank's residual across 2, 4, or 8 ranks. With the subsequent computation fixed, \((J(1/2),J(1/4),J(1/8))=(0,1,0)\), where

\[J(\alpha)=\mathbf1\{P_{20}(F(U(\sigma_0,g_H+\alpha\delta_1)))\ne P_{20}(F(U(\sigma_0,g_H)))\}.\]

For \(N\in\{2,4,8\}\) ranks, \(\alpha=1/N\); \(g_H\) is the healthy reduced gradient and \(\delta_1\) the one-rank difference, before clipping. \(U\) clips and updates initial state \(\sigma_0\) once; \(F\) returns scores after the remaining updates. \(P_{20}\) returns the ordered Top-20 indices with fixed ties; \(\mathbf1\) indicates inequality. In this example the order changes only at the intermediate dilution, \(N=4\).

An eight-rank, two-node continuation has 63 bitwise-identical FP32 losses and synchronized replicas, but 30,411 parameters, 60,025 first moments, 105,921 second moments, and all 1,017 FP64 predictions differ from the reference. Membership and order remain equal in that condition. Across all 160 training continuations, membership is unchanged and order changes in 22 executions. When restoration is delayed by up to 32 updates, 30/32 partial restorations in the two affected captures recover membership, none recovers complete order or state, and all 16 full-state restorations recover their endpoints (Appendix D). 

\subsection{Sequence modeling and numerical propagation}\label{sec:4.3}

A four-state behavioral model predicts when and where the anomaly appears from the sequence of call shapes (Appendix E.3). Across 22 configurations it predicts events better than a static baseline (Brier score 0.090 versus 0.185) and matches magnitude profiles more closely (0.158 versus 0.218). The joint temporal criterion is not met: its composite temporal error is worse than that of a residual-sampling baseline (0.244 versus 0.144). Post-processing can conceal differences: all 2,048 attention-probability rows from the query, key, and value (QKV) projections are finite and normalized, but 58 change their maximum-attention position (Appendix~\ref{sec:E.2}). Clipping a gradient from norm approximately \(1.94\times10^{16}\) to one leaves cosine similarity to the healthy direction near \ensuremath{-}0.00259.

\subsection{Subsequent choices with biological tools}\label{sec:4.4}

PBMC, NCI, and ESOL agent runs test ordinary BF16 score rounding (Table~\ref{tab:3}). ESOL queries reveal offline labels, changing the fitted model and later selections (Appendices B, G, and H).

\begin{table}[!tbp]
\caption{\textbf{Subsequent choices under ordinary precision.} Counts use jointly completed groups.}\label{tab:3}
\centering\small\setlength{\tabcolsep}{4pt}\setlength{\parskip}{0pt}
\begin{tabular*}{\linewidth}{@{}>{\raggedright\arraybackslash}p{0.44\linewidth}@{\extracolsep{\fill}}c l@{}}
\toprule
\textbf{Tool and endpoint} & \textbf{After rounding} & \textbf{Healthy repeats / controls} \\
\midrule
PBMC: actions or marker selection & 9/30 & Repeat and zero: 0/30 each \\
NCI: final molecule membership & 11/61; 13/62 & Repeats: 0/61; 1/62 \\
ESOL: sampled membership & 12/24 each & Repeat and zero: no change \\
\bottomrule
\end{tabular*}
\par
\begin{minipage}{\linewidth}\small
NCI pairs list the healthy-reference, then matched-reference environment. ESOL (each environment): order changes in 15/24 seeds; test RMSE rises in 10, falls in 5, and is unchanged in 9.
\end{minipage}
\end{table}

\section{From tool states to scientific actions}\label{sec:actions}\label{sec:newinputs}\label{sec:herg}\label{sec:agentprov}

\subsection{Changed objects in screening and assay allocation}\label{sec:actions.objects}\label{sec:herg.task}\label{sec:herg.decisions}\label{sec:newinputs.agents}\label{sec:newinputs.donor}

A three-member Chemprop directed message-passing neural network (D-MPNN) ensemble \citep{ref60,ref61} predicted blockade of the hERG potassium channel, a cardiac-safety liability \citep{ref64}, for 2,009 internally held-out compounds. Using GPT-6.1 Sol, Biomni A1's standard (native) agent workflow \citep{ref45} performed two tasks: selecting the 402 highest-scoring compounds (Top-402) for measurement and assigning each of the 2,009 compounds to patch-clamp testing or advancement without testing (triage). A confirmation series comprised 16 saved BF16 passes of the cohort (sources), eight per Case, compared with a healthy BF16 reference from the same GPU model and host. An earlier retrospective series of 16 sources was used to develop the recovery rules. Three fresh trajectories evaluated each source--task pair; benchmark labels were withheld from the agent (Appendix~\ref{sec:I.1}).

\begin{figure}[!tbp]
\centering\includegraphics[width=\linewidth]{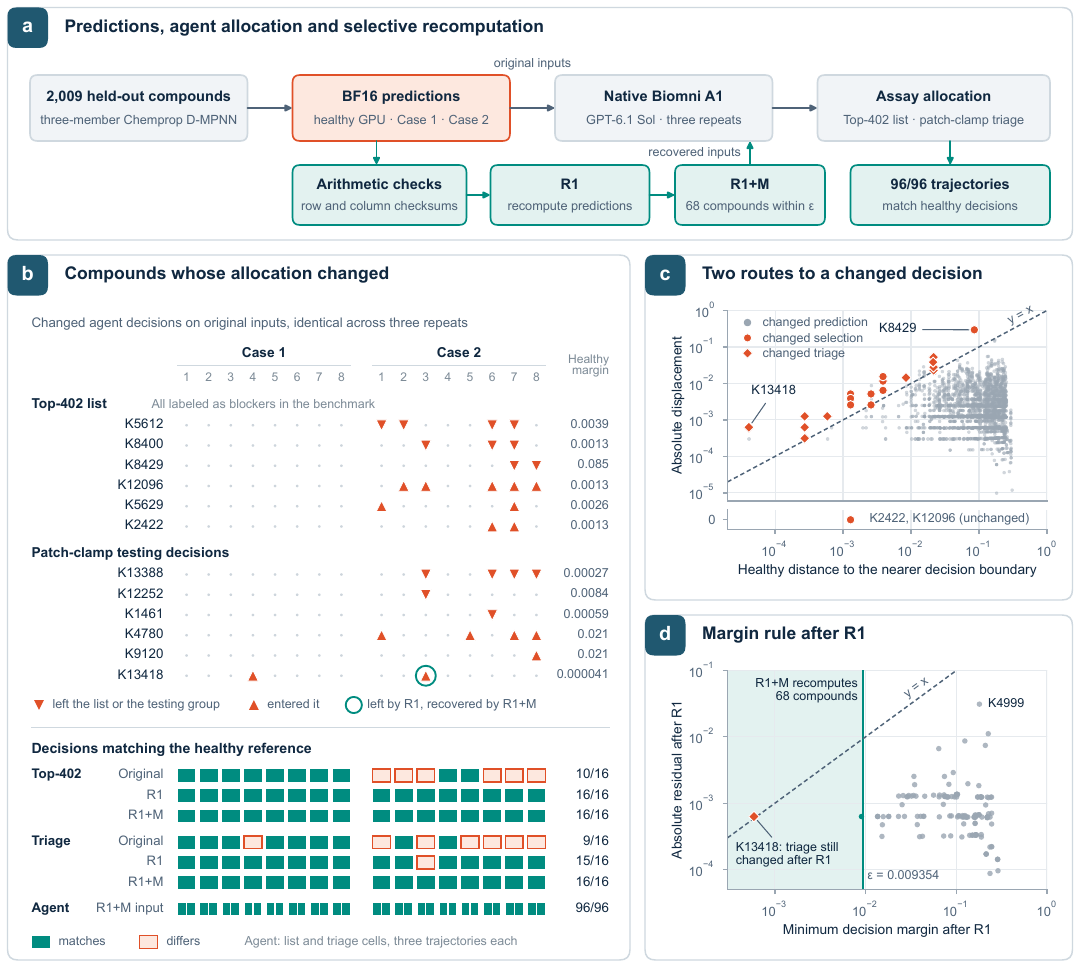}
\caption{Changed hERG assay allocations and recovery. (a) BF16 predictions feed native Biomni selection and triage; R1 recomputes predictions flagged by arithmetic checks, and R1+M adds a margin rule that also recomputes 68 compounds near decision boundaries. (b) Changed decisions across 16 sources; the ring marks the discrepancy left by R1 and recovered by R1+M. Healthy margin is distance to the affected decision's boundary; lower rows show recovery and the decisions of native Biomni runs on R1+M inputs. (c) Prediction displacement versus healthy distance to the nearer boundary; highlights mark changed decisions, and the lower strip shows entry with unchanged predictions. (d) R1 residual versus the minimum decision margin computed from R1 scores; shading marks R1+M selection. Dashed lines in (c,d) mark equality.}\label{fig:5}
\end{figure}

Case~1 preserved Top-402 membership and changed triage in one of eight sources; Case~2 changed each endpoint in six of eight. All nine list replacements exchanged labeled blockers, so the list still held 351 labeled blockers among its 402 compounds. Compound K5612, which left the list in four sources, is CHEMBL3804918, with a published hERG radioligand-displacement IC50 of 434~nM \citep{ref62,ref63}. K12252 and K1461, both labeled as blockers, were assigned to advancement without testing. Changes also reached beyond near-ties: K8429 fell from 0.880 to 0.584 and left the list, whereas K12096 and K2422 entered with unchanged predictions because competitors fell (Figure~\ref{fig:5}b,c; Appendix~\ref{sec:I.2}).

In screening 1,024 cells from held-out donors without an agent, each prespecified primary Case~2 capture created one false positive and corrected another error, preserving the total error count (Appendix~\ref{sec:xtask.2}).

\subsection{Restoration and selective recomputation}\label{sec:actions.recovery}\label{sec:herg.restore}\label{sec:herg.recompute}\label{sec:herg.agents}

Restoring paired reference values at their original coordinates recovered K5612 selection and K7714 triage with 11--254 restored coordinates while hundreds of cohort predictions still differed, consistent with \(\rho_D\le\rho_{\mathrm{state}}\) (Appendix~\ref{sec:I.3}).

For recovery without saved reference predictions, R1 recomputes on healthy hardware the predictions flagged by arithmetic checks, each for one compound from one ensemble member. In the retrospective series, integrity checks on stored data and checks that probabilities were finite and between 0 and 1 flagged none of the 16 faulty sources; row and column checksums with a rounding bound flagged all 16, with no false alarm in eight healthy verifier runs (Appendix~\ref{sec:I.4}). In the confirmation series, R1 recovered every changed decision except one added assay for K13418, the compound closest to the triage threshold.

R1+M adds a margin rule to R1, using only current R1 scores and \(\varepsilon=0.009354\), fixed at twice the largest retrospective R1 residual. Beyond the predictions R1 recomputed, it recomputed predictions from all three members for the 68 of 2,009 compounds with either decision margin below \(\varepsilon\). Selection and triage then matched the healthy reference in all 16 sources; recomputing the same number of compounds chosen at random or by disagreement among ensemble members left the discrepancy. Each recomputed prediction matched its reference value bit for bit; the other predictions kept their R1 values. Decision recovery left larger residuals far from the boundaries (Figure~\ref{fig:5}d; Appendix~\ref{sec:I.5}).

All 96 fresh native Biomni trajectories on R1+M inputs from the same 2,009-compound cohort reproduced the healthy decisions, with three per task for each source. Recovery preceded agent execution; paired runs on the original and R1 inputs, together with healthy controls, brought the total to 114 trajectories (Appendix~\ref{sec:I.6}).

\subsection{Agent behavior and certified commitment}\label{sec:actions.commit}\label{sec:agentprov.behavior}\label{sec:agentprov.commit}

All 36 Case~2 hERG trajectories whose inputs changed a decision followed the supplied decision rule (Appendix~\ref{sec:I.2}). In these tasks, the agents acted correctly on the values they were given, so agent competence did not prevent the changed decisions.

In separate molecular and cell screening tasks, Qwen3.8-27B chose two objects from 24 candidates whose predicted values lay nearest a task threshold; recorded measurements were revealed only when queried. Each context paired a screening input with a target: molecular solubility or a cell surface marker (CD3 or CD14). Forty contexts with repeatable behavioral differences between reference and candidate inputs were evaluated four times per condition. With the shortlist produced by reference computation held fixed, candidate numerical fields changed the set of objects chosen for measurement in 76 of 160 trajectories relative to reference values (Appendix~\ref{sec:xtask.1}).

\begin{samepage}
Execution provenance records the values used \citep{ref55,ref56}. A complete record preserves an incorrectly computed value as faithfully as a correct one. Trust in an action therefore needs an independent check of the computation behind it. Certification constrains which decisions may authorize action.

\end{samepage}

Fix a computation contract (inputs, model, numerical representation and execution settings), a candidate set and a tie rule, with reference scores \(p^\star\in\mathbb R^n\). Let \(g(q)\) denote the target decision computed from score vector \(q\), such as Top-402 membership or triage. After round \(r\), the admissible set \(\mathcal U_r\) fixes exactly recomputed scores and bounds the others. A decision \(a\) may authorize action only if
\[
g(q)=a\qquad\text{for every }q\in\mathcal U_r .
\]

\textbf{Proposition 1 (certified commitment).} \emph{Suppose \(p^\star\in\mathcal U_r\) for every round \(r\). If \(g\) takes the single value \(a\) on \(\mathcal U_r\), then \(a=g(p^\star)\). If, in addition, every round that ends without certification fixes at least one further score at its reference value, and fixed scores stay fixed, certification is reached within \(n\) rounds.} (Proof in Appendix~\ref{sec:commit.1}.)

Corollary~1 (Appendix~\ref{sec:commit.2}) gives sufficient conditions for certifying threshold and Top-\(K\) decisions under valid per-score error bounds: score intervals stay within their threshold decision regions, or every selected interval lies strictly above every unselected interval.

R1+M is an empirical recovery rule: it takes \(e\) as the largest absolute R1 residual in the retrospective sources and sets \(\varepsilon=2e\). A certified deployment uses valid per-score bounds or falls back to trusted full recomputation (Appendix~\ref{sec:commit.2}). Certification fixes the agent's input only when every returned value is determined by the certified target or certified separately (Appendix~\ref{sec:commit.3}).

\begin{samepage}
\textbf{Algorithm 3. Certified commitment.} Hold tool outputs during independent arithmetic verification. On trusted hardware, recompute flagged predictions and scores whose decisions remain unresolved under valid bounds; re-rank the full candidate set until the target and every returned value are certified. If the agent consumed a replaced output, restore its visible history, discard dependent state and pending actions, and invalidate server-side state built from the discarded history or on untrusted hardware, such as key--value caches. A certificate records the certified decision and supporting evidence. A commit service checks action agreement with the current certificate for the same request and contract, then commits at most once per request. If verification fails or the budget is exhausted, recompute every score on trusted hardware or commit nothing.

\end{samepage}

\subsection{Recovery and commitment for each request}\label{sec:actions.request}

In native Biomni, we implemented the fallback of Algorithm~3 that recomputes every score on trusted hardware. It restores the execution state when needed and uses a commit service that accepts only actions agreeing with the current certificate. The 16 confirmation fault sources and one shared healthy reference, each evaluated on two tasks with three repeats, gave 102 requests, each run under three recovery conditions (Appendix~\ref{sec:I.7}). Arithmetic verification flagged all 16 faulty sources and not the healthy reference. Certification used full trusted recomputation in all three recovery conditions. Recovery before the agent read the predictions gave the healthy decision in all 102 requests. After the agent had read the faulty predictions, replacing the prediction file recovered none of the 39 changed decisions, whereas restoring the execution state saved before that read and continuing with recomputed scores recovered all 39 and preserved the other 63. No wrong action was committed. For SciFact citations, joint restoration of history and tool state likewise returned discordance to the level observed in a healthy repeat.

On a healthy data-center GPU, a streaming implementation of the same checks raised a 0.47~s forward pass to 0.59~s, compared with 1.02~s for the original FP64 implementation and 0.88~s for double execution, which repeats the forward pass and compares outputs (Appendix~\ref{sec:I.4}). Across the 96 requests with faulty inputs, continuation from the restored state added about 29,000 tokens per request to the 44,000 already spent analyzing the faulty predictions; recovery before reading needed one analysis of about 44,000 tokens (Appendix~\ref{sec:I.7}).

\section{Related work}\label{sec:5}

ALCE measures citation quality alongside answer correctness \citep{ref57}. BioAgent Bench tests bioinformatics workflows under corrupted inputs and decoy files \citep{ref43}. ToolRobustBench locates failures across tool-calling stages; studies of manipulated stock data and unreliable tool returns track adoption of corrupted content \citep{ref51,ref52,ref53}. Our experiments start from measured GPU outputs and test whether restoring reference values at the original coordinates recovers an agent's reads and citations.

Causal tracing and activation patching intervene on internal representations to locate causal contributions \citep{ref17,ref18,ref19,ref38}. BioAgent-DF intervenes at saved scientific-tool states and measures both endpoint recovery and the fraction of encountered numerical differences replaced. Top-K resilience and decision-kernel analyses explain how score errors affect selection \citep{ref20,ref21}; here the continuation includes a language-model agent.

Production-cluster and LLM-training studies characterize SDC, detect faulty computation, and measure its training effects \citep{ref1,ref2,ref3,ref4,ref5,ref6,ref31,ref32}. We follow captured errors into literature retrieval and subsequent evidence selection. Autonomous research systems, including Biomni and BioDiscoveryAgent, execute tools and experimental feedback loops \citep{ref39,ref45,ref46}. OmniScientist records execution provenance and numerical traceability, and verification-first catalysis calls for auditable evidence \citep{ref55,ref56}. Our replay and restoration experiments test how a change at a recorded numerical site alters that downstream evidence.

Algorithm-based fault tolerance encodes checksums into matrix operations to detect computational errors \citep{ref66}, and selective classification abstains on low-confidence predictions \citep{ref67}. Predictive multiplicity, underspecification and prediction churn describe models or model versions that agree in aggregate but differ on individual predictions \citep{ref68,ref69,ref70}. Here the predictor is fixed, errors arise during execution, and decision margins direct trusted recomputation within a provenance audit.

\section{Discussion and conclusion}\label{sec:6}

The repeatedly confirmed SciFact result is a separation between verdict and evidence path. Numerical diagnostics separate arithmetic anomalies from the ordinary precision effects that also change membership between healthy references. Original-coordinate interventions test which value differences matter: full restoration returned the reference reads and citations in every confirmed case, while no tested partial-row mask both restored the reference path and reproduced the candidate path. For claim 831, the mask selects 8 of 38 global row blocks but replaces every changed score in the visited row, making it a full-row replacement despite its global sparsity.

The repeatedly confirmed evidence-path changes span two claims and two 4B models in Case~2 captures selected for upstream retrieval changes; without that selection, candidate values cause no more discordance than a healthy repeat (Table~\ref{tab:2}). Because the eight batches repeat identical temperature-zero requests, the independent units are captures and claims, compared over jointly completed groups. Stable verdicts can conceal changed evidence paths, so audits should record what an agent reads and cites alongside its verdict.

In hERG assay allocation, decision changes reflected score displacements relative to decision boundaries, including movement of the selection boundary when competing scores changed.

When a research agent directs experiments in an automated laboratory \citep{ref71,ref72}, a numerical fault that changes its decision changes the experiment performed. Aggregate measures need not reveal such changes: in hERG assay allocation, the shortlist kept the same number of labeled blockers while the compounds assigned to testing changed. Verification therefore belongs where a decision is committed: flagged and unresolved scores are recomputed on trusted hardware, an agent that has used superseded values resumes from the execution state saved before it read them, and only certified decisions are committed. In our experiments, the check preceded the commitment of an assay allocation; in an automated laboratory, it would precede the instructions sent to instruments.

We propose extending this framework to decision recovery under a shared budget for computation, agent execution and experimental measurements. A recovery agent would choose which calculations to verify or repeat, which dependent execution state to restore and which measurements to acquire, aiming to resolve as many pending decisions as possible. Valid error bounds or trusted recomputation would support certification of decision agreement with a specified reference computation; new measurements would provide evidence for resolving the underlying biological question under prespecified experimental criteria.

A benchmark with evaluation references hidden from the agent would separately score fault detection, acquisition of trusted evidence, recovery of reference decisions, evidence-based revision and correct commitment. Prospective studies in automated laboratories, with independent measurements and replication across sites, would measure missed or spurious findings, experimental resource use and recovery cost. In that setting, scientific decision fidelity concerns which experiments are performed, not only which conclusions are reported.

\section*{Acknowledgments}

Computations were performed on INTOSYN AI Labs infrastructure.

\section*{Code availability}

Aritheia, a public tool for detecting numerical anomalies on NVIDIA GPUs and collecting diagnostic records, is available at \url{https://github.com/INTOSYN/Aritheia_GPU_Test}. Each GPU runs and evaluates its checks independently.

\label{main-text-end}

\clearpage
\label{references-start}
\clearpage
\appendix
\raggedbottom
\numberwithin{figure}{section}
\numberwithin{table}{section}
\renewcommand{\thefigure}{\thesection\arabic{figure}}
\renewcommand{\thetable}{\thesection\arabic{table}}
The appendices provide numerical interfaces, agent protocols, precision controls, and intervention results. Table~\ref{tab:D1-suite01} represents all 160 continuation endpoints with identical repeats and zero-residual controls consolidated.

\FloatBarrier
\section{Numerical provenance and reference comparisons}\label{app:A}

\subsection{Platforms and experimental units}\label{sec:A.1}

The RTX platform supports native diagnostics, task execution, numerical capture, and healthy reference comparisons. An eight-GPU NVLink/NVSwitch platform supports healthy continuation, high-precision evaluation, and distributed selection. Cross-node layouts combine one healthy RTX GPU with one, three, or seven healthy data-center GPUs.

\textbf{Case 1 (size-dependent)} and \textbf{Case 2 (expanded-trigger)} are two classes of faulty GPUs, distinguished by the fault-activation pattern observed during cluster operation. Case 1 GPUs fail more often on the tested large-matrix paths than on the corresponding short probes. On Case 2 GPUs those short probes also produce arithmetic violations.

All reported capture experiments used the manufacturer-default hardware configuration and default power limits. Matched reference executions used the corresponding inputs, program, and declared numerical conditions; independently healthy executions supplied additional controls. References were measured before, during, and after each capture round.

\subsection{Exact probes and execution-path controls}\label{sec:A.2}

Fault-injection and vulnerability methods control specified error forms and locations \citep{ref7,ref8,ref9,ref10,ref11,ref12,ref13,ref14,ref15,ref16,ref37}. The consequence experiments use values observed during accelerator execution.

\begin{figure}[H]
\centering\includegraphics[width=\linewidth,height=0.78\textheight,keepaspectratio]{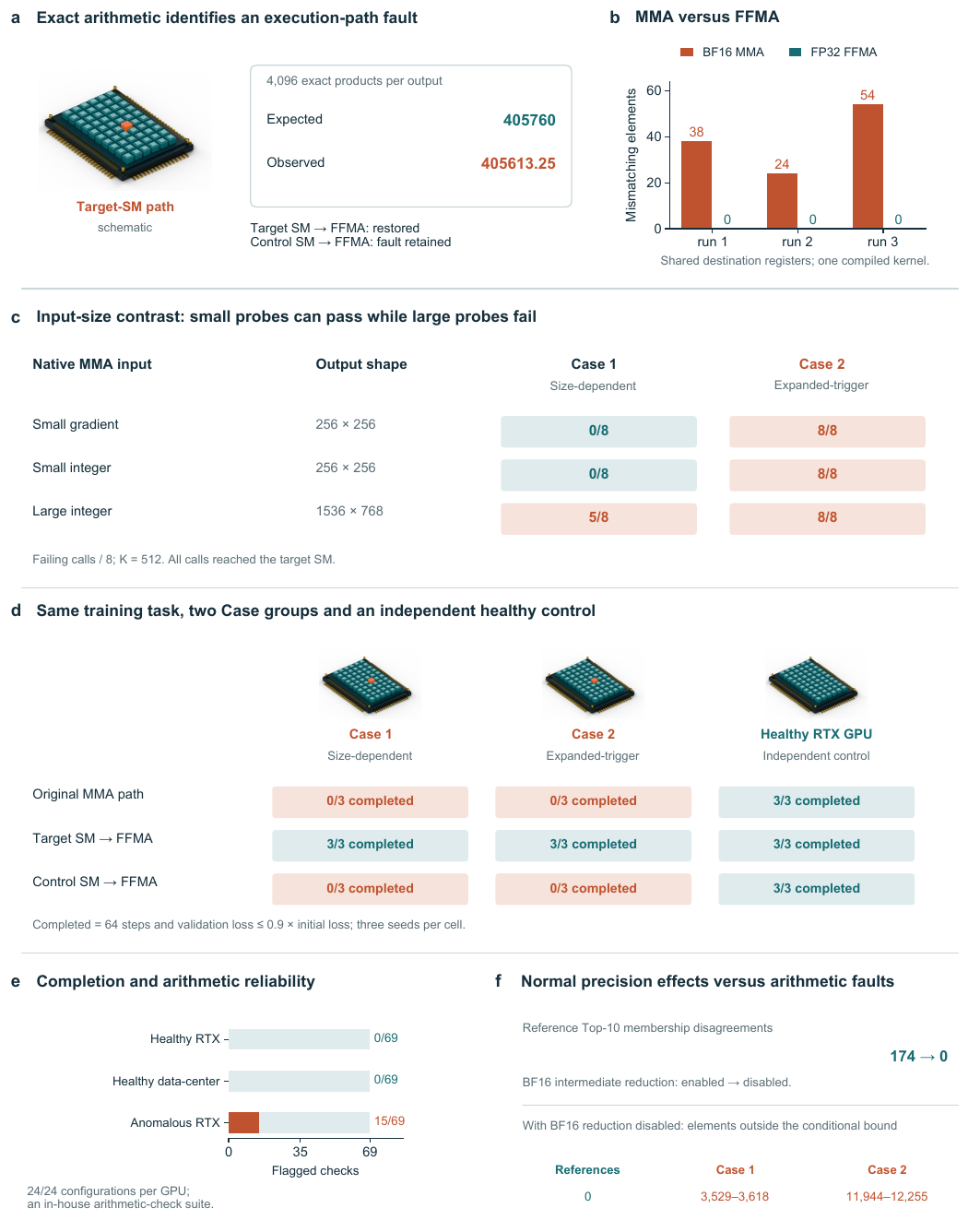}
\caption{Arithmetic execution-path controls. The panels show exact-input, paired-instruction, matrix-shape, training, diagnostic, and precision-path comparisons. Case 1 and Case 2 label the two classes of faulty GPUs. Matrix-call fractions count failures among eight calls; training cells count completions among three seeds.}\label{fig:A1}
\end{figure}

The positive-integer probe keeps each represented product and every correct FP32 partial sum exact under all permitted accumulation orders. The 4,096-term accumulation therefore has zero allowed error: its expected value is 405760, whereas the observed result is 405613.25 (Figure~\ref{fig:A1}a). The fractional part alone contradicts the exact-integer construction. Three comparisons with the same kernel and destination registers produce 38, 24, and 54 MMA mismatches and no FFMA mismatches. Moving the destination group from R4--R7 to R8--R11 or R12--R15 moves the anomalous slot from R5 to R9 or R13, respectively.

With proportional rows or columns, such as [1, 2] and [2, 4], a single fault can appear as proportional changes in several outputs, so the effect of input structure is hard to separate from that of the execution path. A further experiment therefore uses matrices in which no row is a multiple of another row and no column is a multiple of another column; in its core test group, both matrices were also verified to have full rank. The full experiment, including these inputs and other controls, comprises 189 executions: 103/108 candidate calls fail, and all 81 controls pass. All 6,150 anomalous elements are finite, and 650 thread blocks show whole-row errors.

In the Case 1 size comparison, 5/8 large-integer calls fail, while small-gradient and small-integer native calls each fail in 0/8. Another 288 small calls reach the target SM and satisfy their bounds. In Case 2, all three shapes fail in 8/8 calls. The distinction between short and long calls applies only to the tested shapes: a small application batch can still contain large internal matrices.

For each GPU class, a small Transformer completes 0/3 native and 0/3 non-target sham runs; target-SM FFMA substitution completes 3/3. Independent healthy and matched-reference runs provide the remaining comparisons in the full 36-task design. Completion requires 64 steps and independently evaluated loss at most 0.9 times its initial value. One Case 1 trajectory records an operator anomaly at step 1 and first triggers the ordinary NaN/Inf monitor at step 47; it does not complete.

\subsection{Conditional arithmetic bounds and normal implementation differences}\label{sec:A.3}

For represented operand pairs \((a_k,b_k)\), \(k=1,\ldots,K\), with \(K\in\mathbb N_{>0}\), let \(B\geq\sum_{k=1}^K|a_kb_k|\) be an outward bound. Following \citep{ref33,ref34}, the numerical comparison uses

\begin{gather*}
G = \left( 1 + u_{p} \right)\left( 1 + u_{a} \right)^{K + 2}\left( 1 + u_{r} \right)^{\nu_K},\\
E = (G - 1)B + G\{ K\eta_{p} + (K + 2)\eta_{a} + \nu_K\eta_{r}\}.
\end{gather*}

For rounding type \(t\in\{p,a,r\}\), the model is \(|\operatorname{fl}_t(x)-x|\leq u_t|x|+\eta_t\): \(p\) denotes product rounding, \(a\) accumulation, and \(r\) additional intermediate-format rounding. The nonnegative \(u_t\) are dimensionless relative bounds; \(\eta_t\geq0\) are absolute allowances in the operation's output units. Here \(K+2\) bounds ordinary accumulation-rounding operations and \(\nu_K\in\mathbb N_0\) bounds additional intermediate-rounding nodes. The reference interval includes \(E\) and final storage conversion. Its applicability requires the stated computation graph, overflow behavior, and treatment of subnormal values.

PyTorch documents intermediate reduced-precision reduction and platform-dependent floating-point results \citep{ref40,ref41}; cuBLAS specifies reproducibility conditions involving toolkit version, architecture, SM count, and concurrent execution \citep{ref42}. An ordinary autocast calculation does not inherit the exact-integer probe's zero tolerance.

The PBMC diagnosis fixes a 2,638 \ensuremath{\times} 1,838 input and includes 225 computations without profiling. With default BF16 intermediate reduction, the independent healthy RTX result and matched reference differ in Top-10 membership for 174 queries. Disabling only that low-precision reduction removes the membership disagreement, leaving two order differences. The data-center reference and matched reference agree under both BF16 reduction settings.

A post-hoc segmented-rounding model uses eight segments of nominal length 232 or four of nominal length 464, with the last segment truncated to the remaining 214 or 446 coordinates, respectively. It reconstructs 2,915,240 of 2,917,772 nonzero cross-reference differences (99.9132\%). Let \(\mathcal I_1,\ldots,\mathcal I_J\) partition the \(K=1838\) feature coordinates, with \(K_j=|\mathcal I_j|\); \(q_B,d_B\) are the BF16-represented query and reference vectors. The relative accumulation contribution for segment \(j\) is

\[\gamma_{2K_j+4}(u_{32})\sum_{k\in\mathcal I_j}|q_{B,k}d_{B,k}|,\qquad \gamma_n(u)=\frac{nu}{1-nu},\qquad u_{32}=2^{-23}.\]

Here \(nu<1\); \(u_{32}\) is the conservative bound used in the model. The interval also propagates the FP64 reference radius, absolute allowances, intermediate BF16 conversion, cross-segment accumulation, and final conversion. Omitting intermediate BF16 rounding excludes legitimate healthy values in the reported comparison, supporting its inclusion in the numerical model.

Relative to the stated strict-BF16 interval model, Case 1 has 3,529--3,618 out-of-bound elements and Case 2 has 11,944--12,255; healthy, matched, and repeated reference controls have zero. Deterministic settings do not eliminate these anomalies. Retaining BF16 inputs while computing in FP32 produces identical results across comparison groups. FP16 errors occur on some tested paths; the FFMA, TensorFloat-32 (TF32), integer multiply-add (IMAD), and INT8 controls show no anomaly. INT4 and bit matrix multiply-accumulate (BMMA) were not fully evaluated.

\subsection{Diagnostic completion and arithmetic validation}\label{sec:A.4}

The diagnostic evaluation ran an in-house arithmetic-check suite on three GPUs: a faulty RTX GPU that produced anomalous results, a healthy RTX GPU, and a healthy data-center GPU. Each GPU completes all 24 application configurations. On the faulty GPU, 15 of 69 arithmetic checks detect an anomaly; none do on the two healthy GPUs.

\FloatBarrier
\section{Scientific tool interfaces}\label{app:B}

\subsection{Tasks and computational objects}\label{sec:B.1}

\textbf{Original-coordinate residual reconstruction.} Residual implementations with storage conversion \(Q\) are

\[{\widetilde{R}}_{S} = Q\left( FP32(c) - \mathbf{1}_{S} \odot r \right),\quad\quad{\widetilde{T}}_{S} = Q\left( FP32(h) + \mathbf{1}_{S} \odot r \right).\]

Here \((\mathbf1_S)_i=1\) for \(i\in S\) and zero otherwise, and \(\odot\) denotes elementwise multiplication. We verify reconstruction after conversion using empty, full, and zero-residual controls. This check is needed because floating-point subtraction followed by addition can fail to recover the original value. Replay keeps the capture's representation, coordinates, and computational site.

\begin{figure}[H]
\centering\includegraphics[width=\linewidth,height=0.78\textheight,keepaspectratio]{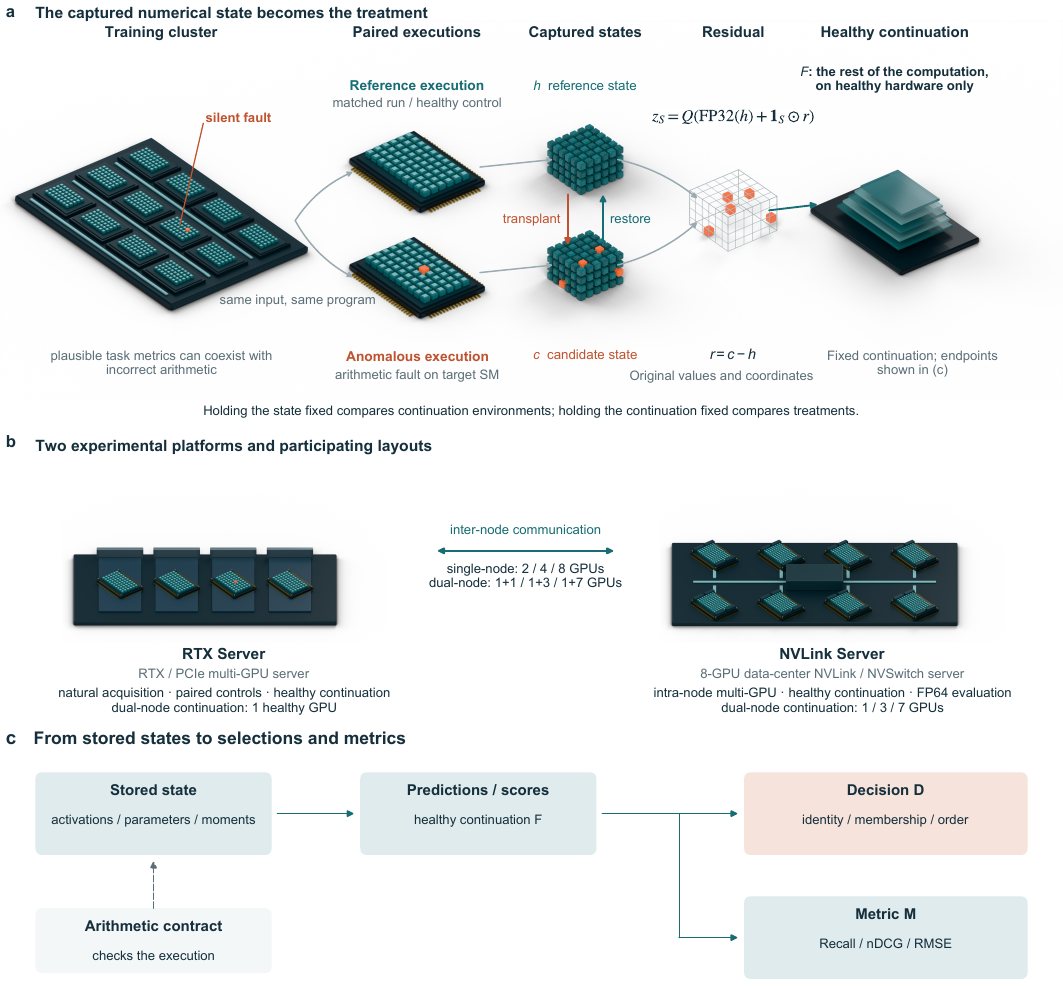}
\caption{Tool-state interventions and computing platforms. (a) Paired states supply original-coordinate restoration and transplantation on healthy hardware. (b) The PCIe and NVLink/NVSwitch layouts indicate computing roles and participation. (c) Arithmetic checks, stored states, selection identities, and task metrics. Table~\ref{tab:B1} lists the twelve interfaces.}\label{fig:B1}
\end{figure}

The interface audit (Table~\ref{tab:B1}; Figure~\ref{fig:B1}) includes all 492 observed outputs: 275 candidates, with 145 in Case 1 and 130 in Case 2, and 217 references or controls. Superconductivity supplies a materials-science comparison.

\begingroup\small\setlength{\tabcolsep}{3pt}
\begin{longtable}{@{}>{\raggedright\arraybackslash}p{0.27\linewidth}>{\raggedright\arraybackslash}p{0.25\linewidth}>{\raggedright\arraybackslash}p{0.43\linewidth}@{}}
\caption{\textbf{Scientific interfaces and input sizes.} Splits count dataset rows.}\label{tab:B1}\\
\toprule
\textbf{Task and data} & \textbf{Size / split} & \textbf{Computation and measured scope} \\
\midrule\endfirsthead
\multicolumn{3}{l}{\small Table~\ref{tab:B1} (continued)}\\
\toprule
\textbf{Task and data} & \textbf{Size / split} & \textbf{Computation and measured scope} \\
\midrule\endhead
\bottomrule\endfoot
RDKit FreeWilson\newline CHEMBL2321810\newline activity example & 1,017; 719/140/158 & 2,048-bit Morgan fingerprints; 2048 \ensuremath{\rightarrow} 512 \ensuremath{\rightarrow} 256 \ensuremath{\rightarrow} 1 MLP; full-candidate Top-10. Top-20 protocols: Appendices C--D. \\
Continuation of the compound checkpoint & 1,017; 719/140/158 & 64 AdamW updates; parameters, moments, predictions, and ranking. \\
RDKit NCI molecular neighbors & 4,999 raw; 4,991 valid; 1,536 queries & Tanimoto similarity, self-match exclusion, fixed Top-10. \\
WDBC morphology features & 569 \ensuremath{\times} 30; 401/83/85 & 30 \ensuremath{\rightarrow} 512 \ensuremath{\rightarrow} 256 \ensuremath{\rightarrow} 2 feature classifier. \\
scikit-learn optical digits & 1,797 \ensuremath{\times} 64; 1271/247/279 & 256-dimensional cached features and a healthy classification head. \\
Processed PBMC & 2,638 \ensuremath{\times} 1,838 & Normalized-dot-product Top-10; five-neighbor voting with computational annotations. \\
BEIR SciFact & 5,183 abstracts; 300 queries; 512-dimensional representations & Fixed latent-semantic retrieval, evaluated independently of agent verdicts. \\
UCI splice & 3,190 sequences of 60 bases; 2232/496/462 & 240-dimensional encoding and three-class prediction. \\
BreastMNIST & 780 images of 28 \ensuremath{\times} 28; 546/78/156 & Binary classification. \\
UCI superconductivity & 21,263 \ensuremath{\times} 81; 14628/3293/3342 & Critical-temperature prediction and ranking. \\
Author-created table questions & 32 four-option questions & Qwen3.5-4B next-token option scoring. \\
Author-created tool-selection questions & 32 intent questions & Tool-name selection; the selected tools are not executed. \\
\end{longtable}\endgroup

The compound-activity task aligns the Act target to Name. Full-candidate ranking includes training rows, while quality evaluation uses the fixed test split. Standardization is fitted on training rows only. Grouping uses canonical structures, complete sequences, or formula strings as appropriate; formula-string grouping does not perform crystal-structure deduplication.

SciFact uses a fixed lexical latent-semantic retriever to isolate numerical scoring from representation construction. Saving the representations provides identical inputs and original query--document coordinates for paired scoring and interventions. CPU tokenization constructs unigrams and bigrams; TF-IDF weighting and randomized SVD run in FP64 on a healthy data-center GPU. Vocabulary selection, IDF, and SVD are fitted only to the document corpus. TF-IDF uses \texttt{min\_df=2}, at most 20,000 terms, log term frequency, smoothed IDF, and L2 normalization. Randomized SVD uses rank parameter 522, five iterations, and seed 20260910, retaining 512 dimensions. Document representations use the retained left singular vectors scaled by their singular values; query TF-IDF vectors are projected onto the same retained right-singular-vector basis. Both representations are then row-normalized and saved in FP32. Matching uses the inner product of these normalized latent representations. Documents follow a fixed lexicographic ID order.

In the earlier free-search agent interface, later searches use a healthy TF-IDF implementation. The fixed-query confirmation binds its single search to the original claim row and imported score array.

\subsection{Neighbor and document identity under unchanged metrics}\label{sec:B.2}

In PBMC, 8/10 Case 1 captures and 4/10 Case 2 captures change neighbor membership while all 2,638 votes remain identical within each capture. Across these twelve captures, 118 Top-10 run--query membership events comprise 78 affecting only ranks 6--10, two changing top-five order only, 17 replacing voters while preserving the label sequence, seven changing label order while preserving counts, and 14 changing label counts while preserving the final vote. Thus 38 events replace actual voting members.

A further 20 events change top-five membership solely through Top-10 reordering. The total of 58 top-five membership changes therefore covers a different event set from the 118 Top-10 membership events. In 11/12 captures, an actual voter changes somewhere while all final votes remain equal.

Three observed Case 2 SciFact outputs change retrieved membership with identical per-query relevance-label sequences, Recall@5, and nDCG@10. The corresponding Case 1 retrieval outputs preserve membership and order.

\subsection{Language-model option scoring and tool execution}\label{sec:B.3}

The table-question and tool-selection tasks use Qwen3.5-4B four-option scores without free generation, key--value (KV) caching, or tool execution. Every initial complete candidate run contains at least one question with NaN or Inf. Selection and correctness comparisons use only questions whose four option scores are finite.

For Case 1, tool prompts padded to 300 tokens and table prompts padded to 300 or 350 tokens each produce 24 finite-score executions. Every execution changes option scores while preserving the chosen option. With original 152--158-token tool prompts, 22/24 executions have finite option scores and 16 of those change the selected option; the other two contain numerical alerts. Padding changes the propagation of numerical differences into option selection.

\FloatBarrier
\section{Selection and recovery}\label{app:C}

\subsection{Recovery cost under a fixed background}\label{sec:C.1}

Recovery for the compound-activity model uses eight development layer-0 captures (0--7) and eight confirmation captures (8--15). The 127 eligible row blocks, indexed 0--126, contain eight rows each. The final singleton at index 1016 remains healthy and is excluded from the restoration budget. Budgets count complete blocks. Both development priorities select block 105 first and block 104 second.

Here \(h,s^{(q)}\in\mathbb R^{1017}\) are reference and development-capture final-score vectors, indexed \(0,\ldots,1016\), and \(B_j=\{8j,\ldots,8j+7\}\) for \(j=0,\ldots,126\). The set \(M_{20}(x)\) contains the 20 highest-scoring indices, breaking ties by increasing original index. The vector \(h[j\leftarrow s]\) equals \(s\) on \(B_j\) and \(h\) elsewhere; \(s[j\leftarrow h]\) reverses this replacement. For a fixed deterministic row-wise continuation, replacing complete saved-state rows in \(B_j\) replaces only their corresponding final scores; the row-block intervention and score substitution agree. With \(\operatorname{mean}_q f_q=\frac18\sum_{q=0}^7 f_q\), the development priorities are

\begin{gather*}
d(s,h) = 20 - \left| M_{20}(s) \cap M_{20}(h) \right|,\\
U_{j} = \operatorname{mean}_{q}d\left( h\left\lbrack j \leftarrow s^{(q)} \right\rbrack,h \right),\\
P_{j} = \operatorname{mean}_{q}\max_{i \in B_{j}}\frac{\left| s_{i}^{(q)} - h_{i} \right|}{\left| h_{i} - t_{20} \right| + \epsilon_{64}},\\
G_{j} = \operatorname{mean}_{q}\{ d\left( s^{(q)},h \right) - d\left( s^{(q)}\lbrack j \leftarrow h\rbrack,h \right)\}.
\end{gather*}

The reference cutoff \(t_{20}=(h_{(20)}+h_{(21)})/2\) uses descending reference scores \(h_{(k)}\). The positive denominator stabilizer is \(\epsilon_{64}=\operatorname{finfo}(\mathrm{float64}).\mathrm{eps}=2^{-52}\), on the score scale. Descending orders \(\left( U_{j},P_{j}, - j \right)\) and \(\left( G_{j},U_{j} + 10^{- 9}P_{j}, - j \right)\) give the same complete priority. The priorities based on current output displacement and hidden difference energy use each candidate's own H/C comparison.

\begingroup\small\setlength{\tabcolsep}{3pt}
\begin{longtable}{@{}>{\raggedright\arraybackslash}p{\dimexpr 0.33000\linewidth-4.0pt\relax}>{\raggedright\arraybackslash}p{\dimexpr 0.24000\linewidth-4.0pt\relax}>{\raggedright\arraybackslash}p{\dimexpr 0.43000\linewidth-4.0pt\relax}@{}}
\caption{\textbf{Recovery over the two confirmation captures with changed membership.} Budgets are 0, 1, 2, 4, 8, 16, 32, 64, and 127 blocks.}\label{tab:C1}\\
\toprule
\textbf{Priority} & \textbf{AUC, linear budget axis} & \textbf{First tested budget recovering both lists} \\
\midrule\endfirsthead
\multicolumn{3}{l}{\small Table~\ref{tab:C1} (continued)}\\
\toprule
\textbf{Priority} & \textbf{AUC, linear budget axis} & \textbf{First tested budget recovering both lists} \\
\midrule\endhead
\bottomrule\endfoot
Single-block decision sensitivity & 0.992 & 2 \\
Co-occurrence & 0.992 & 2 \\
Healthy boundary margin & 0.982 & 4 \\
Current output displacement & 0.858 & 32 \\
Hidden difference energy & 0.622 & 64 \\
Mean of 128 preset random orders & 0.498 & All orders recover at 127 \\
\end{longtable}\endgroup

AUC in Table~\ref{tab:C1} is the trapezoidal summary on linear budget \(b/127\); Figure~\ref{fig:4}a uses \(\log_2(b+1)\) spacing. Capture 8 first recovers reference membership at one block and capture 13 at two. Unequal entries count scalar coordinates that differ between the saved candidate and reference layer-0 activation tensors within the 127 eligible row blocks. At the common two-block budget, the selected blocks contain 16 such entries in each of captures 8 and 13, retaining 938/954 (98.32\%) and 1066/1082 (98.52\%), respectively. Across all eight confirmation captures, 92 of 7,608 unequal entries are replaced, leaving 7,516/7,608 (98.79\%). Full blocks are read and replaced, so memory-access cost follows block size.

\subsection{Score competition}\label{sec:C.2}

In one molecular-ranking example, restoring a single score recovers the complete reference Top-10 order while 55 scores remain different. Reverse transplantation reproduces the changed order. In SciFact capture \texttt{97ab63a85e85} (query batch 8), restoring a selected set of three final-score coordinates recovers all changed memberships while retaining 587/590 score differences and one order difference. Every proper subset of this set fails to recover all reference memberships, so the selected set is minimal under inclusion.

Single-cell interventions use 20 captures, six masks, two directions, and two repeats: 480 executions over 240 conditions. Full restoration and full transplantation reconstruct their respective endpoints.

For claim 684 in SciFact capture \texttt{019f688f6ea3} (configured batch-size limit 1,536; actual query batch 300), relevant document 4942718 moves from rank four to seven after three competitors change, while its own score stays unchanged. Restoring or transplanting any two of three competitor scores recovers or reproduces the Recall@5 loss. In splice classification, either of two changed logits reproduces the correction of a reference error.

Residual recombination preserves each call position's values, multiplicities, and total energy across a panel of eight executions. Two source groups, four block lengths, 64 panels per group/length, and eight executions per panel yield 4,096 constructed executions. Only call-wise recombination, block length one, produces lists absent from the original source group: 24/512 early and 5/512 late. Block lengths 4, 16, and 32 each produce zero such lists.

\begin{figure}[H]
\centering\includegraphics[width=\linewidth,height=0.78\textheight,keepaspectratio]{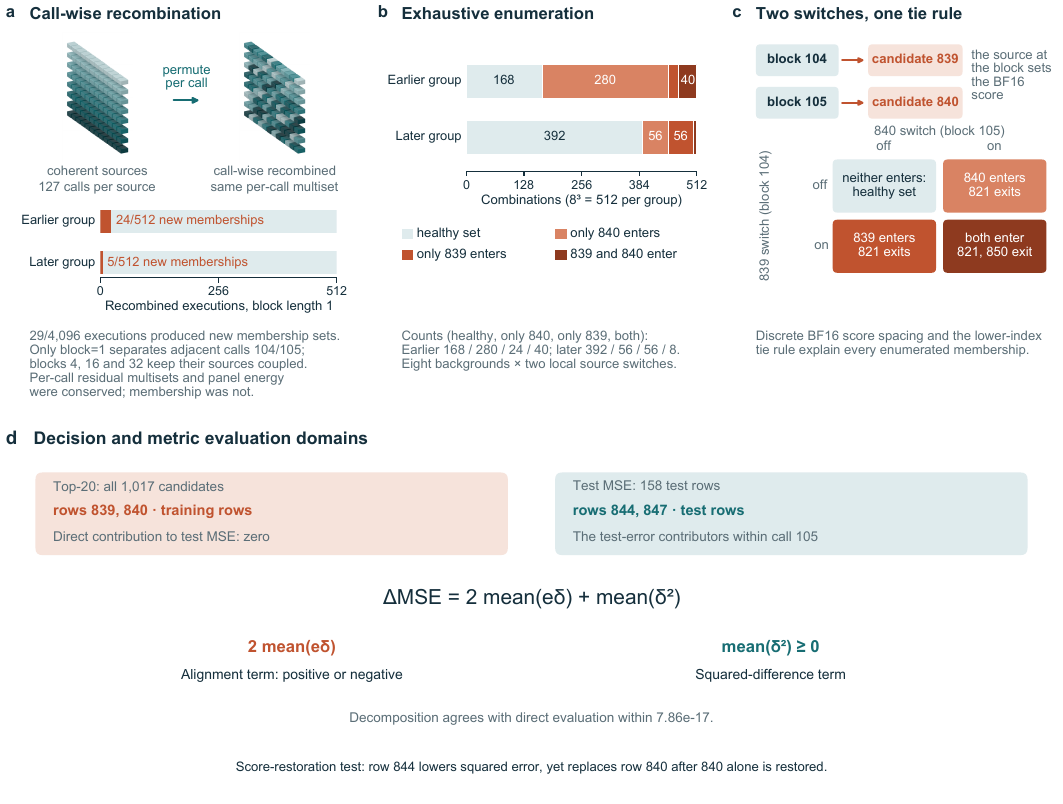}
\caption{Recombination and competition. Finite combinations of captured residuals explain candidate membership and its relation to prediction error. The MSE panel shows the cross term and squared-difference term.}\label{fig:C1}
\end{figure}

The 1,024 enumerated combinations produce four membership states (Figure~\ref{fig:C1}b). Counts are 168/280/24/40 in the early group and 392/56/56/8 in the late group. Blocks 104 and 105 control rows 839 and 840. The fixed block origin keeps those adjacent sources together for the longer partitions. BF16 score spacing and the rule that breaks ties by lower original index explain the outcomes.

Rows 839 and 840 belong to training and make no direct contribution to test MSE; rows 844 and 847 are test rows. The MSE decomposition uses the same 158 test rows as direct evaluation and closes within \(7.86 \times 10^{- 17}\). In a score-restoration search, 840/844 is the only successful coordinate pair. Row 844 has improved prediction error but enters the list only after 840 is restored.

\subsection{Confirmation on new inputs and captures}\label{sec:C.3}

The NCI new-input series uses 1,536 fingerprints with no exact packed-fingerprint overlap with the historical 1,017 inputs. Numerical values change in 32/32 captures; the primary Top-20 membership endpoint changes in 0/32. Exploratory order changes occur in 20/32, including 61 strict crossings and one tie-related reversal. All 64 reference controls agree, and healthy continuation reproduces every final score vector and order. Another batch-512 series has identical tensors in 16/16 captures; 15 serialized files differ in container representation.

\begingroup\small\setlength{\tabcolsep}{3pt}
\begin{longtable}{@{}>{\raggedright\arraybackslash}p{\dimexpr 0.10000\linewidth-5.0pt\relax}>{\raggedright\arraybackslash}p{\dimexpr 0.15000\linewidth-5.0pt\relax}>{\raggedright\arraybackslash}p{\dimexpr 0.16000\linewidth-5.0pt\relax}>{\raggedright\arraybackslash}p{\dimexpr 0.16000\linewidth-5.0pt\relax}>{\raggedright\arraybackslash}p{\dimexpr 0.23000\linewidth-5.0pt\relax}>{\raggedright\arraybackslash}p{\dimexpr 0.20000\linewidth-5.0pt\relax}@{}}
\caption{\textbf{All five new SciFact captures.} The first successful budget uses the fixed development priority.}\label{tab:C2}\\
\toprule
\textbf{Capture} & \textbf{Different scores} & \textbf{Changed-membership queries} & \textbf{Changed-order queries} & \textbf{First successful budget} & \textbf{Remaining score differences} \\
\midrule\endfirsthead
\multicolumn{6}{l}{\small Table~\ref{tab:C2} (continued)}\\
\toprule
\textbf{Capture} & \textbf{Different scores} & \textbf{Changed-membership queries} & \textbf{Changed-order queries} & \textbf{First successful budget} & \textbf{Remaining score differences} \\
\midrule\endhead
\bottomrule\endfoot
00 & 591 & 2 & 2 & 8/38 blocks & 464 \\
01 & 585 & 0 & 0 & No changed list to recover & --- \\
02 & 588 & 2 & 2 & 8/38 blocks & 462 \\
03 & 588 & 2 & 3 & 8/38 blocks & 460 \\
04 & 591 & 1 & 1 & 8/38 blocks & 463 \\
\end{longtable}\endgroup

For the captures in Table~\ref{tab:C2}, every query's relevance-label sequence, Recall@5, and nDCG@10 remain unchanged. Aggregate Recall@5 and nDCG@10 remain 0.525 and 0.4636, respectively. The four changed captures recover membership and order at 8/38 blocks. Priorities ranked by development-capture magnitudes need 32 blocks in each; ranking by each capture's own magnitudes needs 32/32/16/4, so priority performance depends on the capture. The seven membership changes are counted as query--capture events.

New activity capture 03 changes membership while RMSE improves from 0.47488 to 0.47453. This series partitions all 1,017 rows into 128 blocks, including the final singleton. Restoring 1/128 blocks recovers membership and order with 218 differences remaining. Across PBMC neighbor search, SciFact retrieval, and compound-activity prediction, all 9,990 intervention conditions undergo two distributed executions involving eight healthy GPUs, with independent checks of the argmax and of metric summation in a canonical order. The stage sorts saved scores; SciFact uses broadcast, local Top-10 selection, and gather across eight GPUs.

\subsection{Recovery costs and stability conditions}\label{sec:C.4}

Let \(\mathcal R\subseteq2^{\mathcal I}\) be the allowed restoration sets over saved-state coordinates \(\mathcal I\). For independent final-score replacements, \(C_{\mathrm{score}}(S)=|S|\), measured in scalar scores. For prescribed disjoint blocks \(B_j\), write \(S=\bigcup_{j\in J_S}B_j\): \(C_{\mathrm{block}}(S)=|J_S|\) counts blocks, and \(C_{\mathrm{frac}}(S)=|J_S|/N_B\) is the dimensionless fraction of the \(N_B\) eligible blocks. Every selected block counts in full, including unchanged coordinates. Fix one such cost \(C:\mathcal R\to\mathbb R_{\geq0}\) and an endpoint \(D\), either the full decision tuple or a named projection:

\[
\rho_D=\min_{S\in\mathcal R:\ D(F(R_S))=D(F(h))}C(S),\qquad
\rho_{\mathrm{state}}=\min_{S\in\mathcal R:\ R_S=h}C(S).
\]

An empty feasible set has minimum \(+\infty\). The state target is exact reconstruction of \(h\), with all other state that drives the continuation fixed. Under the same allowed edits, cost, and deterministic continuation, state restoration recovers the endpoint, giving \(\rho_D\leq\rho_{\mathrm{state}}\). A successful tested budget upper-bounds the minimum. Stochastic agents require repeated comparisons of endpoint distributions.

For \(x\in\mathbb R^N\) and \(1\leq K\leq N\), \(M_K(x)\) is the set of highest-scoring \(K\) candidate indices under a fixed tie rule. At most \(n_{\mathrm{edit}}\) independent final-score edits replace at most \(n_{\mathrm{edit}}\) members. Thus \(|M_K(h)\mathbin{\bigtriangleup}M_K(c)|/2\), with \(\bigtriangleup\) denoting symmetric difference, lower-bounds the scalar score replacements needed to recover reference membership \citep{ref20,ref27}. Hidden-activation blocks can affect several scores. Valid intervals certify membership stability when every selected score's lower bound strictly exceeds every unselected score's upper bound \citep{ref21}.

On a common evaluation index set \(\mathcal V\), let \(y_i\) be the target, \(\widehat y_i^H,\widehat y_i^C\) the reference and candidate predictions, \(e_i=\widehat y_i^H-y_i\), and \(\delta_i=\widehat y_i^C-\widehat y_i^H\). Writing \(\operatorname{mean}_{\mathcal V}x=|\mathcal V|^{-1}\sum_{i\in\mathcal V}x_i\),

\[\Delta MSE:=MSE_C-MSE_H=2\operatorname{mean}_{\mathcal V}(e\delta)+\operatorname{mean}_{\mathcal V}(\delta^2).\]

The cross term can lower error while ranking changes. The numerical experiments distinguish the 1,017-row candidate universe from the 158-row test set \(\mathcal V\).

Operational recovery cost includes detecting the difference, obtaining a reference, locating values, transferring or recomputing them, reranking, and rerunning the agent. Block fractions such as 2/127 and 8/38 measure replacement scope. A rule that uses the current H/C comparison needs a reference for the capture being repaired; a rule developed on earlier captures does not.

\FloatBarrier
\section{Learned tools and complete-state recovery}\label{app:D}

\subsection{Parameters and optimizer state}\label{sec:D.1}

Gradient pairing swaps which physical GPU runs each logical rank, while each logical rank keeps its own inputs. The experiment records local FP32 gradients before reduction and clipping. Seven local-gradient differences lead to three parameter differences after the first update; each branch then independently recomputes 63 subsequent updates on healthy hardware. Every nonzero intervention changes all 1,017 predictions under the common FP64 evaluator. Some recorded losses remain equal at their stored precision.

Learned scientific tools have an additional persistent state. In AdamW \citep{ref30}, \(t\geq1\) is the saved update count and \(d_\theta\) the parameter count. The parameters \(\theta_t\), first moment \(m_t\), and reduced, clipped gradient \(g_t\) lie in \(\mathbb R^{d_\theta}\); the second moment \(v_t\) lies in \(\mathbb R_{\geq0}^{d_\theta}\). The step size is \(\eta_t>0\), decoupled weight decay is \(\lambda\geq0\), moment-decay rates are \(\beta_1,\beta_2\in[0,1)\), and the stabilizer is \(\epsilon>0\). Squares \(g_t^{\odot2}\), square roots, and division are elementwise; bias correction uses the saved counter:

\begin{gather*}
m_{t} = \beta_{1}m_{t - 1} + \left( 1 - \beta_{1} \right)g_{t},\quad\quad v_{t} = \beta_{2}v_{t - 1} + \left( 1 - \beta_{2} \right)g_{t}^{\odot 2},\\
\theta_{t} = \left( 1 - \eta_{t}\lambda \right)\theta_{t - 1} - \eta_{t}\frac{m_{t}/\left( 1 - \beta_{1}^{t} \right)}{\sqrt{v_{t}/\left( 1 - \beta_{2}^{t} \right)} + \epsilon}.
\end{gather*}

AdamW state includes parameters, first and second moments, step counters, optimizer settings, CPU/CUDA random state, and the sample sequence \citep{ref29,ref30}. These experiments use no scheduler, gradient scaler, or dropout. H/C optimizer states are reconstructed by applying their corresponding gradients on healthy hardware.

Eight parameter/moment combinations, a full-restoration control, and a copy control produce ten continuations. Initial differences involve three parameters and seven entries of each moment; the maximum second-moment difference is approximately \(1.02249 \times 10^{- 9}\). CHH, CHC, and HHC change list order, while all combinations preserve membership. Along CCC \ensuremath{\rightarrow} HHC \ensuremath{\rightarrow} HHH, order discordance follows 0 \ensuremath{\rightarrow} 1 \ensuremath{\rightarrow} 0. Every non-HHH combination changes FP64 predictions and gives a small RMSE improvement.

\subsection{Dilution and nonmonotonic order recovery}\label{sec:D.2}

Real All-Reduce changes one rank's residual to \(\delta_{1}/N\). For \(N = 2,4,8\), the recorded L2 norms are \(1.926785 \times 10^{- 3}\), \(9.633925 \times 10^{- 4}\), and \(4.816962 \times 10^{- 4}\). Order discordance follows 0/1/0, and the numbers of differing losses among 63 are 41/34/0 (Figure~\ref{fig:D2}a). After the initial collective, every branch continues on the same healthy coordinating GPU. The four-rank order change is a strict score crossing. Thus the comparison varies the initial residual without changing the platform or parallel layout of the continuation.

\subsection{State divergence under synchronized training}\label{sec:D.3}

The complete matrix contains 96 synchronous data-parallel continuations, 48 fixed continuations after an initial collective, and 16 single-GPU runs with gradient divisor two: 160 executions forming 80 technical-repeat pairs. Membership is unchanged in every row. Order changes in 22 executions, and ten executions preserve all stored losses while parameter states differ. Replicas agree throughout the 96 data-parallel executions, and 40 ZERO/RESTORE comparisons reproduce the reference endpoint.

The eight-rank, two-node SINGLE example has 63 bitwise-identical FP32 losses but differs from its reference in 30,411 parameters, 60,025 first moments, 105,921 second moments, and all 1,017 FP64 predictions. Membership and order are unchanged.

Table~\ref{tab:D1-suite01} consolidates identical repeats while retaining all 160 outcomes, including runs whose parameters diverge while their losses stay identical.

\begin{figure}[H]
\centering\includegraphics[width=\linewidth,height=0.78\textheight,keepaspectratio]{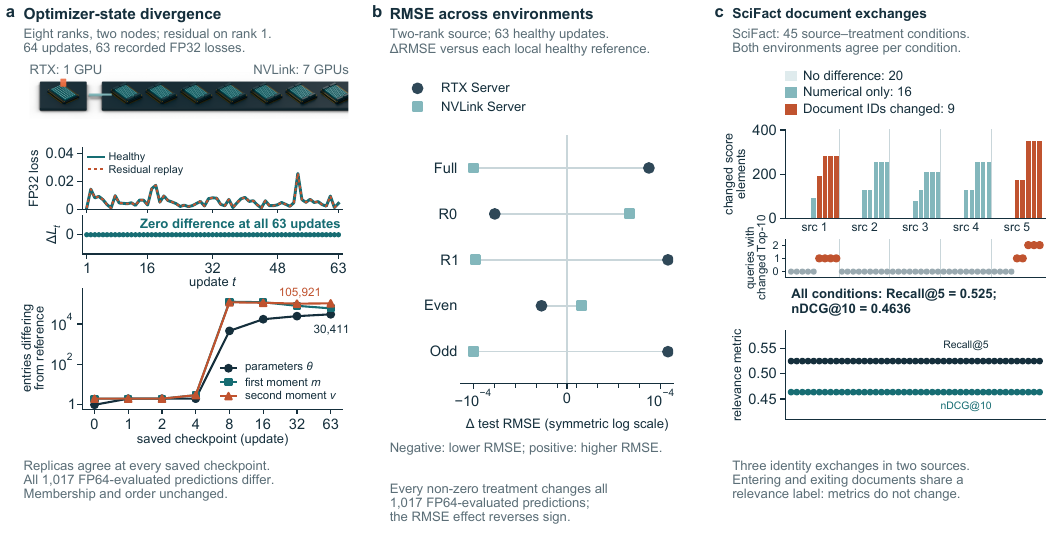}
\caption{Numerical and task endpoints. (a) State divergence with unchanged losses. (b) Reversal of RMSE direction across healthy environments. (c) Document changes with unchanged relevance labels.}\label{fig:D1}
\end{figure}

\subsection{New gradient captures and delayed restoration}\label{sec:D.4}

The new-gradient series has five captures, one of which has no differences. Capture 2 has two local-gradient differences and one parameter difference after update 0. After continuation it differs in 52,128 parameters and all 1,017 FP64 predictions, while 63 stored FP32 losses, membership, and order remain equal.

Captures 0 and 4 change membership and improve RMSE; their losses also change. The delayed-restoration study includes 400 branch continuations (five captures \ensuremath{\times} two healthy environments \ensuremath{\times} four delays \ensuremath{\times} five restoration/control branches \ensuremath{\times} two repeats), plus 40 complete H/C baseline trajectories, yielding 440 executions. Delays are 0, 1, 8, and 32 updates. All 220 repeat pairs agree.

\begin{figure}[H]
\centering\includegraphics[width=\linewidth,height=0.78\textheight,keepaspectratio]{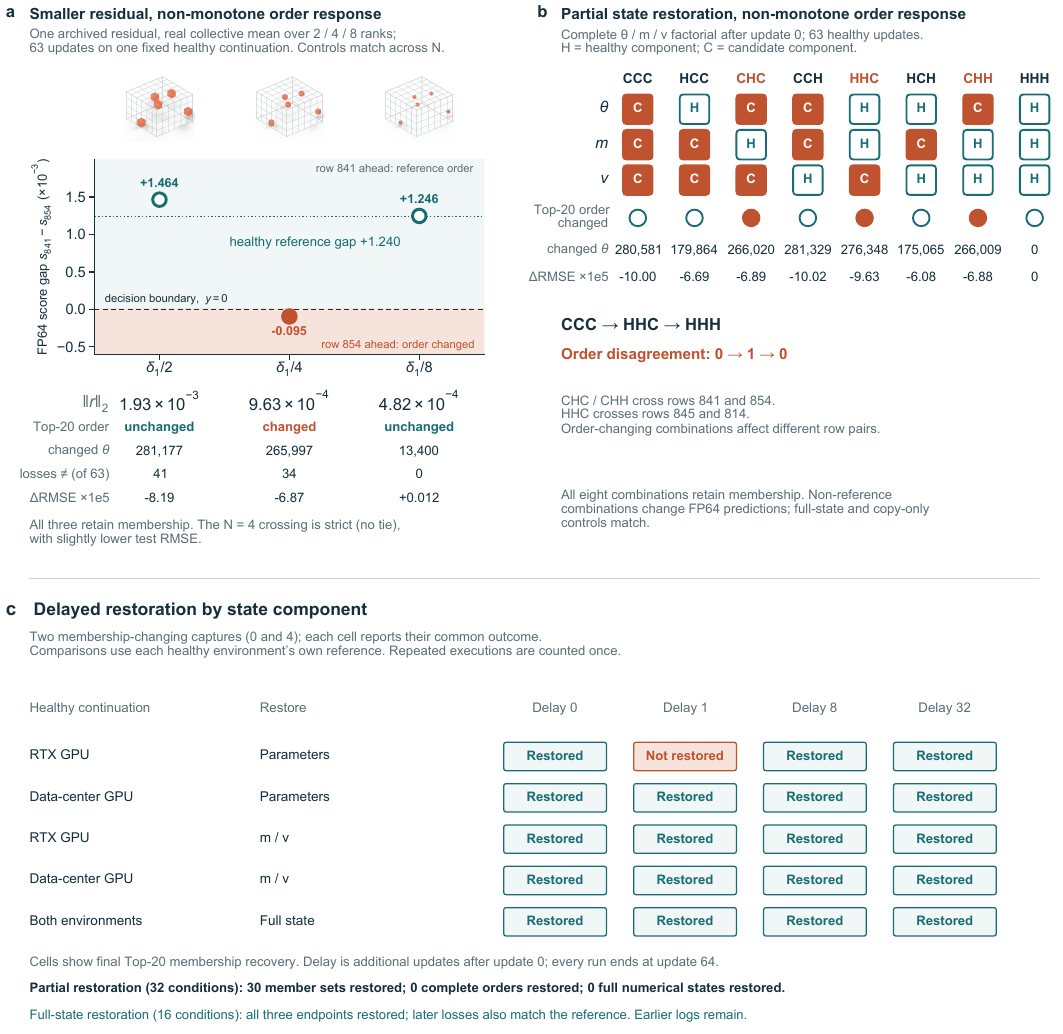}
\caption{Nonmonotonic and delayed recovery. (a) Initial residual dilution with fixed continuation. (b) Parameter/moment combinations from one reconstructed state. (c) Delayed restoration using the two membership-changing new captures. Full-state restoration recovers subsequent computed endpoints without rewriting earlier records.}\label{fig:D2}
\end{figure}

Among the two membership-changing captures, 30/32 partial-restoration configurations recover membership (Figure~\ref{fig:D2}c), and none recovers full order or numerical state. The denominator is two captures \ensuremath{\times} two environments \ensuremath{\times} four delays \ensuremath{\times} two partial-restoration choices. All 16 complete-state configurations recover future endpoints and post-restoration logs; earlier logs remain unchanged. On the healthy RTX platform, parameter-only restoration recovers membership at delays 0, 8, and 32, and fails at delay 1. It recovers membership at all four delays in the data-center environment.

\subsection{Continuation endpoints}\label{sec:D.5}

Table~\ref{tab:D1-suite01} represents all 160 continuations. Each layout contains eight treatments and two technical repeats from the same gradient capture, with 64 updates and 63 subsequent loss observations per execution. All 80 repeat pairs agree on every reported endpoint. H, ZERO, RESTORE, and MATCH\_H have zero in every difference column across all ten layouts (80 executions). The table lists the other four treatments once per repeat pair (80 executions): all change 1,017 predictions and preserve Top-20 membership. Loss/63 counts changed stored losses; O\(\Delta\) indicates changed order; \(\Delta g\) is the reduced-gradient L2 difference. MATCH\_C is compared with MATCH\_H; other treatments are compared with H within their layout.

Let \(\delta_0,\delta_1\) be the saved local-gradient residuals for ranks 0 and 1. SINGLE0 injects \(\delta_0\), SINGLE injects \(\delta_1\), and PAIR injects both before reduction by \(N\); the single-GPU run uses divisor two. ZERO injects a zero residual, and RESTORE replaces the reduced candidate gradient with H before clipping. MATCH\_H sets the whole pre-clipping gradient to a common healthy tensor; MATCH\_C sets it to that tensor plus \((\delta_0+\delta_1)/2\), independent of \(N\).

The single-GPU layout uses one healthy NVLink Server GPU with gradient divisor two. Initial-collective layouts reduce over two nodes with \(N=2,4,8\), then continue on one fixed healthy GPU. Two-node synchronous data parallelism uses one healthy RTX GPU and \(N-1\) healthy NVLink Server GPUs. One-node synchronous data parallelism uses \(N=2,4,8\) GPUs within the NVLink Server.

\begingroup\small\setlength{\tabcolsep}{3pt}
\begin{longtable}{@{}lrrrrr@{}}
\caption{\textbf{Continuation endpoints by computing layout.} Each row represents two identical technical repeats. The four zero-difference controls per layout are consolidated in the text.}\label{tab:D1-suite01}\\
\toprule
Treatment & $\Delta g$ L2 & Changed $\theta$ & Loss/63 & O$\Delta$ & $\Delta$RMSE \\
\midrule\endfirsthead
\multicolumn{6}{l}{\small Table~\ref{tab:D1-suite01} (continued)}\\
\toprule
Treatment & $\Delta g$ L2 & Changed $\theta$ & Loss/63 & O$\Delta$ & $\Delta$RMSE \\
\midrule\endhead
\bottomrule\endfoot
\multicolumn{6}{@{}l}{\textbf{Single GPU; divisor 2}}\\*
SINGLE0 & \(3.856\times10^{-4}\) & 164,218 & 9 & 0 & \(3.240\times10^{-6}\) \\
SINGLE & \(1.927\times10^{-3}\) & 281,177 & 41 & 0 & \(-8.191\times10^{-5}\) \\
PAIR & \(1.965\times10^{-3}\) & 280,581 & 41 & 0 & \(-9.996\times10^{-5}\) \\
MATCH\_C & \(1.965\times10^{-3}\) & 280,581 & 41 & 0 & \(-9.996\times10^{-5}\) \\
\multicolumn{6}{@{}l}{\textbf{Initial collective; N=2}}\\*
SINGLE0 & \(3.856\times10^{-4}\) & 164,218 & 9 & 0 & \(3.240\times10^{-6}\) \\
SINGLE & \(1.927\times10^{-3}\) & 281,177 & 41 & 0 & \(-8.191\times10^{-5}\) \\
PAIR & \(1.965\times10^{-3}\) & 280,581 & 41 & 0 & \(-9.996\times10^{-5}\) \\
MATCH\_C & \(1.965\times10^{-3}\) & 280,581 & 41 & 0 & \(-9.996\times10^{-5}\) \\
\multicolumn{6}{@{}l}{\textbf{Initial collective; N=4}}\\*
SINGLE0 & \(1.928\times10^{-4}\) & 174,718 & 12 & 0 & \(-1.425\times10^{-5}\) \\
SINGLE & \(9.634\times10^{-4}\) & 265,997 & 34 & 1 & \(-6.875\times10^{-5}\) \\
PAIR & \(9.825\times10^{-4}\) & 181,439 & 12 & 0 & \(-5.861\times10^{-5}\) \\
MATCH\_C & \(1.965\times10^{-3}\) & 280,581 & 41 & 0 & \(-9.996\times10^{-5}\) \\
\multicolumn{6}{@{}l}{\textbf{Initial collective; N=8}}\\*
SINGLE0 & \(9.641\times10^{-5}\) & 173,996 & 12 & 0 & \(-1.423\times10^{-5}\) \\
SINGLE & \(4.817\times10^{-4}\) & 13,400 & 0 & 0 & \(1.201\times10^{-7}\) \\
PAIR & \(4.912\times10^{-4}\) & 176,552 & 12 & 0 & \(-1.420\times10^{-5}\) \\
MATCH\_C & \(1.965\times10^{-3}\) & 280,581 & 41 & 0 & \(-9.996\times10^{-5}\) \\
\multicolumn{6}{@{}l}{\textbf{Two-node DDP; N=2}}\\*
SINGLE0 & \(3.856\times10^{-4}\) & 75,017 & 2 & 0 & \(1.719\times10^{-5}\) \\
SINGLE & \(1.927\times10^{-3}\) & 210,288 & 18 & 1 & \(8.112\times10^{-5}\) \\
PAIR & \(1.965\times10^{-3}\) & 208,785 & 18 & 1 & \(7.602\times10^{-5}\) \\
MATCH\_C & \(1.965\times10^{-3}\) & 263,824 & 31 & 1 & \(-3.061\times10^{-5}\) \\
\multicolumn{6}{@{}l}{\textbf{Two-node DDP; N=4}}\\*
SINGLE0 & \(1.928\times10^{-4}\) & 901 & 0 & 0 & \(-2.192\times10^{-9}\) \\
SINGLE & \(9.634\times10^{-4}\) & 270,748 & 35 & 0 & \(-1.169\times10^{-4}\) \\
PAIR & \(9.825\times10^{-4}\) & 270,768 & 35 & 0 & \(-1.168\times10^{-4}\) \\
MATCH\_C & \(1.965\times10^{-3}\) & 275,172 & 38 & 1 & \(2.416\times10^{-4}\) \\
\multicolumn{6}{@{}l}{\textbf{Two-node DDP; N=8}}\\*
SINGLE0 & \(9.641\times10^{-5}\) & 203,896 & 16 & 1 & \(1.075\times10^{-5}\) \\
SINGLE & \(4.817\times10^{-4}\) & 30,411 & 0 & 0 & \(1.245\times10^{-7}\) \\
PAIR & \(4.912\times10^{-4}\) & 203,735 & 15 & 1 & \(2.633\times10^{-5}\) \\
MATCH\_C & \(1.965\times10^{-3}\) & 240,107 & 26 & 1 & \(-5.218\times10^{-5}\) \\
\multicolumn{6}{@{}l}{\textbf{One-node DDP; N=2}}\\*
SINGLE0 & \(3.856\times10^{-4}\) & 15,893 & 0 & 0 & \(1.586\times10^{-6}\) \\
SINGLE & \(1.927\times10^{-3}\) & 265,710 & 25 & 1 & \(2.597\times10^{-4}\) \\
PAIR & \(1.965\times10^{-3}\) & 264,585 & 25 & 1 & \(1.543\times10^{-4}\) \\
MATCH\_C & \(1.965\times10^{-3}\) & 280,581 & 41 & 0 & \(-9.996\times10^{-5}\) \\
\multicolumn{6}{@{}l}{\textbf{One-node DDP; N=4}}\\*
SINGLE0 & \(1.928\times10^{-4}\) & 2,691 & 1 & 0 & \(5.856\times10^{-8}\) \\
SINGLE & \(9.634\times10^{-4}\) & 29,909 & 0 & 0 & \(1.622\times10^{-7}\) \\
PAIR & \(9.825\times10^{-4}\) & 30,552 & 1 & 0 & \(2.020\times10^{-7}\) \\
MATCH\_C & \(1.965\times10^{-3}\) & 249,601 & 30 & 0 & \(-5.807\times10^{-5}\) \\
\multicolumn{6}{@{}l}{\textbf{One-node DDP; N=8}}\\*
SINGLE0 & \(9.641\times10^{-5}\) & 153,661 & 9 & 0 & \(-2.996\times10^{-5}\) \\
SINGLE & \(4.817\times10^{-4}\) & 202,551 & 20 & 0 & \(-2.941\times10^{-5}\) \\
PAIR & \(4.912\times10^{-4}\) & 202,687 & 20 & 0 & \(-2.937\times10^{-5}\) \\
MATCH\_C & \(1.965\times10^{-3}\) & 241,825 & 22 & 1 & \(2.300\times10^{-5}\) \\
\end{longtable}\endgroup

\FloatBarrier
\section{Error propagation and simulation}\label{app:E}

\subsection{What correct communication propagates}\label{sec:E.1}

All 7,680 recorded transfers are exact, while exchanging shard ownership changes molecular-ranking failures from 10/10 to 0/10. PBMC split into 1,319-column shards has no numerical difference in 20/20 executions. These comparisons show that scheduling changes the actual computation exposed to an anomaly, while correct communication propagates the values already produced.

Five SciFact captures and nine interventions per capture produce 45 conditions: nine change decisions, 16 change only numerical values, and 20 remain identical. Full endpoints agree across two healthy environments. The nine decision-changing conditions correspond to three document exchanges in two captures. Both incoming and outgoing documents have relevance label zero, and each exchange occurs at rank ten. Recall@5 = 0.525 and nDCG@10 approximately 0.4636 therefore remain constant in all 45 conditions, with equality for every query.

Dynamic training behaves differently: the RMSE-change direction reverses across the two healthy environments for all five nonzero gradient interventions: the full residual, rank-0-only and rank-1-only residuals, and the even-index and odd-index parts of the flattened full residual (Figure~\ref{fig:D1}b). Each effect is evaluated against its own environment's reference.

\subsection{Normalization and clipping}\label{sec:E.2}

Changed digit-recognition features produce different logits when passed to a fixed healthy classification head, while all predicted classes remain the same. In QKV continuation, 66 of 2,048 probability rows have total variation at least 0.01. Section 4.3 reports normalization, maximum-attention changes, and gradient clipping.

\subsection{Behavioral components and sequence modeling}\label{sec:E.3}

Across 22 configurations, the dynamic predictor has Brier score 0.0902 and accuracy 86.36\%, compared with 0.1851 and 68.18\% for the static baseline. For the sequence model, each task has 4,096 correctly predicted call labels and 2,158 spatial events meeting the criterion. Magnitude-profile L1 distance is 0.15836 against 0.21794 for the baseline.

For replay, the residual interface keeps 16,437 measured residual records in a residual bank, one entry per record. Each record comes from one computation call and holds the coordinates and values of its differences together with the call information. The records trace back to 240 captures; a capture contains several calls and can contribute several records. At a BF16 storage boundary, 878/1,024 selected differences survive conversion.

\textbf{Four-state readout and spatial rules.} Let \(z_t\in\{0,1,2,3\}\) be the internal state before call \(t=0,1,\ldots\), \(n_t\) its input row count, and \(w\) the row-count mapping below. Starting at \(z_0=0\), the protocol updates

\[z_{t + 1} = \left( z_{t} + w\left( n_{t} \right) \right)\bmod 4.\]

Row counts \(n_{t} \in \{ 31,65,879,930,945,946\}\) receive weights 2, 1, 2, 1, 1, and 1. An event is predicted when \(n_{t} \in \{ 930,945,946\}\), or when \(n_{t} = 65\) and \(z_{t} = 3\). The latter predicts row 33 and columns 128--143. Large shapes use row sets \{929,945\} or \{897,913\} according to state parity, omitting rows outside the shape. States 1/2 use columns 32--63; the others use columns 0--31. The spatial criterion requires the correct row set, nonempty changed columns, and containment of those columns in the predicted interval.

Sequence-model experiments use 8, 16, and 32 session seeds, with 128 calls per session.

The transition-distribution, switching-fraction, lag-1, and run-length errors meet their absolute thresholds, although each is larger than the corresponding baseline error. The alternating-condition fraction error exceeds its threshold (0.15625 versus 0.15), and the temporal composite error exceeds that of the residual-sampling baseline, which samples within each condition (0.24392 versus 0.14430). The joint temporal criterion is therefore not met in this fixed-seed evaluation.

For each of \(n=32\) conditions, let \(p_i,\widehat p_i\in\{x\in[0,1]^3:\sum_{k=1}^3x_k=1\}\) be the observed and simulated magnitude profiles. Each component is the fraction of calls selected by the predeclared event rule that falls in its bin; the denominator is the condition's positive event-call count. Observed magnitude is \(\max_k|c_k-h_k|\), on the captured tensor's numerical scale. Bins are \([0,10^{10}]\), \((10^{10},10^{24}]\), and \((10^{24},\infty)\). The profiles are matched as unlabeled collections:

\[L_{\mathrm{profile}}=\min_{\pi\in\mathfrak S_n}\frac1n\sum_{i=1}^n\|p_i-\widehat p_{\pi(i)}\|_1,\qquad \|x\|_1=\sum_{k=1}^3|x_k|.\]

Here \(\mathfrak S_n\) contains all permutations of \(\{1,\ldots,n\}\), matching each profile once. The distance lies in \([0,2]\). Sorted-channel MAE is 0.03219 versus 0.06549; it averages absolute differences after independently sorting each bin's proportions across conditions, then averages over the three bins.

\FloatBarrier
\section{Agent cohorts and paired analysis}\label{app:F}

\subsection{Claim-level records and execution protocol}\label{sec:F.1}

One Qwen3.5-4B series records 642 paired groups and 2,652 selected execution records. Its inventory contains 184 initial groups, 37 resumed groups, and 421 sustained-batching groups. The 184 initial groups comprise 5 groups for mechanism analysis and 179 unselected groups.

Of the 37 resumed groups, 24 complete the core conditions and access the saved numerical site. Scores change in 13 of these and remain equal in 11. Three mechanism records concern claim 831 and two captures. Reference branches read document 12737132 on TEAD/YAP regulation; candidate branches read 735130 on myosin phosphorylation in Drosophila oogenesis, alongside a shared document. Every condition in these three records returns the correct NEI verdict.

For records 00/0, 00/2, and 04/0, H1/H2 read and citation sets agree, C changes both sets, full restoration recovers H1 in 3/3, and transplantation reproduces C in 3/3. Selected restoration recovers H1 in 2/3, failing for 04/0. The 11 unchanged-score controls do not show this combination of outcomes. Section 2.4 and Appendix \ref{sec:J.1} show that the selected mask restores the full affected row in all three records.

The original SciFact test split lacks verdict labels. Evaluation aligns development labels and BEIR queries by their actual IDs. Labels remain evaluator-only. Tools expose structured actions, and log invalid JSON, unknown actions, numerical alerts, rejected premature finishes, exhausted budgets, and service failures. Search passes IDs, titles, and numerical scores to the model; \texttt{read} delivers abstract sentences. Masking removes title and abstract text with IDs and scores unchanged.

\subsection{Common subsets and sensitivity analyses}\label{sec:F.2}

Let \(\mathcal U\) be a nonempty jointly valid group set, with execution trajectories \(C_u,H1_u,H2_u\) for group \(u\) and selected endpoint map \(D\). The paired summary is

\[\widehat\Delta_D=\frac1{|\mathcal U|}\sum_{u\in\mathcal U}\left[\mathbf1\{D(C_u)\ne D(H1_u)\}-\mathbf1\{D(H2_u)\ne D(H1_u)\}\right].\]

Here \(\mathbf1\) is the binary indicator.

Candidate and reference-repeat discordances are reported together. The summary above weights groups equally and differs from the capture-balanced estimates: within each task--capture unit, technical repeats are averaged first, then tasks within a capture, then captures equally. Capture bootstrap resamples captures; crossed capture \ensuremath{\times} task bootstrap applies weights on both axes. Table~\ref{tab:F2} reports the estimates and intervals using five captures and the original 10,000 fixed-seed resamples; all intervals include zero.

\begingroup\small\setlength{\tabcolsep}{3pt}
\begin{longtable}{@{}>{\raggedright\arraybackslash}p{\dimexpr 0.25000\linewidth-4.5pt\relax}>{\raggedright\arraybackslash}p{\dimexpr 0.25000\linewidth-4.5pt\relax}>{\raggedright\arraybackslash}p{\dimexpr 0.25000\linewidth-4.5pt\relax}>{\raggedright\arraybackslash}p{\dimexpr 0.25000\linewidth-4.5pt\relax}@{}}
\caption{\textbf{Recorded sensitivity estimates.} Intervals use the original 10,000 fixed-seed resamples. Units are percentage points of excess discordance.}\label{tab:F2}\\
\toprule
\textbf{Series / endpoint} & \textbf{Capture-balanced estimate} & \textbf{Capture-bootstrap interval} & \textbf{Crossed-bootstrap interval} \\
\midrule\endfirsthead
\multicolumn{4}{l}{\small Table~\ref{tab:F2} (continued)}\\
\toprule
\textbf{Series / endpoint} & \textbf{Capture-balanced estimate} & \textbf{Capture-bootstrap interval} & \textbf{Crossed-bootstrap interval} \\
\midrule\endhead
\bottomrule\endfoot
Initial, unselected / actions & 1.626 & {[}\ensuremath{-}3.871, 8.750{]} & {[}\ensuremath{-}6.818, 12.121{]} \\
Initial, unselected / read membership & \ensuremath{-}0.667 & {[}\ensuremath{-}2.000, 0.000{]} & {[}\ensuremath{-}7.194, 4.511{]} \\
Initial, unselected / citations & 0.793 & {[}\ensuremath{-}3.871, 6.250{]} & {[}\ensuremath{-}7.194, 9.783{]} \\
Initial, unselected / verdict & 0.000 & {[}0, 0{]} & {[}0, 0{]} \\
Sustained / actions & \ensuremath{-}0.376 & {[}\ensuremath{-}3.455, 2.609{]} & {[}\ensuremath{-}5.159, 3.975{]} \\
Sustained / read membership & \ensuremath{-}0.189 & {[}\ensuremath{-}0.566, 0.000{]} & {[}\ensuremath{-}1.551, 1.181{]} \\
Sustained / citations & \ensuremath{-}0.555 & {[}\ensuremath{-}2.264, 0.600{]} & {[}\ensuremath{-}3.600, 2.024{]} \\
Sustained / verdict & 0.016 & {[}\ensuremath{-}0.577, 0.625{]} & {[}\ensuremath{-}1.124, 1.029{]} \\
\end{longtable}\endgroup

The sensitivity records report jointly valid subsets of 158/179 for the unselected groups in the initial series and 382/421 for sustained batching. The 421 sustained-batching groups comprise 382 valid comparisons and 39 others.

The unweighted verdict counts are equal, two versus two, while the capture-balanced estimate is 0.016 percentage points.

Appendix F.3 reports the 168 continuations from saved prefixes; Appendix G.1 gives PBMC and NCI completion denominators and verification.

\textbf{Native language-model execution.} In the Case 1 series, none of 220 candidate episodes calls a tool; 220 requests are flagged by vLLM's request-level NaN diagnostic. The paired healthy branch completes 198/220. The earlier Case 2 series records 240 such flagged requests and zero candidate completions, versus 222/240 healthy completions. A matched-reference series completes 8/8. The counts are increments of \texttt{vllm:corrupted\_requests\_total}: 228 \ensuremath{-} 8 for Case 1: size-dependent and 240 \ensuremath{-} 0 for Case 2: expanded-trigger. The corresponding public vLLM 0.28.0 implementation clears the periodic logger's local count before printing “Corrupted”, while updating the Prometheus counter independently. Each candidate episode made one model request and executed no tool. The NaN records contain request-level flags without raw-logit tensors or token/operation traces, so their origin is unknown.

\textbf{Execution windows.} A sustained SciFact phase runs 1,684 episodes in 846 seconds at 486.6 output tokens/s. The prefix series takes 28.8 seconds for 168 continuations. Baseline Qwen episodes take 257.0 seconds, excluding model loading, for 145,742 output tokens, or 567.1 output tokens/s. Gemma episodes take 579.0 seconds for 413,558 output tokens, or 714.2 output tokens/s.

\subsection{Saved prefixes and visible history}\label{sec:F.3}

All 24 completed, exposed groups from the resumed series receive seven continuation conditions, yielding 168 continuations. Relative to the healthy prefix, the candidate prefix and its repeat each change citations in 6/24 groups; tool-only replacement changes them in 7/24. A repeat of the healthy prefix, joint replacement of visible history and tool state, and a fresh healthy replay each have 2/24. Final verdicts agree throughout.

\begin{figure}[H]
\centering\includegraphics[width=\linewidth,height=0.78\textheight,keepaspectratio]{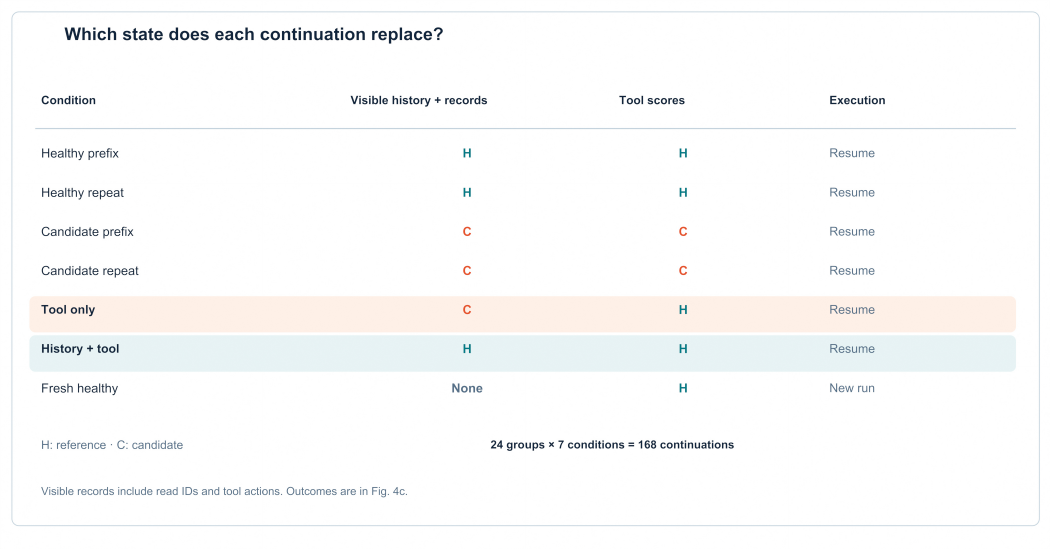}
\caption{State replacement by continuation condition. H/C denote reference and candidate contents. Tool-only replacement retains candidate visible history; joint replacement restores both history and tool state; a fresh execution starts without an existing prefix. Each condition has 24 groups, giving 168 continuations.}\label{fig:F2}
\end{figure}

The replacement scope includes visible messages, delivered documents, citable sentences, tool-action records, and tool scores as specified by condition (Figure~\ref{fig:F2}). We record whether each continuation revisits the repaired query. Prefix interventions leave the server KV cache unchanged, and some continuations do not revisit the query.

\FloatBarrier
\section{Biological-tool evaluations}\label{app:G}

\subsection{Language-model tool execution under ordinary precision}\label{sec:G.1}

NCI provides similarity search and descriptor reads, allowing the agent to choose its next query \citep{ref35,ref49}. PBMC fixes preprocessing, cell/gene IDs, self-match masking, and normalization, then records neighbor inspection, marker checks, and review selections \citep{ref48}. The cell labels come from Louvain clustering. Mean scaled-feature differences guide marker inspection.

Ordinary BF16 score rounding changes PBMC actions or marker selections in 9/30 pairs, compared with 0/30 each for healthy repeat and zero treatment. Three PBMC runs contain 3,264 records, each comprising eight development and 1,080 confirmation records. Verification reconstructs 2,160 cross-run pairs: healthy versus repeat has 1,068/1,080 jointly completed pairs, and repeat versus matched reference has 1,071/1,080. Their actions and final structured selections agree.

NCI's two reference environments use the same model. Confirmation completions are 250/256 and 249/256. Subsets complete under all four conditions contain 61/64 and 62/64 tasks; rounding changes final membership in 11/61 and 13/62, versus healthy-repeat differences of 0/61 and 1/62. Reference repeats differ in their valid actions in 5/61 and 7/62 tasks, so action sequences can differ while final membership stays the same. Rejected premature finishes are excluded from valid-action comparisons; invalid calls and budget exhaustions remain recorded. Verification reconstructs similarity scores, BF16 rounding, ranking, self-exclusion, descriptors, and final selections.

ESOL executes fit, predict, assay, and finish \citep{ref47}. Here assay reveals the requested offline query-pool label; the tool then refits within a fixed budget. The scaffold split separates initially labeled rows, query-pool rows, and test rows. The 24 seeds share one scaffold split. Neither measured nor precomputed predicted solubility is an input feature.

Initially, all four development runs in each environment fail to finish. Adding explicit progress feedback without increasing the budget yields 4/4 development and 96/96 confirmation completions per environment. In each environment, rounding changes sampled membership in 12/24 seeds and sample order in 15/24; healthy repeat and zero treatment change neither. Test RMSE increases in ten seeds, decreases in five, and is unchanged in nine in each environment. RMSE is recalculated from saved predictions and official labels, without refitting the random forest.

NCI and ESOL use progress feedback and produce 720 saved records. Across all 200 ESOL records and 3,616 model requests, reconstructed initial and tool observations match the saved request contents. Model messages exclude test IDs, labels, predictions, and RMSE. Pool queries reveal only the requested row's label. All 200 rejected premature finishes remain recorded.

\subsection{Deterministic-controller protocols}\label{sec:G.2}

Two CPU-controller protocols each contain 352 executions on real datasets with ordinary BF16 score rounding (Protocol II in Table~\ref{tab:G1}). The controllers deterministically choose subsequent tool actions. WDBC supplies cytological morphology features \citep{ref50}.

\begingroup\small\setlength{\tabcolsep}{3pt}
\begin{longtable}{@{}>{\raggedright\arraybackslash}p{\dimexpr 0.14286\linewidth-5.143pt\relax}>{\raggedright\arraybackslash}p{\dimexpr 0.14286\linewidth-5.143pt\relax}>{\raggedright\arraybackslash}p{\dimexpr 0.14286\linewidth-5.143pt\relax}>{\raggedright\arraybackslash}p{\dimexpr 0.14286\linewidth-5.143pt\relax}>{\raggedright\arraybackslash}p{\dimexpr 0.14286\linewidth-5.143pt\relax}>{\raggedright\arraybackslash}p{\dimexpr 0.14286\linewidth-5.143pt\relax}>{\raggedright\arraybackslash}p{\dimexpr 0.14286\linewidth-5.143pt\relax}@{}}
\caption{\textbf{Protocol II.} Counts refer to the same fixed task units across conditions.}\label{tab:G1}\\
\toprule
\textbf{Task} & \textbf{Units} & \textbf{Executions} & \textbf{H/H action changes} & \textbf{Zero-treatment action changes} & \textbf{BF16 action changes} & \textbf{BF16 final-membership changes} \\
\midrule\endfirsthead
\multicolumn{7}{l}{\small Table~\ref{tab:G1} (continued)}\\
\toprule
\textbf{Task} & \textbf{Units} & \textbf{Executions} & \textbf{H/H action changes} & \textbf{Zero-treatment action changes} & \textbf{BF16 action changes} & \textbf{BF16 final-membership changes} \\
\midrule\endhead
\bottomrule\endfoot
NCI, four rounds of descriptor feedback & 64 & 256 & 0 & 0 & 10 & 3 \\
WDBC, five active-sampling batches & 24 & 96 & 0 & 0 & 0 & 0 \\
\end{longtable}\endgroup

NCI Protocol II changes query paths in 2/64 tasks. With fixed initial queries and read budgets 1, 2, 3, 5, and 10, membership-change counts are 0/0/0/2/3 and order-change counts are 0/0/0/2/7. This budget comparison reuses the 64 tasks.

Protocol I reports NCI action and membership changes in 11/64 and 3/64, and WDBC changes in 3/24 and 1/24; the two protocols are separate implementations with different dataset snapshots.

\FloatBarrier
\section{Numerical captures and agent confirmation}\label{app:J}

\subsection{Existing masks and the actual exposed budget}\label{sec:J.1}

For five complete SciFact H/C arrays, the recorded verification checks original values, row and column IDs, query text, and the mask applied to claim 831. Table~\ref{tab:J1} gives global and visited-row mask counts.

\begingroup\small\setlength{\tabcolsep}{3pt}
\begin{longtable}{@{}>{\raggedright\arraybackslash}p{\dimexpr 0.16667\linewidth-5.0pt\relax}>{\raggedright\arraybackslash}p{\dimexpr 0.16667\linewidth-5.0pt\relax}>{\raggedright\arraybackslash}p{\dimexpr 0.16667\linewidth-5.0pt\relax}>{\raggedright\arraybackslash}p{\dimexpr 0.16667\linewidth-5.0pt\relax}>{\raggedright\arraybackslash}p{\dimexpr 0.16667\linewidth-5.0pt\relax}>{\raggedright\arraybackslash}p{\dimexpr 0.16667\linewidth-5.0pt\relax}@{}}
\caption{\textbf{Global and claim-831 mask scope.}}\label{tab:J1}\\
\toprule
\textbf{Capture} & \textbf{Globally different scores} & \textbf{Globally replaced different scores} & \textbf{Different scores in claim-831 row} & \textbf{Within-row replacement} & \textbf{Remaining row differences} \\
\midrule\endfirsthead
\multicolumn{6}{l}{\small Table~\ref{tab:J1} (continued)}\\
\toprule
\textbf{Capture} & \textbf{Globally different scores} & \textbf{Globally replaced different scores} & \textbf{Different scores in claim-831 row} & \textbf{Within-row replacement} & \textbf{Remaining row differences} \\
\midrule\endhead
\bottomrule\endfoot
00 & 591 & 127 & 16 & 16/16 & 0 \\
01 & 585 & 127 & 16 & 16/16 & 0 \\
02 & 588 & 126 & 14 & 14/14 & 0 \\
03 & 588 & 128 & 16 & 16/16 & 0 \\
04 & 591 & 128 & 16 & 16/16 & 0 \\
\end{longtable}\endgroup

The three mechanism records use captures 00 and 04. The global mask selects 331,712/1,554,900 coordinates, corresponding to 8/38 row blocks, and replaces every changed value of the visited query. Selected restoration 2/3 versus full restoration 3/3 therefore cannot be explained by a smaller within-row budget. The selected mask performs full-row replacement while leaving differences elsewhere in the array.

\subsection{New captures and matched references for both GPU classes}\label{sec:J.2}

Initial acquisition produces 44 complete arrays: two Cases \ensuremath{\times} two query batches \ensuremath{\times} eight repeats gives 32 candidate arrays, plus 12 reference arrays measured around the capture rounds (Table~\ref{tab:J2}). All use the same 300 \ensuremath{\times} 512 query and 5,183 \ensuremath{\times} 512 document representations. Scores from BF16-autocast general matrix multiplication (GEMM) are stored in FP32. The captured SciFact retrieval scores contain no NaN or Inf values and are replayed to agents running on healthy hardware. Appendix F.2 reports the separate native-model logit diagnostic.

All captures reuse the same saved query and document representations. The tested operation is a CUDA BF16-autocast GEMM of a \(B\times512\) query matrix and a \(512\times5183\) document matrix, with \(B=8\) (final batch 4) or \(B=300\). TF32 is disabled and BF16 reduced-precision reduction is enabled. Outputs are converted to FP32, concatenated into a \(300\times5183\) array, transferred to host memory, and saved as \texttt{.npy} files.

\begingroup\small\setlength{\tabcolsep}{3pt}
\begin{longtable}{@{}>{\raggedright\arraybackslash}p{\dimexpr 0.20000\linewidth-4.8pt\relax}>{\raggedright\arraybackslash}p{\dimexpr 0.20000\linewidth-4.8pt\relax}>{\raggedright\arraybackslash}p{\dimexpr 0.20000\linewidth-4.8pt\relax}>{\raggedright\arraybackslash}p{\dimexpr 0.20000\linewidth-4.8pt\relax}>{\raggedright\arraybackslash}p{\dimexpr 0.20000\linewidth-4.8pt\relax}@{}}
\caption{\textbf{Initial capture comparison.}}\label{tab:J2}\\
\toprule
\textbf{GPU class} & \textbf{Query batch} & \textbf{Numerically changed captures} & \textbf{Different scores per capture} & \textbf{Captures with changed Top-10 membership} \\
\midrule\endfirsthead
\multicolumn{5}{l}{\small Table~\ref{tab:J2} (continued)}\\
\toprule
\textbf{GPU class} & \textbf{Query batch} & \textbf{Numerically changed captures} & \textbf{Different scores per capture} & \textbf{Captures with changed Top-10 membership} \\
\midrule\endhead
\bottomrule\endfoot
Case 1: size-dependent & 8 & 8/8 & 54--88 & 0/8 \\
Case 1: size-dependent & 300 & 7/8 & 0--24 & 0/8 \\
Case 2: expanded-trigger & 8 & 8/8 & 588--592 & 7/8 \\
Case 2: expanded-trigger & 300 & 8/8 & 328--471 & 8/8 \\
\end{longtable}\endgroup

Five additional rounds of 20 reference/candidate arrays add 100 arrays, including acquisition while a healthy agent is running. The six rounds total 144 arrays and 72 matched references. Every matched reference agrees bitwise with independent healthy outputs at its batch size; the two independent healthy repeats per batch also agree. A round of 20 capture computations runs while the Gemma agent episodes execute, with reference measurements accompanying the round. The agent runs on healthy hardware.

For the tested inputs and batch shapes, the bitwise agreement of matched and independent healthy references makes the candidate score differences directly comparable. The separate exact probes, conditional arithmetic checks, and target-SM controls comparing FFMA substitution with sham routing in Appendices~\ref{sec:A.2}--\ref{sec:A.3} support an anomalous execution path in each GPU class. The PBMC precision control in Appendix~\ref{sec:A.3} shows why this distinction matters: normal implementation-dependent BF16 reduction can also change neighbor membership across healthy references.

\subsection{Capture split and agent interface}\label{sec:J.3}

Within each Case/batch combination, captures 0--1 develop coordinate priorities and 2--7 confirm them. At most one coordinate per query is selected, ordered by development Top-10 membership-change frequency, score-difference magnitude, and original index. A selected coordinate may coincide with no actual H/C difference in a later row; such a condition has zero effective replacement.

The free-text interface attempts 567 episodes: 402 finish and 165 exhaust their budget, while all bypass the original query's saved numerical state. Binding an explicit query ID to the saved row resolves this access failure.

The confirmation interface binds an explicit query ID to the original claim text and saved row. After one search, the model independently chooses up to six returned documents to read, then supplies a verdict and sentence citations within 16 calls. Feedback reports remaining budget and available actions. Three development checks precede confirmation; Appendix \ref{sec:J.6} reports the Gemma outcomes.

The task measures evidence verification after fixed retrieval. Query projection and scoring are completed before agent execution. During each fixed-query episode, \texttt{search} reads and stably sorts the saved score row, returning document IDs, titles, and scores.

Qwen3.5-4B~\citep{ref58} uses official, unquantized BF16 weights; checkpoint-provided FP32 auxiliary tensors remain unchanged. The H1, H2, C, restoration, and transplantation conditions run on healthy hardware with 64 concurrent requests and matched service settings. Gemma 3 4B IT~\citep{ref59} uses its official BF16 checkpoint under the same task schedule, with model-specific template details below.

\subsection{Baseline Qwen results}\label{sec:J.4}

The mechanism cohort, selected by upstream membership change, contains 27 Case 2 capture--claim combinations and 11 distinct claims. Three technical repeats of each request across seven conditions yield 567 completed, exposed trajectories. There are 30 comparison groups at query batch 8 and 51 at query batch 300, totaling 81.

The 12 batch-300 read changes involve claim 431 in nine groups and claim 729 in three. Full restoration recovers every reference read set. Healthy repeat and zero treatment match H1 reads, citations, and verdicts in all 81 groups. Selected restoration does not recover the reference read sets, and transplantation does not reproduce the candidate read sets. For one claim-431 capture in two repeats, restoring one score with 95 row differences remaining returns verdict and citations to H, but reads remain different. Transplantation retains H's verdict instead of reproducing C's.

Claim 431 is SUPPORT-labeled. Two candidate records return SUPPORT with delivered citations, whereas their reference/restored branches return NEI. One claim-56 record changes H's SUPPORT to NEI with unchanged reads and citations. Full restoration reverses all three verdict differences.

Masking titles and abstracts runs H1/H2/C over the same 81 groups, totaling 243 completed and exposed trajectories, all with uncited NEI. H1 changes verdict relative to full evidence in 57/81 groups. Masked read-membership discordance is 15/30 for C versus H1 and 0/30 for H2 versus H1 at batch 8, and 38/51 and 5/51 at batch 300. The interface still allows the model to choose listed IDs without document content.

Unselected confirmation fixes claims 350 and 1180 across two Cases, two batches, six confirmation captures, three repeats, and four conditions: 576 trajectories. Each Case/batch stratum has 36 completed, exposed groups, with zero differences in reads, citations, actions, or verdicts.

The baseline Qwen schedule saves 1,386 trajectories and 5,155 model requests.

\subsection{Baseline Gemma under the matched task protocol}\label{sec:J.6}

Gemma uses the official google/gemma-3-4b-it BF16 checkpoint and chat template with text inputs, a 16,384-token context limit, 64 concurrent tasks, 95\% GPU-memory allocation, and the built-in sampler. Speculative decoding is disabled.

Task order, arrays, masks, zero temperature, repeat parameters, 384-token output limit, and 16-call budget match Qwen. Three development checks search and read; two finish and one exhausts the budget after an invalid finish format. Rejected actions consume budget, without automatic argument correction or selective retry.

Each mechanism condition schedules 81 episodes. Completions are H1 40, H2 39, C 39, full restoration 41, selected restoration 37, transplantation 34, and zero treatment 39. All 27 invalid-JSON outputs reach the token limit. The main rejected calls have invalid finish structure or citation arguments.

Joint completion of H1/H2/C/full restoration gives 3/30 groups at batch 8 and 33/51 at batch 300, covering claims 56, 268, 431, 507, and 729. Healthy repeats match H1 reads, citations, and verdicts within these groups. Two batch-300 read changes come from one claim-729 capture and two repeats: H reads 7114092 and C reads 34753204; both return NEI. Full restoration recovers both reference read sets. One selected restoration also returns H but replaces no row difference while 101 scores remain changed. One transplantation fails to complete.

Eight claim-268 groups across three captures change citations from sentence 1 to sentence 2 of document 29381091, with unchanged reads and SUPPORT verdict. Full restoration recovers every reference citation. The selected restoration that returns H replaces no row difference. No baseline Gemma result shows a partial-row mask that both restores H and reproduces C.

Six claim-431 verdict differences involve two captures and three repeats each. H1/H2/C read 19922508 and cite sentence 3, but C changes NEI to correct SUPPORT. Full restoration returns H's verdict, while selected restoration and transplantation miss their respective targets. Appendix \ref{sec:J.10} reports the different result under the explicit-example prompt.

All 243 masked trajectories return uncited NEI. Joint full/masked comparisons have H1 40 pairs with 28 verdict changes, H2 39/27, and C 39/36. Masked read discordance is 0/30 for C versus H1 and 0/30 for H2 versus H1 at batch 8, and 31/51 and 0/51 at batch 300. Completed unselected-control groups cover only claim 350: all 288 claim-1180 attempts exhaust budget. Each of the four Case/query-batch strata has 18 completed groups and zero differences in its structured endpoints.

\subsection{Same-request variability and restoration denominators}\label{sec:J.8}

A post-hoc comparison examines C/selected-restoration pairs in which no changed row value is replaced. Each model schedules 39 such pairs. All 39 Qwen and 18 Gemma pairs jointly complete. Their search observations and first requests are identical, yet some later endpoints differ (Table~\ref{tab:J4}).

\begingroup\small\setlength{\tabcolsep}{3pt}
\begin{longtable}{@{}>{\raggedright\arraybackslash}p{\dimexpr 0.16667\linewidth-5.0pt\relax}>{\raggedright\arraybackslash}p{\dimexpr 0.16667\linewidth-5.0pt\relax}>{\raggedright\arraybackslash}p{\dimexpr 0.16667\linewidth-5.0pt\relax}>{\raggedright\arraybackslash}p{\dimexpr 0.16667\linewidth-5.0pt\relax}>{\raggedright\arraybackslash}p{\dimexpr 0.16667\linewidth-5.0pt\relax}>{\raggedright\arraybackslash}p{\dimexpr 0.16667\linewidth-5.0pt\relax}@{}}
\caption{\textbf{Endpoint variability under identical initial requests.}}\label{tab:J4}\\
\toprule
\textbf{Model} & \textbf{Complete pairs} & \textbf{Different read sets} & \textbf{Different citations} & \textbf{Different verdicts} & \textbf{Different structured actions} \\
\midrule\endfirsthead
\multicolumn{6}{l}{\small Table~\ref{tab:J4} (continued)}\\
\toprule
\textbf{Model} & \textbf{Complete pairs} & \textbf{Different read sets} & \textbf{Different citations} & \textbf{Different verdicts} & \textbf{Different structured actions} \\
\midrule\endhead
\bottomrule\endfoot
Qwen & 39 & 0 & 1 & 1 & 2 \\
Gemma & 18 & 2 & 1 & 0 & 3 \\
\end{longtable}\endgroup

Five technical pairs account for the overlapping endpoint differences. Their first response divergence occurs under identical request content. The other 21 Gemma pairs are incomplete. These differences motivate repeating the candidate state.

\subsection{Effective replacement and restoration}\label{sec:J.9}

All 12 Qwen read-change groups complete selected restoration and transplantation, and neither condition recovers its target set; only nine groups actually replace a changed score. Both Gemma read-change groups complete selected restoration with zero effective replacement; only one transplantation completes. The eight citation-change and six verdict-change groups complete both selected conditions, but no partial-row mask succeeds in both directions. In two Qwen groups, selected restoration recovers the verdict and citations but not the reads, and reverse transplantation fails to reproduce the candidate endpoint.

\subsection{Prompt sensitivity and all attempted outcomes}\label{sec:J.10}

All 559 baseline Gemma budget-exhausted episodes contain rejected calls. Inspection of requests, responses, and prefixes finds no crossed request/response associations. Common errors put the verdict outside \texttt{arguments} or encode sentence indices as strings. All 27 invalid-JSON responses truncate at the output limit.

The explicit-example prompt is specified after the baseline analysis and before the second schedule runs. Tasks, arrays, masks, order, repeats, 16 calls, six reads, and 384 output tokens remain fixed. Examples cover all three verdict labels and recommend no particular document, sentence, or answer. Rejected actions continue to consume budget.

\begin{table}[H]
\centering\small\setlength{\tabcolsep}{3pt}
\caption{\textbf{Completion by model, prompt, and cohort.}}\label{tab:J5}
\begin{tabular}{@{}>{\raggedright\arraybackslash}p{\dimexpr 0.25000\linewidth-4.5pt\relax}>{\raggedright\arraybackslash}p{\dimexpr 0.25000\linewidth-4.5pt\relax}>{\raggedright\arraybackslash}p{\dimexpr 0.25000\linewidth-4.5pt\relax}>{\raggedright\arraybackslash}p{\dimexpr 0.25000\linewidth-4.5pt\relax}@{}}
\toprule
\textbf{Model / prompt} & \textbf{Mechanism} & \textbf{Masking} & \textbf{Unselected controls} \\
\midrule
Qwen / baseline & 567/567 & 243/243 & 576/576 \\
Qwen / explicit examples & 553/567 & 243/243 & 576/576 \\
Gemma / baseline & 269/567 & 243/243 & 288/576 \\
Gemma / explicit examples & 494/567 & 29/243 & 555/576 \\
\bottomrule
\end{tabular}
\end{table}

For baseline Gemma (Table~\ref{tab:J5}), mechanism attempts include 271 budget exhaustions and 27 invalid-JSON outputs; masking has neither, and unselected controls have 288 budget exhaustions. Totals are 800/1,386 completed, 559 budget-exhausted, and 27 invalid-JSON episodes, with no service failures in any cohort.

The schedule under the explicit-example prompt has 2,772 attempts: 2,450 complete and 322 exhaust calls, with no invalid JSON or service failure. Gemma's 73 mechanism and 21 unselected-control failures search but never successfully read, commonly because \texttt{sentence} is added to the read arguments. Its 214 masked failures repeatedly cite undelivered sentences. Of 2,803 invalid masked citations, 2,797 accompany NEI.

Core conditions are H1/H2/C/full restoration. Requiring only H1/H2/C leaves the counts and denominators unchanged. Three Qwen claim-729 groups exhaust budget under H1, H2, and full restoration but finish under C and selected restoration, showing condition-dependent completion.

Qwen's nine citation changes are three claim-431 captures with three repeats each. Gemma's 15 read/citation changes are five combinations for claims 343, 431, and 729, also with three repeats. Full restoration recovers all reference endpoints. Qwen's nine pairs of selected restoration and transplantation complete without recovering their targets. For Gemma, each selected condition has 12 completions without target recovery and three noncompletions.

Among 36 Gemma groups complete under both prompts, read differences rise from two to nine, citation differences from eight to nine, and verdict differences fall from six to zero. Under the explicit-example prompt, only one Gemma H1 full/masked pair jointly completes, compared with 40 under the baseline prompt.

\subsection{Three technical repeats and source retesting}\label{sec:J.11}

At zero temperature, the three seed parameters identify technical request repeats. Reanalysis first requires all three repeats of a capture--claim--condition to complete, then examines within-state agreement and recurrence of C--H1 differences.

All 27 baseline Qwen combinations have three complete repeats. Four change reads in all three, although one has internally varying candidate read sets. Verdict changes occur in two repeats of one combination and one repeat of another. Baseline Gemma verdict differences involve two combinations with three repeats each; read differences appear in two repeats of one combination.

With explicit examples, Qwen's three citation-changing combinations and Gemma's five read/citation-changing combinations recur in all three repeats, without verdict changes. The selected conditions can still vary internally, and completion depends on model and condition.

Two additional rounds of 20 arrays retest the source comparisons, giving eight rounds and 184 arrays in total. Their 96 matched references agree with four independent healthy reference arrays at the corresponding shapes. Reference controls follow each round.

\subsection{Read and citation identities}\label{sec:J.12}

Table~\ref{tab:J7} lists Gemma's evidence selections under the explicit-example prompt.

\begingroup\small\setlength{\tabcolsep}{3pt}
\begin{longtable}{@{}>{\raggedright\arraybackslash}p{\dimexpr 0.09000\linewidth-4.8pt\relax}>{\raggedright\arraybackslash}p{\dimexpr 0.14000\linewidth-4.8pt\relax}>{\raggedright\arraybackslash}p{\dimexpr 0.17000\linewidth-4.8pt\relax}>{\raggedright\arraybackslash}p{\dimexpr 0.17000\linewidth-4.8pt\relax}>{\raggedright\arraybackslash}p{\dimexpr 0.43000\linewidth-4.8pt\relax}@{}}
\caption{\textbf{Documents read and sentences cited by Gemma under the explicit-example prompt.} Sentence indices are zero-based.}\label{tab:J7}\\
\toprule
\textbf{Claim} & \textbf{Capture--claim combinations} & \textbf{H1 document / sentence} & \textbf{C document / sentence} & \textbf{Both verdicts} \\
\midrule\endfirsthead
\multicolumn{5}{l}{\small Table~\ref{tab:J7} (continued)}\\
\toprule
\textbf{Claim} & \textbf{Capture--claim combinations} & \textbf{H1 document / sentence} & \textbf{C document / sentence} & \textbf{Both verdicts} \\
\midrule\endhead
\bottomrule\endfoot
343 & 1 & 7873737 / 0 & 19308127 / 4 & SUPPORT, label-correct \\
431 & 3 & 19922508 / 3 & 22696649 / 1 & SUPPORT, label-correct \\
729 & 1 & 9154703 / 2 & 52944377 / 7 & SUPPORT, label-correct \\
\end{longtable}\endgroup

The corresponding annotations identify 5884524/5, 28937856/5, and 26851674/1. Neither branch matches the annotated evidence. For claim 431, the candidate's cited sentence (22696649/1) describes ROS sensing microbe size and provides no evidence for ROS-mediated FoxO3a activation in neuronal death. Both branches nevertheless return the benchmark-correct SUPPORT label.

Qwen's nine citation changes all involve claim 431: both branches read document 19922508, H1 cites sentence 0, and C cites none. Both return an incorrect NEI. These are three captures with three technical repeats each.

\subsection{Repeated-request design}\label{sec:J.13}

Before any request is run, the final confirmation specifies eight batches covering all 27 combinations under the explicit-example prompt, 11 claims, and seven conditions. Each model schedules 1,512 attempts; the two models schedule 3,024. The model's request seed is zero throughout. Scheduling varies, while arrays, masks, budgets, feedback, precision, and templates remain fixed. Each model uses a new service instance.

During each model schedule, source retesting adds 20 arrays. Ten rounds total 224 arrays, including 120 matched references that agree bitwise with four independent healthy references. Every round includes repeated reference controls.

\subsection{Within-state and between-state contrasts}\label{sec:J.14}

For each capture--claim combination, let \(\mathcal B\) contain batches in which H1/H2/C/full restoration all complete and access the numerical site. For \(m=|\mathcal B|\geq2\), let \(C_i,H1_i\) denote their execution trajectories, \(D\) the selected endpoint, and \(\operatorname{mean}_{i\ne j}a_{ij}=[m(m-1)]^{-1}\sum_{i\in\mathcal B}\sum_{j\in\mathcal B,\;j\ne i}a_{ij}\) the mean over ordered distinct-batch pairs. Using the binary indicator \(\mathbf1\),

\begin{gather*}
d_{CH} = \operatorname{mean}_{i \neq j}\mathbf{1}\{ D\left( C_{i} \right) \neq D\left( H1_{j} \right)\},\\
d_{HH} = \operatorname{mean}_{i \neq j}\mathbf{1}\{ D\left( H1_{i} \right) \neq D\left( H1_{j} \right)\},\quad\quad d_{CC} = \operatorname{mean}_{i \neq j}\mathbf{1}\{ D\left( C_{i} \right) \neq D\left( C_{j} \right)\},\\
\delta_{D} = d_{CH} - \frac{1}{2}\left( d_{HH} + d_{CC} \right).
\end{gather*}

Read membership and cited-sentence sets are primary; verdicts, actions, correctness, and annotation matches are also evaluated. Table~\ref{tab:J9} reports summaries that weight combinations or claims equally, using each combination's available common batches. Because multiple ordered pairs share records, \(\delta_{D}\) is descriptive and missing batches are not treated as unchanged outcomes.

\subsection{Eight-batch confirmation results}\label{sec:J.15}

Strict repeatability requires all four core conditions in all eight batches, one stable H1/H2/full-restoration endpoint, and a different internally stable C endpoint. Strict partial-row bidirectional recovery additionally requires eight complete batches of selected restoration and transplantation, a replacement fraction strictly between zero and one, positive remaining row differences, and recovery of the respective H and C targets.

\begingroup\small\setlength{\tabcolsep}{3pt}
\begin{longtable}{@{}>{\raggedright\arraybackslash}p{\dimexpr 0.14286\linewidth-5.143pt\relax}>{\raggedright\arraybackslash}p{\dimexpr 0.14286\linewidth-5.143pt\relax}>{\raggedright\arraybackslash}p{\dimexpr 0.14286\linewidth-5.143pt\relax}>{\raggedright\arraybackslash}p{\dimexpr 0.14286\linewidth-5.143pt\relax}>{\raggedright\arraybackslash}p{\dimexpr 0.14286\linewidth-5.143pt\relax}>{\raggedright\arraybackslash}p{\dimexpr 0.14286\linewidth-5.143pt\relax}>{\raggedright\arraybackslash}p{\dimexpr 0.14286\linewidth-5.143pt\relax}@{}}
\caption{\textbf{All attempts and jointly completed core groups in the eight-batch series.}}\label{tab:J8}\\
\toprule
\textbf{Model} & \textbf{Completed / attempted episodes} & \textbf{Budget exhausted} & \textbf{Joint core groups / scheduled} & \textbf{Read C--H1} & \textbf{Cite C--H1} & \textbf{Verdict C--H1} \\
\midrule\endfirsthead
\multicolumn{7}{l}{\small Table~\ref{tab:J8} (continued)}\\
\toprule
\textbf{Model} & \textbf{Completed / attempted episodes} & \textbf{Budget exhausted} & \textbf{Joint core groups / scheduled} & \textbf{Read C--H1} & \textbf{Cite C--H1} & \textbf{Verdict C--H1} \\
\midrule\endhead
\bottomrule\endfoot
Qwen & 1472/1512 & 40 & 208/216 & 1 & 25 & 0 \\
Gemma & 1154/1512 & 358 & 152/216 & 32 & 35 & 0 \\
\end{longtable}\endgroup

Together, 2,626 attempts complete and 398 exhaust the budget (Table~\ref{tab:J8}), with no service failure or invalid JSON. H2--H1 and full-restoration--H1 discordances are zero for reads, citations, and verdicts in both models.

Qwen has 26/27 combinations complete in all eight core batches. Gemma has 18/27 complete in all eight and two more with some complete batches. All seven Gemma claim-343 combinations have no jointly complete batch. Claim 343 therefore cannot be evaluated in this eight-batch series.

The three strict Qwen combinations all concern claim 431: both branches read 19922508, H1 cites sentence 0, C cites none, and full restoration recovers H's citation in every batch. Both verdicts are incorrect NEI. Gemma's four strict combinations are three claim-431 captures and one claim-729 capture, with the identities shown in Table~\ref{tab:J7}. Each has stable but different H/C paths, complete restoration, and a correct SUPPORT verdict, while neither branch matches the annotated evidence.

Neither model satisfies strict partial-row bidirectional recovery with the tested mask.

\begingroup\small\setlength{\tabcolsep}{3pt}
\begin{longtable}{@{}>{\raggedright\arraybackslash}p{\dimexpr 0.14286\linewidth-5.143pt\relax}>{\raggedright\arraybackslash}p{\dimexpr 0.14286\linewidth-5.143pt\relax}>{\raggedright\arraybackslash}p{\dimexpr 0.14286\linewidth-5.143pt\relax}>{\raggedright\arraybackslash}p{\dimexpr 0.14286\linewidth-5.143pt\relax}>{\raggedright\arraybackslash}p{\dimexpr 0.14286\linewidth-5.143pt\relax}>{\raggedright\arraybackslash}p{\dimexpr 0.14286\linewidth-5.143pt\relax}>{\raggedright\arraybackslash}p{\dimexpr 0.14286\linewidth-5.143pt\relax}@{}}
\caption{\textbf{Between-state differences adjusted for within-state variation.} Fractions are percentages; adjusted differences are percentage points.}\label{tab:J9}\\
\toprule
\textbf{Model / endpoint} & \textbf{Eligible combinations / claims} & \textbf{\(d_{CH}\)} & \textbf{\(d_{HH}\)} & \textbf{\(d_{CC}\)} & \textbf{\(\delta_D\), combinations weighted equally} & \textbf{\(\delta_D\), claims weighted equally} \\
\midrule\endfirsthead
\multicolumn{7}{l}{\small Table~\ref{tab:J9} (continued)}\\
\toprule
\textbf{Model / endpoint} & \textbf{Eligible combinations / claims} & \textbf{\(d_{CH}\)} & \textbf{\(d_{HH}\)} & \textbf{\(d_{CC}\)} & \textbf{\(\delta_D\), combinations weighted equally} & \textbf{\(\delta_D\), claims weighted equally} \\
\midrule\endhead
\bottomrule\endfoot
Qwen / read membership & 26 / 10 & 0.481\% & 0.000\% & 0.962\% & 0.000 & 0.000 \\
Qwen / cited sentences & 26 / 10 & 12.019\% & 0.000\% & 0.962\% & 11.538 & 10.000 \\
Gemma / read membership & 20 / 10 & 20.000\% & 0.000\% & 0.000\% & 20.000 & 20.000 \\
Gemma / cited sentences & 20 / 10 & 21.875\% & 0.000\% & 3.393\% & 20.179 & 20.119 \\
\end{longtable}\endgroup

For one Qwen claim-507 capture, C changes its read set in only one batch. The resulting \(d_{CH} = 0.125\), \(d_{CC} = 0.25\), and \(d_{HH} = 0\) give \(\delta_{D} = 0\). Gemma has within-candidate citation variation in two claim-268 combinations; the three additional single-batch citation differences are excluded from its four strict combinations.

Among first responses given while request histories are still identical, 26 Qwen and 504 Gemma record pairs differ in response content. Pairs share records and need not differ at the structured endpoints. Requests after their histories diverge are excluded from same-input comparisons. Strict repeatability is evaluated on the sets of documents read and sentences cited.

\FloatBarrier
\section{hERG assay allocation experiments}\label{app:I}

\subsection{Model, data and execution}\label{sec:I.1}

The model and execution settings were fixed across conditions (Table~\ref{tab:Iparams}); prediction calibration and the selection and triage rules were fixed before capture. Internal evaluation on 2,009 held-out compounds achieved an area under the receiver operating characteristic curve (AUROC) of 0.814; the healthy Top-402 contained 351 compounds labeled as blockers in the benchmark (87.31\%) at about 20\% of the pool. A pass computes the cohort with all three members; a source is one saved pass under one condition, and a technical block groups captures with their matched references. Molecular graphs and input order are preserved. Selection ranks calibrated probabilities with lexical identifier ties; triage assigns compounds to patch-clamp testing at or above the fixed threshold and to advancement without testing otherwise.

\begingroup\small\setlength{\tabcolsep}{3pt}
\begin{longtable}{@{}>{\raggedright\arraybackslash}p{0.24\linewidth}>{\raggedright\arraybackslash}p{0.72\linewidth}@{}}
\caption{\textbf{Fixed settings of the hERG experiments.}}\label{tab:Iparams}\\
\toprule
\textbf{Setting} & \textbf{Value} \\
\midrule\endfirsthead
\toprule
\textbf{Setting} & \textbf{Value} \\
\midrule\endhead
\bottomrule\endfoot
Ensemble & Three Chemprop D-MPNN members; hidden size 300; message-passing depth 3; mean aggregation; two-layer feed-forward classifier; batch normalization off \\
Batching & 128 compounds per batch, the last holding 89; seven linear calls per member per batch; 336 linear observations per pass \\
Calibration temperature & 1.0546289419746864 \\
Triage threshold & 0.30475174995423615; assign to patch-clamp testing if the calibrated probability is at least this value \\
Selection & 402 of 2,009 compounds; exact ties ordered by compound identifier \\
Margin threshold \(\varepsilon\) & 0.009354012791251587, twice the largest absolute calibrated R1 residual in the retrospective series (0.0046770063956257935) \\
Agent framework & Native Biomni A1; 224 standard tools and the project prediction tool \\
Agent model & \texttt{gpt-6.1-sol}; temperature zero; no explicit seed \\
Agent execution & Each trajectory ran in a fresh process \\
\end{longtable}\endgroup

The retrospective series used two technical blocks with four repeats per condition, the first primary. Native Biomni cohorts covered selection, triage and 51 trajectories used to evaluate restoration budgets without rerunning the predictor. Confirmation used eight balanced blocks, each with a pass under each Case, BF16 matched references on the hardware that produced the faulty passes before and after its two Case passes, and a healthy reference obtained with the same GPU model on the same host. The eight healthy reference states and 16 matched reference states had identical saved predictions and linear-layer inputs and outputs. Biomni's execution graph was unchanged, and the agent received neither expected decisions nor fault locations. The single-request experiment reused the 16 confirmation fault sources and one shared healthy reference without new captures; trusted recomputation on a healthy data-center GPU reproduced that reference bit for bit (Appendix~\ref{sec:I.7}).

\subsection{Object-level agent decisions}\label{sec:I.2}

\begingroup\small\setlength{\tabcolsep}{3pt}
\begin{longtable}{@{}>{\raggedright\arraybackslash}p{0.17\linewidth}>{\centering\arraybackslash}p{0.09\linewidth}>{\centering\arraybackslash}p{0.17\linewidth}>{\centering\arraybackslash}p{0.15\linewidth}>{\centering\arraybackslash}p{0.15\linewidth}>{\centering\arraybackslash}p{0.17\linewidth}@{}}
\caption{\textbf{hERG predictions and compound allocation in the confirmation series.} Each source covers all 2,009 compounds and three ensemble members. Comparisons use the healthy BF16 reference obtained with the same GPU model on the same host.}\label{tab:5}\\
\toprule
\textbf{BF16 condition} & \textbf{Sources} & \textbf{Changed ensemble means per source} & \textbf{Changed Top-402 sets} & \textbf{Changed triage vectors} & \textbf{Labeled blockers in Top-402} \\
\midrule\endfirsthead
\toprule
\textbf{BF16 condition} & \textbf{Sources} & \textbf{Changed ensemble means per source} & \textbf{Changed Top-402 sets} & \textbf{Changed triage vectors} & \textbf{Labeled blockers in Top-402} \\
\midrule\endhead
\bottomrule\endfoot
Healthy reference & 8 & 0 & 0/8 & 0/8 & 351/402 \\
Case 1 & 8 & 18--61 & 0/8 & 1/8 & 351/402 \\
Case 2 & 8 & 287--313 & 6/8 & 6/8 & 351/402 \\
\end{longtable}\endgroup

Decisions in this subsection are outputs of native Biomni trajectories in the confirmation series. Each combination of source and task had three repeats with identical primary decisions, and all 36 Case~2 trajectories whose inputs changed a decision applied the stated rule to their input. Probabilities are calibrated ensemble means, the reference is the healthy BF16 state from the same host, and the triage threshold is given in Table~\ref{tab:Iparams}.

\begingroup\small\setlength{\tabcolsep}{3pt}
\begin{longtable}{@{}>{\raggedright\arraybackslash}p{0.10\linewidth}>{\raggedright\arraybackslash}p{0.46\linewidth}>{\raggedright\arraybackslash}p{0.40\linewidth}@{}}
\caption{\textbf{Compounds that left or entered the 402-compound list.} Values are the healthy rank, then healthy \(\rightarrow\) source calibrated probability. R1 restored every changed list.}\label{tab:I1}\\
\toprule
\textbf{Case 2 block} & \textbf{Left the list} & \textbf{Entered the list} \\
\midrule\endfirsthead
\toprule
\textbf{Case 2 block} & \textbf{Left the list} & \textbf{Entered the list} \\
\midrule\endhead
\bottomrule\endfoot
1 & K5612 (397; 0.798067 \(\rightarrow\) 0.786460) & K5629 (406; 0.792903 \(\rightarrow\) 0.795484) \\
2 & K5612 (397; 0.798067 \(\rightarrow\) 0.791613) & K12096 (403; 0.794193, unchanged) \\
3 & K8400 (402; 0.795484 \(\rightarrow\) 0.790324) & K12096 (403; unchanged) \\
4, 5 & none & none \\
6 & K5612 (\(\rightarrow\) 0.786460); K8400 (\(\rightarrow\) 0.791613) & K12096 (403), K2422 (404); both unchanged \\
7 & K5612 (\(\rightarrow\) 0.782601); K8400 (\(\rightarrow\) 0.792903); K8429 (211; 0.879586 \(\rightarrow\) 0.584045) & K12096, K2422 (unchanged); K5629 (\(\rightarrow\) 0.798067) \\
8 & K8429 (\(\rightarrow\) 0.584045) & K12096 (unchanged) \\
\end{longtable}\endgroup

All six exchanged compounds carry blocker labels. Except for K8429, healthy ranks were 397--406, within 0.0039 of the selection boundary. K12096 and K2422 entered with unchanged predictions when competitors fell; their shared probability 0.794193 gave K12096 the single vacancy in block~8 by lexical order. K8429 fell from rank 211 to 821 in block~7, with ensemble standard deviation rising from 0.0194 to 0.4165, above the reference maximum 0.3517.

\begingroup\small\setlength{\tabcolsep}{3pt}
\begin{longtable}{@{}>{\raggedright\arraybackslash}p{0.14\linewidth}>{\raggedright\arraybackslash}p{0.11\linewidth}>{\centering\arraybackslash}p{0.06\linewidth}>{\raggedright\arraybackslash}p{0.20\linewidth}>{\raggedright\arraybackslash}p{0.11\linewidth}>{\raggedright\arraybackslash}p{0.11\linewidth}>{\raggedright\arraybackslash}p{0.11\linewidth}@{}}
\caption{\textbf{Compounds whose assignment to patch-clamp testing changed.} The signed margin is the healthy probability minus the threshold.}\label{tab:I2}\\
\toprule
\textbf{Source} & \textbf{Compound} & \textbf{Label} & \textbf{Healthy probability (signed margin)} & \textbf{Source probability} & \textbf{Change} & \textbf{Restored by} \\
\midrule\endfirsthead
\toprule
\textbf{Source} & \textbf{Compound} & \textbf{Label} & \textbf{Healthy probability (signed margin)} & \textbf{Source probability} & \textbf{Change} & \textbf{Restored by} \\
\midrule\endhead
\bottomrule\endfoot
Case 2, block 1 & K4780 & 0 & 0.283293 (\(-0.021459\)) & 0.321968 & assay added & R1 \\
Case 2, block 3 & K12252 & 1 & 0.313189 (\(+0.008437\)) & 0.298736 & assay removed & R1 \\
Case 2, block 3 & K13388 & 0 & 0.305025 (\(+0.000273\)) & 0.303768 & assay removed & R1 \\
Case 2, block 3 & K13418 & 0 & 0.304710 (\(-0.000041\)) & 0.305339 & assay added & R1+M \\
Case 2, block 5 & K4780 & 0 & 0.283293 (\(-0.021459\)) & 0.335736 & assay added & R1 \\
Case 2, block 6 & K13388 & 0 & 0.305025 (\(+0.000273\)) & 0.304710 & assay removed & R1 \\
Case 2, block 6 & K1461 & 1 & 0.305339 (\(+0.000587\)) & 0.304082 & assay removed & R1 \\
Case 2, block 7 & K13388 & 0 & 0.305025 (\(+0.000273\)) & 0.304396 & assay removed & R1 \\
Case 2, block 7 & K4780 & 0 & 0.283293 (\(-0.021459\)) & 0.306282 & assay added & R1 \\
Case 2, block 8 & K13388 & 0 & 0.305025 (\(+0.000273\)) & 0.304710 & assay removed & R1 \\
Case 2, block 8 & K4780 & 0 & 0.283293 (\(-0.021459\)) & 0.311306 & assay added & R1 \\
Case 2, block 8 & K9120 & 0 & 0.283609 (\(-0.021143\)) & 0.321968 & assay added & R1 \\
Case 1, block 4 & K13418 & 0 & 0.304710 (\(-0.000041\)) & 0.305339 & assay added & R1 \\
\end{longtable}\endgroup

Against 1,407 reference decisions to assign patch-clamp testing, Case~2 produced six removals and six additions. Removals included blockers K12252 and K1461; the other changes concerned compounds labeled as non-blockers in the benchmark. K13418 and K13388 were closest to the healthy threshold, with adjacent attainable calibrated probabilities about 0.0003 apart. R1 restored eleven of twelve changes; R1+M recovered the remaining K13418 assay in block~3 (R1 residual 0.000629). R1 also restored the Case~1 block~4 change to K13418.

\begingroup\small\setlength{\tabcolsep}{3pt}
\begin{longtable}{@{}>{\raggedright\arraybackslash}p{0.30\linewidth}>{\centering\arraybackslash}p{0.33\linewidth}>{\centering\arraybackslash}p{0.33\linewidth}@{}}
\caption{\textbf{Visibility of changed compounds.} Three trajectories per combination of source and task; none called the prediction tool. Tables concern boundaries, cut-off ties or threshold neighborhoods; compound identities are listed above.}\label{tab:I3}\\
\toprule
\textbf{Case 2 block, task} & \textbf{Tables listing affected compounds} & \textbf{Reports naming affected compounds} \\
\midrule\endfirsthead
\toprule
\textbf{Case 2 block, task} & \textbf{Tables listing affected compounds} & \textbf{Reports naming affected compounds} \\
\midrule\endhead
\bottomrule\endfoot
1, selection & 2/3 & 0/3 \\
2, selection & 2/3 & 3/3 \\
3, selection & 2/3 & 3/3 \\
6, selection & 3/3 & 1/3 \\
7, selection & 2/3 & 3/3 \\
8, selection & 2/3 & 1/3 \\
1, triage & 0/3 & 0/3 \\
3, triage & 3/3 & 1/3 \\
5, triage & 0/3 & 0/3 \\
6, triage & 2/3 & 2/3 \\
7, triage & 3/3 & 1/3 \\
8, triage & 3/3 & 1/3 \\
Total & 24/36 & 16/36 \\
\end{longtable}\endgroup

Boundary or tie tables appeared in 30/36 trajectories whose inputs changed a decision; 24 listed an affected compound (Table~\ref{tab:I3}). K8429, displaced to rank 821, appeared in none. K4780 appeared in all three tables for block~7 at margin 0.0015, and none of the six tables for blocks~1 and~5 at margins 0.017 and 0.031. A report for block~6 advanced K13388 at margin \(-0.000041\), versus its healthy \(+0.000273\). Named compounds were boundary or near-threshold cases. All 36 reports stated that the inputs were finite; 18 reported reproducing the calibration within \(1.67\times10^{-16}\). No output used ensemble disagreement as a decision criterion, called the predictor or referred to hardware faults.

In the 16 retrospective BF16 sources, decisions implied by the inputs changed for eight of the twelve confirmation compounds: K5612, K8400, K12096, K5629 and K2422 in selection, and K13388, K4780 and K13418 in triage. K13388 was assigned to advancement without testing in six retrospective sources, and K5612 left the list in three. The faults moved compounds at healthy margins from 0.00004 to 0.085, so the affected set is defined jointly by the decision boundaries and by where the faulty arithmetic displaces values.

\subsection{Retrospective coordinate budgets and biochemical anchors}\label{sec:I.3}

Continuation resumed at the output of the second feed-forward layer, before the final activation and output layer. All reference and candidate member predictions replayed bit for bit in three BF16 passes. The editable target scope was \(3\times300\) coordinates, against 1,808,100 for the cohort; every integer prefix of each prespecified coordinate order was tested.

Matches of the complete InChIKey against ChEMBL 35 link K5612 to CHEMBL3804918 (radioligand displacement, IC50 434 \(\pm\) 29~nM, Ki 217~nM) and K7714 to CHEMBL2207660 (radioligand displacement, IC50 1,995~nM) \citep{ref62,ref63}. In the retrospective BF16 selection tasks, K5612 fell from rank 397 to 415 or 421, and matching restoration returned it to the list. In the retrospective triage task, K7714 was assigned to advancement without testing while the number assigned to testing remained 1,407.

Coordinates were ordered before scanning by their mean absolute difference across eight earlier development passes, with ties broken by index; a co-occurrence order and three random orders were retained, and biological labels did not determine priority. Strict locality required each touched member row to retain differences and the visible prediction vector to differ from the destination. The numbers of differing coordinates were 291, 567 and 850 in the target rows, and 135,225, 138,222 and 134,646 across the cohort.

\begingroup\small\setlength{\tabcolsep}{3pt}
\begin{longtable}{@{}>{\raggedright\arraybackslash}p{0.36\linewidth}>{\centering\arraybackslash}p{0.25\linewidth}>{\centering\arraybackslash}p{0.35\linewidth}@{}}
\caption{\textbf{First successful prefixes and numbers of strict bidirectional partial prefixes with coordinates ordered by mean absolute differences in the development captures.} R is reference restoration and T is reciprocal transplantation.}\label{tab:I5}\\
\toprule
\textbf{Source} & \textbf{First R / T prefix} & \textbf{Number of strict bidirectional partial prefixes} \\
\midrule\endfirsthead
\toprule
\textbf{Source} & \textbf{First R / T prefix} & \textbf{Number of strict bidirectional partial prefixes} \\
\midrule\endhead
\bottomrule\endfoot
K5612, block 1 & 64 / 27 & 227 \\
K5612, block 2 & 254 / 24 & 312 \\
K7714, triage task & 11 / 547 & 281 \\
\end{longtable}\endgroup

The three restored decisions coexisted with 303, 317 and 276 calibrated cohort predictions still differing from the reference, respectively. Each touched member row retained differences under the strict locality rule.

With the co-occurrence order and three random orders, first successful prefixes ranged from 64 to 458 restored and 24 to 160 transplanted coordinates for K5612, and from 1 to 40 restored and 635 to 846 transplanted coordinates for K7714. Reference inputs restored the full decisions. The 22 conditions contained 17 distinct inputs, each evaluated in three trajectories (51 total).

\subsection{Arithmetic verification and timing}\label{sec:I.4}

The audit covered eight BF16 passes per Case and eight each for reference computation on the same device and restoration. Removing duplicate calculations with the same ensemble member, layer, operands and outputs left 5,232 linear calculations. For \(Y=XW^{\top}+b\), BF16-quantized operands gave FP64 expected row and column sums on a healthy GPU, with allowances for accumulation, intermediate and final rounding. Scope and thresholds were fixed before audit; an independent CPU implementation agreed on all calculations. Checksums localized 695 of 991 Case~1 observations with errors identified by elementwise checks and all 2,256 Case~2 observations. The 16 reference and restoration passes were unflagged, with no false alarm in eight healthy verifier runs.

Per Case, integrity checks on stored data and checks that final probabilities were finite and between 0 and 1 each flagged 0/8 passes; row and column checksums with a fixed rounding bound and elementwise arithmetic recomputation each flagged 8/8.

In the confirmation series, recomputing the R1 units with the original batch geometry retained the K13418 triage discrepancy, and restoration with reference values recovered every member prediction in all eight Case~2 sources.

\begingroup\small\setlength{\tabcolsep}{3pt}
\begin{longtable}{@{}>{\raggedright\arraybackslash}p{0.31\linewidth}>{\centering\arraybackslash}p{0.19\linewidth}>{\centering\arraybackslash}p{0.21\linewidth}>{\centering\arraybackslash}p{0.23\linewidth}@{}}
\caption{\textbf{Timing on healthy GPUs.} Medians of eight repetitions per method; columns report total times for separate methods. Data-center GPU rows use healthy inputs on the same device, include calibration and output writing, and share the BF16 forward and double-execution baselines.}\label{tab:I7}\\
\toprule
\textbf{Context} & \textbf{BF16 forward} & \textbf{BF16 forward with verification} & \textbf{Two BF16 passes and comparison} \\
\midrule\endfirsthead
\toprule
\textbf{Context} & \textbf{BF16 forward} & \textbf{BF16 forward with verification} & \textbf{Two BF16 passes and comparison} \\
\midrule\endhead
\bottomrule\endfoot
Confirmation on the same host (RTX) & 0.4883~s & 0.8956~s & 0.9761~s \\
Margin experiment (RTX) & 0.50383~s & 1.18486~s & 1.00587~s \\
Data-center GPU, original FP64 checks & 0.4673~s & 1.0204~s & 0.8816~s \\
Data-center GPU, streaming FP64 checks & 0.4673~s & 0.5877~s & 0.8816~s \\
\end{longtable}\endgroup

The confirmation R1 medians were 0.2164--0.3197~s for 2,168--3,283 recomputed predictions from individual ensemble members per Case~2 source. The margin step took a median of 0.0333~s, including selection, graph repacking, transfer to the GPU, inference and writeback. Timings exclude model loading, graph construction, cross-machine transfer and archival.

The streaming implementation shares signed and absolute FP64 tile reductions for row and column checks. A conservative arithmetic guard determines the original FP64 check's flag decision when rounding bounds leave it unambiguous, and otherwise invokes that check. Across the 16 fault sources, flags and localization matched the original checks, and R1 and R1+M arrays were bitwise identical. Relative to the original FP64 implementation, streaming reduced the verified forward time by 42.4\% and the added verification time by 78.2\%; its overhead relative to the native forward path was 25.8\%.

\subsection{Adding a margin rule to R1}\label{sec:I.5}

Development used the 16 retrospective R1 sources. The threshold \(\varepsilon\) was twice their largest absolute calibrated residual relative to the healthy reference (Table~\ref{tab:Iparams}); it was fixed before evaluation on the 16 confirmation sources. The rule uses only current R1 probabilities \(\hat p_i\). The triage margin is \(|\hat p_i-t|\). With compounds sorted by decreasing probability and lexical ties, the selection margin is \(\hat p_i-\hat p_{(403)}\) for selected compounds and \(\hat p_{(402)}-\hat p_i\) otherwise. The union of compounds with either margin strictly below \(\varepsilon\) was recomputed once, all three members, in original order and in batches of at most 128, while unselected predictions remained unchanged. Reference predictions and labels were inaccessible during recomputation.

The rule selected 68 compounds, 39 through the triage margin and 29 through the ranking margin, with no overlap, and the same 68 in every source evaluated. Multipliers 0, 0.25, 0.5, 1, 2 and 4 of \(\varepsilon\) selected 0, 16, 34, 68, 146 and 306 compounds; triage agreement was 15/16 at multiplier 0 and 16/16 at every nonzero multiplier, and multiplier 1 remained the primary policy. Residuals that changed decisions were small, 0.000629 for K13388 in a retrospective Case~1 source and 0.000629 for K13418, whereas a residual of 0.0311 for K4999, whose healthy probability is 0.6468, changed neither decision; its R1 score remained 0.18 from the nearer boundary. The five confirmation residuals larger than the largest retrospective residual, two of them also larger than \(\varepsilon\) (0.0311 for K4999 and 0.0111 for K8408), all lay at least 0.064 from the nearer boundary after R1 and changed no decision. Six benign BF16 settings on healthy hardware, each repeated eight times, reproduced the BF16 predictions and decisions bit for bit.

\begingroup\small\setlength{\tabcolsep}{3pt}
\begin{longtable}{@{}>{\raggedright\arraybackslash}p{0.24\linewidth}>{\centering\arraybackslash}p{0.13\linewidth}>{\centering\arraybackslash}p{0.13\linewidth}>{\centering\arraybackslash}p{0.11\linewidth}>{\centering\arraybackslash}p{0.13\linewidth}>{\centering\arraybackslash}p{0.13\linewidth}@{}}
\caption{\textbf{Additional recomputation in the 16 confirmation sources.} Compounds and predictions are recomputed per source; selection and triage count sources whose outputs match the healthy reference; the last column counts R1 residual decisions recovered.}\label{tab:I8}\\
\toprule
\textbf{Strategy} & \textbf{Compounds} & \textbf{Predictions} & \textbf{Selection} & \textbf{Triage} & \textbf{Residuals} \\
\midrule\endfirsthead
\toprule
\textbf{Strategy} & \textbf{Compounds} & \textbf{Predictions} & \textbf{Selection} & \textbf{Triage} & \textbf{Residuals} \\
\midrule\endhead
\bottomrule\endfoot
R1 only & 0 & 0 & 16/16 & 15/16 & 0/1 \\
R1+M & 68 & 204 & 16/16 & 16/16 & 1/1 \\
Random, each of 20 fixed seeds & 68 & 204 & 16/16 & 15/16 for every seed & 0/1 for every seed \\
Ordered by ensemble disagreement & 68 & 204 & 16/16 & 15/16 & 0/1 \\
\end{longtable}\endgroup

\subsection{Native agent continuation}\label{sec:I.6}

\begingroup\small\setlength{\tabcolsep}{3pt}
\begin{longtable}{@{}>{\raggedright\arraybackslash}p{0.33\linewidth}>{\raggedright\arraybackslash}p{0.12\linewidth}>{\centering\arraybackslash}p{0.12\linewidth}>{\centering\arraybackslash}p{0.17\linewidth}>{\centering\arraybackslash}p{0.20\linewidth}@{}}
\caption{\textbf{All 114 continuation trajectories.} The reference decision is the agent's output on contemporary healthy inputs.}\label{tab:I9}\\
\toprule
\textbf{Input condition} & \textbf{Task} & \textbf{Trajectories} & \textbf{Agreement with own input rule} & \textbf{Agreement with reference decision} \\
\midrule\endfirsthead
\toprule
\textbf{Input condition} & \textbf{Task} & \textbf{Trajectories} & \textbf{Agreement with own input rule} & \textbf{Agreement with reference decision} \\
\midrule\endhead
\bottomrule\endfoot
R1+M, 16 sources & Selection & 48 & 48/48 & 48/48 \\
R1+M, 16 sources & Triage & 48 & 48/48 & 48/48 \\
Case 1 block 4, original input & Selection & 3 & 3/3 & 3/3 \\
Case 1 block 4, original input & Triage & 3 & 3/3 & 0/3 \\
Case 1 block 4, R1 input & Selection & 3 & 3/3 & 3/3 \\
Case 1 block 4, R1 input & Triage & 3 & 3/3 & 3/3 \\
Healthy BF16 reference & Selection & 3 & 3/3 & 3/3 \\
Healthy BF16 reference & Triage & 3 & 3/3 & 3/3 \\
\end{longtable}\endgroup

The three triage outputs on the original inputs for Case~1 block~4 consistently added K13418, giving 1,408 decisions to assign patch-clamp testing against 1,407 in the reference, and the R1 outputs recovered the reference action. All 38 combinations of source and task had three valid trajectories with identical primary decisions, and all 19 selection groups also had identical order within the list. The R1 and R1+M inputs for block~4 were identical but kept separate planned repeats, so the 19 conditions represent 18 distinct prediction tables. No trajectory called the prediction service.

\subsection{Recovery and commitment for each request}\label{sec:I.7}

Each request is defined by one source, task and repeat: the 16 confirmation fault sources and one shared healthy reference, two tasks and three repeats give 102 requests. The 39 changed decisions came from 13 combinations of source and task, each repeated three times. Each request was evaluated separately under three recovery conditions: recovery before the agent read the predictions, file replacement after it had read them, and restoration of the execution state saved before that read. The two later conditions shared one executed prefix in which native Biomni read the prediction table, wrote its report and prepared an action. File replacement corrected the prediction file while retaining visible history, Python variables, derived files and pending actions. State restoration returned to the checkpoint before the first read, restored files and the Python environment, discarded dependent state and pending actions, and continued with the certified table. Prompts and tools were unchanged, and no prompt mentioned faults or recomputation.

Every condition obtained a certificate from full trusted recomputation, including runs without an arithmetic flag. Each condition had a separate local commit record, so the limit of one commitment per request applied within each condition. In one transaction, it checked the request, certificate, computation contract, input view, agreement with the certified decision and prior commitment; it recorded accepted actions without changing the agent's proposal. Commitment was a local record of an accepted assay allocation. When only the prediction file was replaced, all 96 actions on fault inputs were rejected because the input view used to prepare them had been superseded, including 57 that agreed with the healthy decision. The six accepted actions were healthy controls (Table~\ref{tab:I10}).

\begingroup\small\setlength{\tabcolsep}{3pt}
\begin{longtable}{@{}>{\raggedright\arraybackslash}p{0.43\linewidth}*{3}{>{\centering\arraybackslash}p{0.17\linewidth}}@{}}
\caption{\textbf{Decisions and commitments by recovery condition.} The first four rows count agreement with the healthy decision. Rejected actions are grouped by agreement with that decision; all 96 rejections recorded a superseded input view. Across the 306 outcomes from the three conditions, 210 actions were committed, none deviating from the certified decision or recorded twice.}\label{tab:I10}\\
\toprule
\textbf{Outcome} & \textbf{Before reading} & \textbf{File replacement} & \textbf{State restoration} \\
\midrule\endfirsthead
\toprule
\textbf{Outcome} & \textbf{Before reading} & \textbf{File replacement} & \textbf{State restoration} \\
\midrule\endhead
\bottomrule\endfoot
Case 1 & 48/48 & 45/48 & 48/48 \\
Case 2 & 48/48 & 12/48 & 48/48 \\
Healthy controls & 6/6 & 6/6 & 6/6 \\
Changed decisions recovered & 39/39 & 0/39 & 39/39 \\
Actions committed & 102 & 6 & 102 \\
Actions rejected, decision deviating & 0 & 39 & 0 \\
Actions rejected, decision matching & 0 & 57 & 0 \\
\end{longtable}\endgroup

On the saved arrays, R1 recovered both decisions in all 16 sources except one triage vector (Case~2, block~3), and R1+M in all 16, while 5--16 and 4--15 of 6,027 member predictions, respectively, still differed. Full trusted recomputation eliminated the remaining differences; all 306 certified computations produced the same prediction array.

\begingroup\small\setlength{\tabcolsep}{3pt}
\begin{longtable}{@{}>{\raggedright\arraybackslash}p{0.66\linewidth}>{\raggedright\arraybackslash}p{0.30\linewidth}@{}}
\caption{\textbf{Numerical processing costs and agent token use.} Entries for saved states list Case~1 / Case~2 medians across eight sources each, with one paired timing per source and method; verification followed by recomputation includes reading, checks, trusted recomputation, calibration and writeback. Timings that include forward execution use healthy inputs and cover one verified forward pass followed by full recomputation, with eight repetitions per method. State operations cover file restoration, checkpoint decoding, table writeback and commit recording. Token counts cover successful responses, with cached input included within input and the shared prefix counted once.}\label{tab:I11}\\
\toprule
\textbf{Item} & \textbf{Value} \\
\midrule\endfirsthead
\toprule
\textbf{Item} & \textbf{Value} \\
\midrule\endhead
\bottomrule\endfoot
Saved state, original checks with trusted recomputation & 8.091 / 8.117~s \\
Saved state, streaming checks with trusted recomputation & 2.731 / 2.783~s \\
Saved state, direct trusted recomputation & 0.455 / 0.453~s \\
Saved state, double execution & 0.870 / 0.863~s \\
Verified forward plus full trusted recomputation (original / streaming checks) & 1.4339 / 1.0021~s \\
Tokens, 96 requests with faulty inputs: analysis after recovery before reading / initial analysis of the faulty predictions & 4,264,281 / 4,224,138 \\
Tokens, 96 requests with faulty inputs: continuation after file replacement / state restoration & 1,252,274 / 2,769,479 \\
All 102 requests: tokens / successful calls & 13,308,407 / 1,569 \\
State operations per request & 5.66~ms \\
\end{longtable}\endgroup

Raw predictions and saved tensors were finite, and the captures raised no warning. All 102 prefixes applied the rule to their own input, and the three repeats of each combination of source and task made the same decision. No trajectory called the prediction service, and no report identified a numerical fault.

\FloatBarrier
\section{Screening experiments}\label{app:xtask}

\subsection{27B agents on new screening inputs}\label{sec:xtask.1}

The schedule fixed 48 ESOL molecular and 48 CITE-seq cell operations before fresh capture \citep{ref47,ref65}. CD3 and CD14 yield 144 contexts, each run under 14 conditions, including repeated reference and fault states and complete reference replay: 2,016 serial Qwen3.8-27B trajectories. Repeated reference and repeated Case~2 states agreed within condition on queried objects, their order and hit count. Measurements were unavailable until queried; the predictor was not refitted (Table~\ref{tab:Jparams}).

Case~2 changed the shortlist in 62 contexts, the objects chosen for measurement in 38 and the number of measurements meeting the task threshold in 11. Every report matched returned measurements; rerunning the agent on reference values recovered all 38 changed choices.

\begingroup\small\setlength{\tabcolsep}{3pt}
\begin{longtable}{@{}>{\raggedright\arraybackslash}p{0.24\linewidth}>{\raggedright\arraybackslash}p{0.72\linewidth}@{}}
\caption{\textbf{Fixed settings of experiments on new inputs.}}\label{tab:Jparams}\\
\toprule
\textbf{Setting} & \textbf{Value} \\
\midrule\endfirsthead
\toprule
\textbf{Setting} & \textbf{Value} \\
\midrule\endhead
\bottomrule\endfoot
ESOL endpoint & Aqueous solubility; supported if logS is at least \(-2\) \\
CITE-seq endpoints & Normalized antibody-derived tag readouts; CD3 threshold 5.937278695830723; CD14 threshold 3.539439707328391 \\
Predictor & Saved input representations, cosine similarities to landmark samples and fitted ensembles of four ridge predictors, retained across numerical conditions \\
Shortlist & 24 candidates nearest the threshold; ties broken by identifier; displayed in identifier order with predicted mean, predictor dispersion and threshold comparison \\
Queries & Two objects measured in sequence; public measurements revealed only when queried \\
27B agent & Qwen3.8-27B served locally; serial client requests; temperature zero; seed zero; thinking disabled; output limit 384 tokens \\
Donor operation & 1,024 by 512 matrix multiplied by a 512 by 512 matrix; BF16-represented operands and output; reduced-precision intermediate reduction disabled; saved in FP32 \\
\end{longtable}\endgroup

\begingroup\small\setlength{\tabcolsep}{3pt}
\begin{longtable}{@{}lccccc@{}}
\caption{\textbf{Case 2 query outcomes on new inputs.} Each domain has 48 contexts. Changed: contexts whose set of measured objects changed. Intervals are in percent; CD3 and CD14 share cell operations, so the cluster bootstrap resamples operations. Order and hit count: contexts whose query order or number of measured objects meeting the threshold changed.}\label{tab:4}\\
\toprule
\textbf{Domain} & \textbf{Changed} & \textbf{Wilson 95\%} & \textbf{Bootstrap 95\%} & \textbf{Order} & \textbf{Hit count} \\
\midrule\endfirsthead
\toprule
\textbf{Domain} & \textbf{Changed} & \textbf{Wilson 95\%} & \textbf{Bootstrap 95\%} & \textbf{Order} & \textbf{Hit count} \\
\midrule\endhead
\bottomrule\endfoot
ESOL & 7/48 & 7.25--27.17 & 6.25--25.00 & 8 & 3 \\
CD3 & 18/48 & 25.22--51.64 & 25.00--52.08 & 18 & 3 \\
CD14 & 13/48 & 16.57--41.00 & 14.58--39.58 & 14 & 5 \\
\end{longtable}\endgroup

Case~2 changed a median of 71.5 coordinates per ESOL operation and 199 per CITE-seq operation, with median maximum displacements of 0.24 and 0.33. For the tool's 24-item shortlist, let \(d_i=|\mu_i-t|\) for reference means \(\mu_i\) and fixed threshold \(t\). Sort \(d\) increasingly and set \(g=d_{(25)}-d_{(24)}\). If \(e=\max_i|\widehat\mu_i-\mu_i|\), then \(\big||\widehat\mu_i-t|-d_i\big|\le e\); hence \(g>2e\) certifies unchanged membership (Appendix~\ref{sec:C.4}). Of the 144 Case~2 comparisons, this sufficient condition certified 16 ESOL shortlists, all unchanged; the shortlist changed in 62.

All 40 Case~2 context pairs with a repeated behavioral difference, 38 changes in the objects chosen for measurement and two changes in secondary endpoints, entered a factorial comparison of list membership and numerical values. Each condition combined reference or candidate list membership with the reference or candidate numerical fields of the same identifiers, with four repetitions per pair (Table~\ref{tab:xtask1}). In mixed displays, choices did not follow the source of the list alone: reference lists with candidate values matched the reference query set in 84 of 160 trajectories, and candidate lists with reference values in 32 of 160. Both which objects are listed and the values attached to them steer the agent's choices.

\begingroup\small\setlength{\tabcolsep}{3pt}
\begin{longtable}{@{}>{\raggedright\arraybackslash}p{0.30\linewidth}>{\centering\arraybackslash}p{0.15\linewidth}>{\centering\arraybackslash}p{0.17\linewidth}>{\centering\arraybackslash}p{0.17\linewidth}>{\centering\arraybackslash}p{0.15\linewidth}@{}}
\caption{\textbf{Factorial comparison of list membership and numerical values.} The first factor is list membership and the second the numerical fields. Matching the reference and matching the candidate are not mutually exclusive in the two contexts whose reference and candidate query sets coincide; the four candidate/candidate trajectories that differ from the candidate come from one context.}\label{tab:xtask1}\\
\toprule
\textbf{List / values} & \textbf{Trajectories} & \textbf{Match reference} & \textbf{Match candidate} & \textbf{Report errors} \\
\midrule\endfirsthead
\toprule
\textbf{List / values} & \textbf{Trajectories} & \textbf{Match reference} & \textbf{Match candidate} & \textbf{Report errors} \\
\midrule\endhead
\bottomrule\endfoot
Reference / reference & 160 & 160 & 8 & 0 \\
Reference / candidate & 160 & 84 & 28 & 0 \\
Candidate / reference & 160 & 32 & 84 & 0 \\
Candidate / candidate & 160 & 8 & 156 & 0 \\
\end{longtable}\endgroup

\subsection{Screening cells from held-out donors}\label{sec:xtask.2}

Donors P7 and P8 were absent from earlier captured rows, the labeled fitting set and landmarks. Existing transformations, predictor and thresholds were retained. Two technical blocks ran in opposite Case order, each with four captures under matched reference, Case~1, Case~2 and fresh recovery; first captures were prespecified as primary. The independent CPU check used represented BF16 operands and bounds for FP32 accumulation and final BF16 rounding; it took 0.061~s, with array comparisons taking 4.7--6.0~ms. All 32 captures completed with finite outputs and raised no warning or exception. Each Case~2 capture had 116--127 coordinates whose errors exceeded the numerical bounds; fresh reference computation recovered both changed primary CD3 classifications per block. Case~1 and CD14 had no changed primary classification.

\begingroup\small\setlength{\tabcolsep}{3pt}
\begin{longtable}{@{}>{\centering\arraybackslash}p{0.06\linewidth}>{\raggedright\arraybackslash}p{0.27\linewidth}>{\centering\arraybackslash}p{0.12\linewidth}>{\centering\arraybackslash}p{0.12\linewidth}>{\centering\arraybackslash}p{0.15\linewidth}>{\raggedright\arraybackslash}p{0.17\linewidth}@{}}
\caption{\textbf{The four distinct changed cells in the primary Case 2 captures.} Values are normalized CD3 readouts; the threshold is given in Table~\ref{tab:Jparams}. All four cells come from donor P7.}\label{tab:xtask2}\\
\toprule
\textbf{Block} & \textbf{Cell} & \textbf{Reference prediction} & \textbf{Case 2 prediction} & \textbf{Measurement} & \textbf{Effect} \\
\midrule\endfirsthead
\toprule
\textbf{Block} & \textbf{Cell} & \textbf{Reference prediction} & \textbf{Case 2 prediction} & \textbf{Measurement} & \textbf{Effect} \\
\midrule\endhead
\bottomrule\endfoot
1 & \texttt{E2L8\_AGCGATTGTGGAAGTC} & 5.905539 & 5.958529 & 5.294847 & new false positive \\
1 & \texttt{E2L8\_AGGACTTGTGGAATGC} & 6.067270 & 5.834704 & 5.769720 & false positive corrected \\
2 & \texttt{E2L8\_CTCAACCGTCTCCTGT} & 5.918845 & 6.041544 & 5.965956 & false negative corrected \\
2 & \texttt{E2L8\_GACTATGTCGAGAGCA} & 5.910636 & 6.008034 & 5.628356 & new false positive \\
\end{longtable}\endgroup

\FloatBarrier
\section{Certified commitment}\label{app:commit}

\subsection{Contract, admissible sets and Proposition 1}\label{sec:commit.1}

Fix inputs, model, representation, execution settings, candidate set and tie rule, defining reference scores \(p^\star\in\mathbb R^n\) and decision map \(g:\mathbb R^n\to\mathcal A\). The hERG contract is healthy BF16 computation (Table~\ref{tab:Iparams}). Trusted recomputation fixes a score only when it reproduces its reference exactly, using the original geometry or regrouped batches when the arithmetic for each item is independent of batch size and composition. Recomputed hERG scores matched their references bit for bit in the tested sources (Section~\ref{sec:herg.recompute}). Other values enter through valid bounds. At round \(r\), current scores are \(s^{(r)}\) and \(F_r\) indexes fixed scores. Assume validity, \(p^\star\in\mathcal U_r\), and persistence, \(F_r\subseteq F_{r+1}\) with \(q_i=p^\star_i\) for all \(q\in\mathcal U_r\), \(i\in F_r\).

\emph{Correctness.} If \(g(q)=a\) for every \(q\in\mathcal U_r\), validity gives \(g(p^\star)=a\), so a committed decision equals the reference decision.

\emph{Termination.} Suppose every round that ends without certification adds at least one index to \(F_r\). If no earlier round certifies, then \(F_n=\{1,\ldots,n\}\); persistence gives \(\mathcal U_n\subseteq\{p^\star\}\), validity gives \(\mathcal U_n=\{p^\star\}\), and \(g\) is constant on \(\mathcal U_n\). Certification is therefore reached within \(n\) rounds.

Certification concerns agreement with the reference computation; Algorithm~3 handles agent and service failures.

\subsection{Interval evidence, thresholds and Top-K sets}\label{sec:commit.2}

Let \(s\) be the current scores with valid bounds \(|s_i-p^\star_i|\le e_i\), where \(e_i=0\) for \(i\in F_r\).
The admissible set is the box \(\mathcal U=\{q:|q_i-s_i|\le e_i\ \text{for all}\ i\}\), which contains \(p^\star\).

\textbf{Corollary 1 (interval evidence).}
\emph{Let \(|s_i-p^\star_i|\le e_i\) for the current scores \(s\), with \(e_i=0\) for recomputed scores.
For the rule \(q_i\ge t\), the decision for item \(i\) is certified when \(s_i-e_i\ge t\) or \(s_i+e_i<t\).
A Top-\(K\) set \(S\) is certified when \(s_i-e_i>s_j+e_j\) for every \(i\in S\) and \(j\notin S\).}

\emph{Proof.}

\emph{Threshold.}
For the rule \(q_i\ge t\), if \(s_i-e_i\ge t\) then \(q_i\ge t\) for every \(q\in\mathcal U\), and if \(s_i+e_i<t\) then \(q_i<t\).
The decision for item \(i\) is therefore constant on \(\mathcal U\).

\emph{Top-K.}
For observed Top-\(K\) set \(S\), the interval condition gives \(q_i\ge s_i-e_i>s_j+e_j\ge q_j\) for every \(q\in\mathcal U\), \(i\in S\), \(j\notin S\).
Thus \(M_K(q)=S\), independently of ties.
For two scores outside \(F_r\) with common bound \(e\), this requires \(s_i-s_j>2e\).

\emph{Re-ranking.}
After each round, update \(S\), boundary scores and margins over the full candidate set.
Changes to one score can move unchanged candidates across the boundary; certification uses the updated bounds and comparisons.

\emph{Relation to R1+M.}
With the margins of Appendix~\ref{sec:I.5}, recomputing all compounds at margin below \(2e\) leaves a gap of at least \(2e\) between unrecomputed scores on opposite sides of the original selection boundary.
Unrecomputed triage scores remain at least \(2e\) from the threshold.
Certification checks the full re-ranked candidate set, including equalities and pairs involving recomputed scores.
R1+M used the retrospective maximum residual as \(e\) in one empirical round with \(\varepsilon=2e\).
Certified intervals require valid per-score bounds.
Row and column checksums locate faulty computations but do not bound individual scores, because errors can cancel within a sum.

The Top-\(K\) condition restates the interval stability criterion in Appendix~\ref{sec:C.4}.

\subsection{Observations, history and commitment}\label{sec:commit.3}

For \(a_t=\pi(H_t)\), certifying \(g\) fixes the input when every returned field is determined by the certified target; other fields must be certified or omitted. In the experiments with selective recomputation, the hERG agents received the full prediction table after R1+M and reproduced the healthy decisions (Section~\ref{sec:actions.recovery}); the single-request experiment certified the full table through trusted recomputation (Appendix~\ref{sec:I.7}). Under fixed continuation conditions, restored input determines the endpoint for deterministic agents and its conditional distribution for stochastic agents (Appendix~\ref{sec:C.4}).

Recovery after consumption replaces the affected output in visible history, discards dependent messages, tool records and pending actions, and invalidates server-side state derived from them, including KV-cache entries for discarded tokens or entries computed on untrusted hardware.

Each pending action is bound to its request, computation contract and certificate. In one atomic step, the commit service checks action agreement, certificate currency and uncommitted status, then records commitment; mismatches are rejected. For hERG, it checks the certified selection set or triage vector. Each request commits at most once; exactly-once external execution also requires an idempotent action service keyed by that request. Before commitment, verification failure, budget exhaustion, agent failure and service failure lead to trusted full recomputation or no commitment.

The single-request experiment evaluates verification before the agent reads the scores and after it has used them, with restoration of the execution state and local records of actions accepted under the current certificate (Appendix~\ref{sec:I.7}).

\end{document}